%% file: main.tex
\documentclass[aps,prd,floatfix,showpacs,superscriptaddress,groupedaddress, onecolumn, nofootinbib,longbibliography]{revtex4-2}
\input{setup.tex}
\begin{document}

\title{Agents for self-driving laboratories applied to quantum computing}

\author{Shuxiang Cao$^{1,\blacklozenge}$}
\email{shuxiang.cao@physics.ox.ac.uk}
\altaffiliation[Present address: ]{
    NVIDIA Corporation, 2788 San Tomas Expressway, Santa Clara, 95051, CA, USA%
  }%
\author{Zijian Zhang$^{2,3, \blacklozenge}$}
\email{zijian@cs.toronto.edu}
\author{Mohammed Alghadeer$^{1}$}
\author{Simone D Fasciati$^{1}$}
\author{Michele Piscitelli$^{1}$}
\author{Mustafa Bakr$^{1}$}
\author{Peter Leek$^{1}$}
\email{peter.leek@physics.ox.ac.uk}
\author{Alán Aspuru-Guzik$^{2,3,4,5,6,7,8}$}
\email{alan@aspuru.com}
\affiliation{$^1$Clarendon Laboratory, Department of Physics, University of Oxford, Oxford, OX1 3PU, UK}
\affiliation{$^2$Department of Computer Science, University of Toronto, Toronto, ON M5S 2E4, Canada}
\affiliation{$^3$Vector Institute for Artificial Intelligence, Toronto, ON, M5G 1M1, Canada}
\affiliation{$^4$Department of Chemistry, University of Toronto, Toronto, ON M5S 3H6, Canada}
\affiliation{$^5$Department of Materials Science \& Engineering, University of Toronto, Toronto, ON M5S 3E4, Canada}
\affiliation{$^6$Department of Chemical Engineering \& Applied Chemistry, University of Toronto, Toronto, ON M5S 3E5, Canada}
\affiliation{$^7$Canadian Institute for Advanced Research
(CIFAR), Toronto, ON M5G 1M1, Canada}
\affiliation{$^8$Acceleration Consortium, Toronto, ON M5S 3H6, Canada}

\def\thefootnote{$\blacklozenge$}\footnotetext{These two authors contributed equally to this work}

\begin{abstract}
Fully automated self-driving laboratories are promising to enable high-throughput and large-scale scientific discovery by reducing repetitive labour. However, effective automation requires deep integration of laboratory knowledge, which is often unstructured, multimodal, and difficult to incorporate into current AI systems.
This paper introduces the \textit{k-agents} framework, designed to support experimentalists in organizing laboratory knowledge and automating experiments with agents. Our framework employs large language model-based agents to encapsulate laboratory knowledge including available laboratory operations and methods for analyzing experiment results. To automate experiments, we introduce execution agents that break multi-step experimental procedures into \sx{agent-based state machine}s, interact with other agents to execute each step and analyze the experiment results. The analyzed results are then utilized to drive state transitions, enabling closed-loop feedback control.
To demonstrate its capabilities, we applied the agents to calibrate and operate a superconducting quantum processor, where they autonomously planned and executed experiments for hours, successfully producing and characterizing entangled quantum states at the level achieved by human scientists.
Our knowledge-based agent system opens up new possibilities for managing laboratory knowledge and accelerating scientific discovery. 
\end{abstract}
\maketitle

\input{1.introduction}

\input{3.results}

\input{4.conclusion}

\input{2.methodology}

\section*{Code availablility}
The source codes of \textit{k-agents} can be found in \cite{kagentsRepo}. The codes for experiments and benchmarks can be found in \cite{LeeQRepo}.

\section*{Conflict of interest disclosure}

A.A.-G. is a founder of Kebotix, Inc., a company specializing in closed-loop molecular discovery, and IntrepidLabs, Inc. a company using self-driving laboratories for pharmaceuticals. P.L. is the founder and Chief Science Officer of Oxford Quantum Circuits Limited, a company developing superconducting circuit quantum computers.

\acknowledgements
The authors thank Yuning Zhang for the insightful discussion. This project is supported by Schmidt Sciences, LLC. S.C.~acknowledges support from Schmidt Science. P.L.~acknowledges support from the EPSRC [EP/T001062/1, EP/N015118/1, EP/M013243/1]. M.B. acknowledges support from EPSRC QT Fellowship grant [EP/W027992/1]. A.A.-G. thanks Anders G. Frøseth for his generous support. A.A.-G. also acknowledges the generous support of Natural Resources Canada and the Canada 150 Research Chairs program. This research is part of the University of Toronto’s Acceleration Consortium, which receives funding from the Canada First Research Excellence Fund (CFREF). 

\bibliography{bib}

\appendix
\input{appendix}

\end{document}

%% file: setup.tex
\usepackage{latexsym}
\usepackage{graphicx}
\usepackage{newtxmath}
\usepackage[colorlinks]{hyperref}
\usepackage{rotating}
\usepackage[normalem]{ulem}
\usepackage{hyperref}
\usepackage{amsmath,amsfonts}
\usepackage{mathtools}
\usepackage{bm}
\usepackage{color}
\usepackage{textgreek}
\usepackage[greek,english]{babel}
\usepackage[many]{tcolorbox}
\usepackage{listings}
\usepackage{tikz} 
\usepackage{pythonhighlight}
\hypersetup{colorlinks=true,urlcolor=blue,linkcolor=blue}

\usepackage{xcolor}
\usepackage{amsfonts}
\usepackage{mathrsfs}
\usepackage{physics}
\usepackage{multirow}
\usepackage{braket}
\usepackage{caption}
\usepackage{subcaption}
\usepackage{algorithm2e}
\RestyleAlgo{ruled}
\usepackage{algpseudocode}
\hypersetup{colorlinks=true,urlcolor=blue,linkcolor=blue}

\newcommand{\sx}[1]{#1}

%% file: 1.introduction.tex
\section{Main}

Automating laboratory experiments has the potential to accelerate scientific discovery by closing the loop between experimental execution, artificial intelligence (AI) and human-in-the-loop decision-making \cite{tom2024self}. Although it is increasingly common in laboratories that experiments can be implemented through programming interfaces \cite{sim2024chemos, seifrid2024chemspyd, 2309.10333}, automating experiments still requires detailed laboratory knowledge to set the parameters of each experiment, interpret the results of the experiment, and execute complicated experiment workflows. 
Traditional methods of automating experiments require human experts to translate their knowledge into code for machine execution \cite{Stein2019,AutoChem2021,XIE2023101043,10.1063/1.5096007,ChemistryAutoLab2022}. 
However, the knowledge required for laboratory work has increased significantly over the past decades due to the growing complexity of the experimental apparatus, which now involves more intricate technical details for operation. 
This makes translating the required knowledge and maintaining consistency an increasingly significant human effort, which compromises the efficiency brought about by automation. In addition, the multimodal and complex nature of laboratory knowledge also makes it challenging to develop automated programs that can perform these tasks at a level comparable to that of humans.

The advent of large language models (LLMs) and multimodal LLMs has sparked new hope for more efficient experiment automation due to their universal ability to process text and image information \cite{brown2020language, openai2023gpt4, team2023gemini, dubey2024llama, liu2024visual,liu2024improved, zhang2024theory}. 
Although limited by the length of their context windows, LLMs offer the hope of efficiently understanding laboratory documents, performing analyses, generating code, and interpreting experimental images. The development of multi-agent systems has further enhanced LLM's capabilities by integrating multiple LLM-based agents to work together on a complex objective, improving both accuracy and scalability \cite{li2023camel,chan2023chateval,wu2023autogen,talebirad2023multi,zhang2024towards,guo2024large,li2024survey}. 
The merits of LLM-based agents make them well suited for assisting laboratory automation without the need for extensive human intervention. 
However, several challenges remain. First, much of the media, such as code and documents, that carry laboratory knowledge is proprietary and inaccessible online, meaning that it is unlikely to be included in the training data of publically available LLMs. In addition, the amount of text that includes such specialized laboratory knowledge is often insufficient to effectively fine-tune an LLM \cite{kaplan2020scaling}. Furthermore, laboratory knowledge is usually dynamic and frequently updated, making fine-tuning models impractical, as it is often hard to rewrite the knowledge in LLMs \cite{li2024unveiling, hoelscher2023detecting, zhang2024comprehensive}. As a result, incorporating laboratory-specific knowledge into LLMs remains difficult. 
Recent work on retrieval-augmented generation (RAG) offers a foundation to equip LLM-based agents with larger-scale knowledge without fine-tuning \cite{lewis2020retrieval, gao2023retrieval, karpukhin2020dense}. However, laboratory knowledge is usually heterogeneous and multimodal, making it challenging to apply standard RAG methods directly \cite{franklin1998experiment, ChemistryAutoLab2022, Stein2019}. 

Despite challenges, pioneering efforts to develop automated LLM-based agents to carry out experiments are already underway \cite{boiko2023autonomous, m2024augmenting, darvish2024organa, yoshikawa2023large}. For example, in \cite{boiko2023autonomous} and \cite{m2024augmenting}, researchers equipped LLM-based agents with tools that allow them to acquire professional knowledge from the expert-designed toolkit and even the Internet. Then, these agents are capable of performing chemistry experiments on the basis of this knowledge. However, in general, previous results lack a scalable memory system and rely on keeping the conversation history when calling LLMs. 
This inherently restricts the agents' ability to automate long-duration tasks involving many sequential steps, as most current LLMs exhibit diminishing performance as the input length increases \cite{liu2023lost, li2024long, li2023long}. This construction also prohibits further extensions of the system, such as supporting the handling of scientific plots.

\begin{figure*}[!t]
    \centering
    \includegraphics[width=0.85\linewidth]{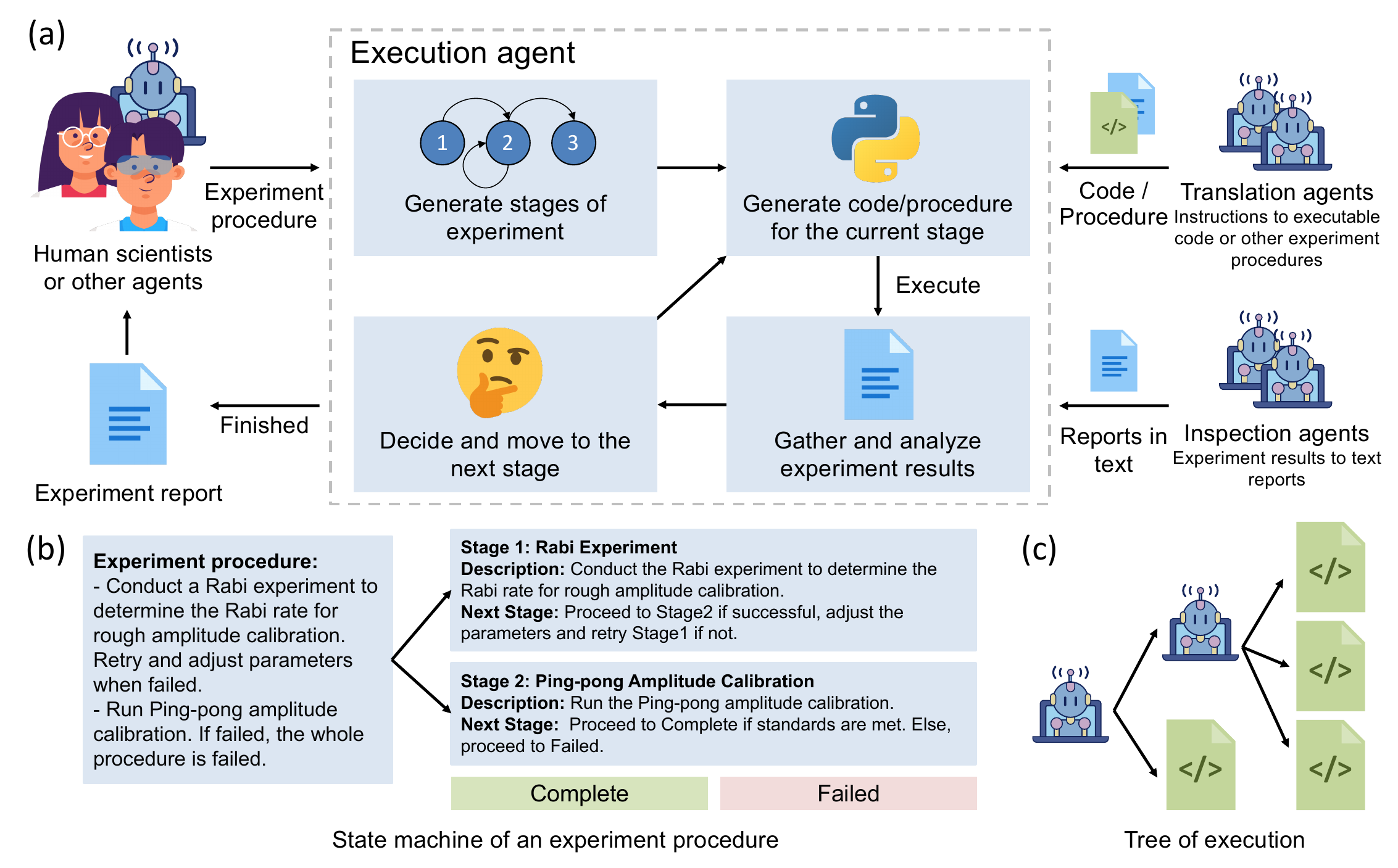}
    \caption{(a) Overview of the \textit{k-agents} framework. Given a procedure in natural language, the execution agent first decomposes it into \sx{an agent-based state machine}, which contains experiment stages. Each stage holds an independent experiment instruction to be translated with the translation agents. The transition between stages is driven by the generated transition rules and the reports from the inspection agents. (b) \sx{Agent-based state machine} of instructions. The figure demonstrates how an experiment procedure can be decomposed into \sx{an agent-based state machine}. The decomposition divides experiment procedures into experiment stages that hold single instructions, along with a transition rule for deciding the next stage. (c) Tree of translation. The translating of instructions can be viewed as an expansion of a tree, in which the translation results are represented as the child nodes. An instruction can be translated as a simple experiment (code node) or an experiment procedure (agent node) that needs further execution and translation with another execution agent.
    }
    \label{fig:framework-overview}
\end{figure*}

In this work, we introduce \textit{k-agents} (see Figure~\ref{fig:framework-overview}), a knowledge-based multi-agent system designed to automate experiments, particularly those requiring large-scale, multimodal laboratory knowledge and complex workflow.
We developed tools for users to transfer their knowledge, creating LLM-based knowledge agents to manage laboratory knowledge without fine-tuning. These knowledge agents can help operate laboratories by holding knowledge ranging from single experiments to complex procedures and how to inspect the results of each experiment. These agents are designed to be activated selectively, allowing the system to scale efficiently as the number of agents increases.
In order to fully exploit the knowledge agents, we further introduce the \textit{execution agent}, which is responsible for coordinating the knowledge agents, gathering and filtering knowledge, generating code (scripts) to operate the laboratory and controlling the progress of complex experiment procedures with intelligent closed-loop feedback planning.
To handle multi-step experiments, the execution agent decomposes complex procedures into independent experiment stages and creates \sx{an agent-based state machine}. \sx{ The agent-based state machine differs from a traditional state machine in that state transitions are determined by agents, rather than by rigid deterministic rules.} This state-machine-based approach minimizes the experimental history that needs to be loaded into LLMs, making it feasible to conduct long-duration experiments at a human-like level of performance. \sx{The agents are created by prompting a language model on a distinct context and the details of the prompt construction can be found in Appendix \ref{app:agents}.}

As a demonstration, we applied our framework to automate the calibration and characterization of single- and two-qubit gates on our superconducting quantum processor. Superconducting qubits have become one of the most widely adopted platforms for quantum computing, with recent advancements pushing the scale to hundreds of qubits, along with active quantum error correction \cite{acharya2024quantum,reichardt2024demonstration,putterman2024hardware,brock2024quantum,2411.10406}. As these systems grow in complexity, calibrating the operations of hundreds of qubits has emerged as a substantial bottleneck. Given this rapid progress, it is timely and essential to study automation solutions that can support the scalability and performance requirements of large-scale superconducting quantum devices. In addition, our framework demonstrates its ability to conduct custom experiments. Specifically, preparing the Greenberger–Horne–Zeilinger (GHZ) state requires the calibration of multiple qubits to execute the desired quantum operations. We selected this experiment to demonstrate the capabilities of our automation technique in generating an entangled quantum state and evaluating its fidelity. We propose that this framework could be adapted to other fields as a model for future research and industrial methodologies.

\begin{figure*}
    \centering
    \includegraphics[width=1\linewidth]{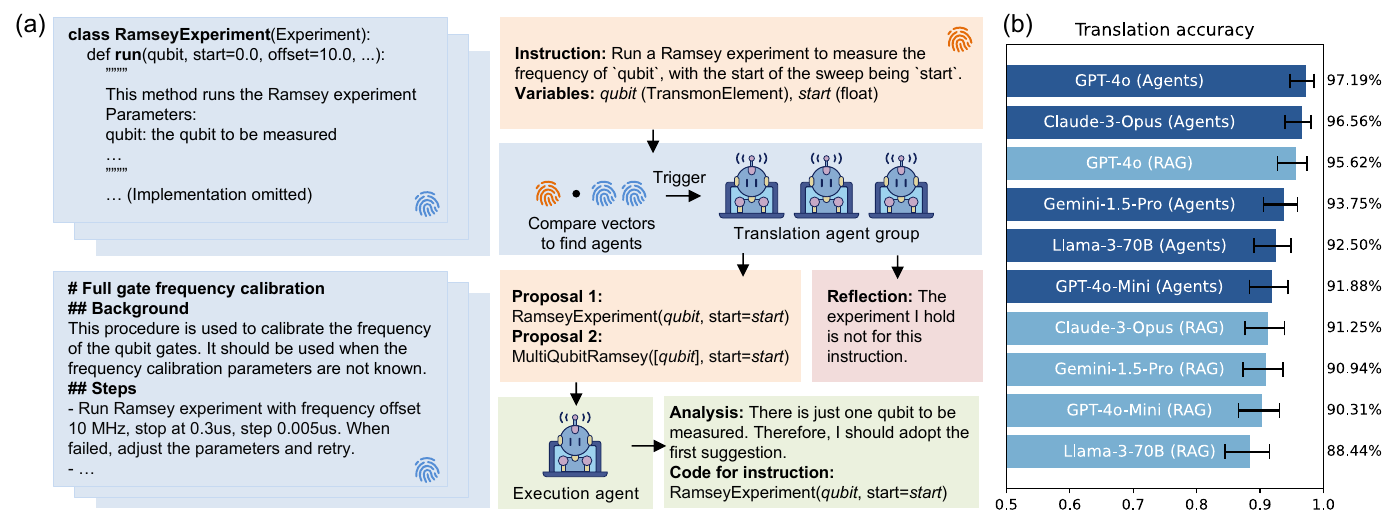}
    \caption{(a) Translation agents. Translation agents are responsible for translating an incoming instruction into executable code. The agents will be activated when their characterizing vectors (blue fingerprints) have a significant overlap with the instruction's vector (orange fingerprint). The activated agents will try to translate the instruction based on their knowledge. If the translation agent deems the translation as valid, the result will be output to the execution agent for a final selection and execution. 
    (b) Accuracy of instruction translation. We prepare a test set containing 80 instructions from 8 experiments to be translated.
    We compare \textit{k-agents} (marked as Agents), which uses Algorithm~\ref{alg:trans}, with standard RAG methods (marked as RAG), which directly load the signature of the experiment class into the prompt. In our setup with 17 code translation agents, we found that \textit{k-agents} demonstrates better performance than standard RAG methods in choosing the correct experiment class. Further, \textit{k-agents} can support heterogeneous translation agents working together, which is hard to achieve with standard RAG methods. Details of this benchmark can be found in the Appendix \ref{app:benchmark}.
    }
    \label{fig:interpret}
\end{figure*}
\subsection{Knowledge agents}

We define \textit{knowledge agents} as AI agents whose performance is measured by their ability to receive and transfer knowledge. For instance, an agent qualifies as a strong knowledge agent if it accepts knowledge from natural language inputs and responds correctly to related queries using natural language. Developing more advanced knowledge agents offers significant advantages. First, a more general ability to accept knowledge reduces the effort needed for humans to translate knowledge into computer-friendly formats. Furthermore, the ability to transfer knowledge facilitates more rigorous testing of agents \cite{krenn2022scientific} and avoids treating the system as a black box, improving the general trustworthiness of AI systems.

\sx{Prior works \cite{boiko2023autonomous, m2024augmenting, darvish2024organa, yoshikawa2023large} has limitations in scalability, which constrain their ability to manage complex toolsets and long-horizon tasks. These systems typically construct a single prompt that embeds both task instructions and the entire tool set. However, as shown in \cite{yao2024tau, wang2024appbench}, performance degrades significantly as the length of input increases. Moreover, these approaches are generally limited to simpler tasks where the entire execution context can be accommodated within the prompt window.}

\sx{To address some of these issues, works such as \cite{boiko2023autonomous} introduce modular architectures that allows the developer to mannually divide tools into components. 
Additionally, they rely on retrieval-augmented generation (RAG) \cite{boiko2023autonomous, m2024augmenting} for data access, which often screens and retrieves documents based on relevance rather than deeper procedural or understanding.}

\sx{To resolve the above issue,} we introduce the \textit{k-agents} framework, which provides tools to implement the aforementioned bidirectional knowledge transferability and \sx{the scalablity towards large number of tools and long-horizen tasks}. In \textit{k-agents}, we implement user-friendly interfaces that enable users to encapsulate knowledge within different LLM-based agents. These agents can then transfer their knowledge to the users and other agents in the format of natural language with LLMs. \sx{Our framework has a scalable memory architecture with an agent-based retrieval mechanism. This enables agents to dynamically interact with a large, indexed memory of procedural knowledge and existing experiments. During retrieval, the agents conduct the reflection and selection processes to improve the accuracy and contextual relevance of retrieved information. This design allows our framework to support more complex and long-duration tasks.}

\textbf{(Code translation agent)} 
This first kind of knowledge agent in \textit{k-agents} is the \textit{code translation agent}, which is designed to accept and apply the knowledge of the available experiment interfaces that can be called by code. Here, the term ``experiment'' stands for not only a direct experiment on an instrument, but any general action that is available in the laboratory. It can also be a call to a tool or even another agent. Given instructions in natural language, these agents are responsible for translating natural language instructions into the corresponding code.
To help the users of \textit{k-agents} construct code translation agents, we provide an abstract Python class that allows users to define the set of experiments available in the laboratory explicitly. The users can then define all possible experiment interfaces, ensuring that the AI system can access and utilize them. As illustrated in Figure~\ref{fig:interpret}(a), human experimentalists are expected to document background knowledge, required parameters, and implementation code for each experiment in the {\sf run} method of the subclass representing the experiment. 
After indexing all available experiment classes, \textit{k-agents} will construct a code translation agent for each of the classes.

\textbf{(Procedure translation agent)} 
Experiment procedures in laboratories can involve complex workflows. These workflows may require experimentalists to determine a sequence of experiments based on the results of the experiments executed.
Traditionally, this knowledge has been maintained either through unstructured documentation or by relying on the memory of the experimentalists themselves. 
To address this, as shown in Figure~\ref{fig:interpret}(a), \textit{k-agents} introduces a standardized format to store examples of how to implement instructions using multistep experiment procedures. 
These examples are then used to produce another kind of knowledge agent that we call the \textit{procedure translation agent}, which is capable of translating an instruction into its corresponding procedure based on stored examples. Similarly to code translation agents, procedure translation agents also output code. The code will call an execution agent to execute the procedure, which we will introduce later. 

\begin{figure*}
    \centering
    \includegraphics[width=1\linewidth]{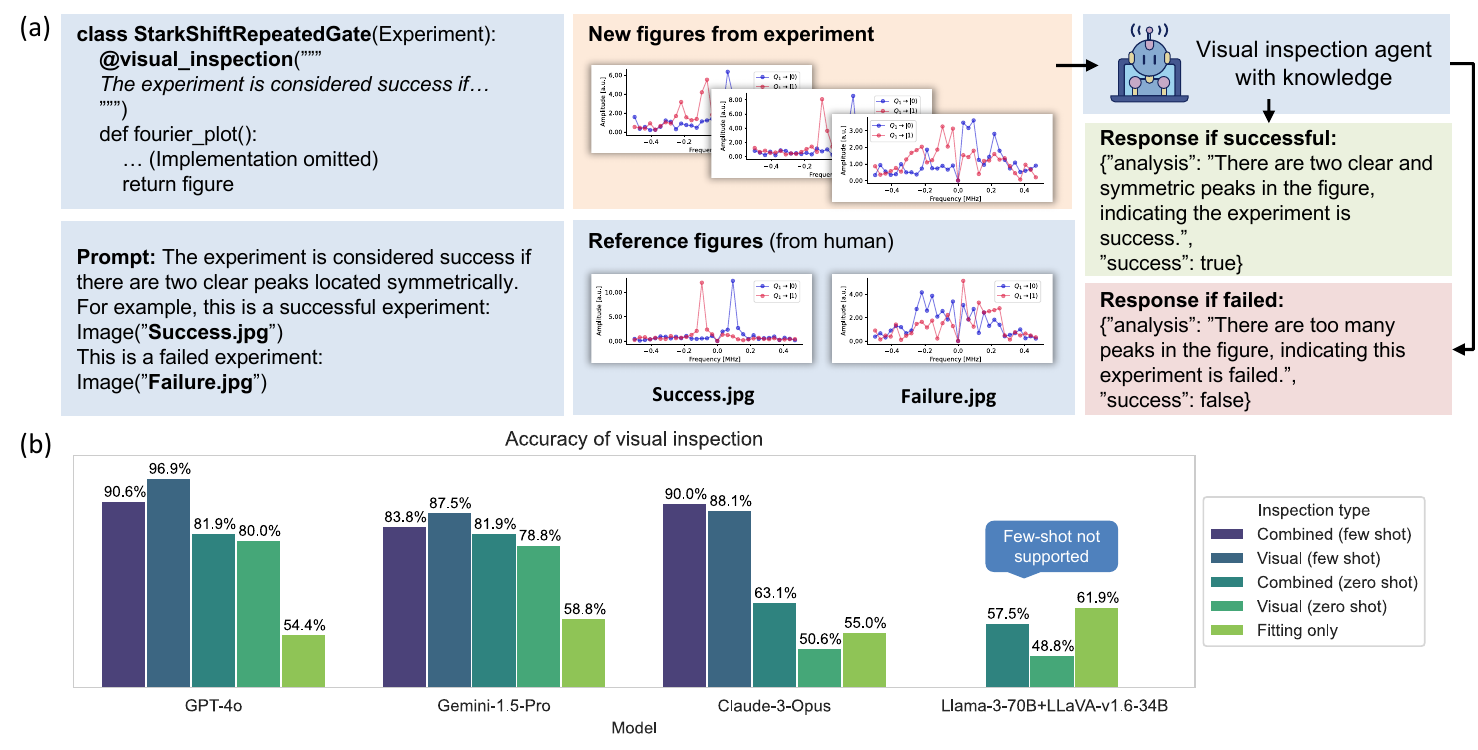}
    \caption{
(a) Visual inspection agent. Our framework introduces visual inspection agents by attaching decorators to the member methods of the experiment class that produce figures. The decorator accepts a prompt as input, and users can add example figures to the prompt. Whenever the experiment is carried out, the execution agent can call the inspection agents to generate a text-based report based on the outcome figure and their knowledge (prompts). (b) Performance benchmarking of visual inspection. We compare the performance of the visual inspection method in determining whether a certain experiment is successful or not. We added three types of inspection agents. Visual agents with a text description of success criteria (zero-shot), visual agents with text descriptions and example figures (few-shot), and fitting agents who provide reports based on fitting results. We test the accuracy of inspection on each type of agent. We also combine the results of fitting agents and visual agents in the ``Combined'' setup. We found that providing example figures significantly improves accuracy. Additionally, we found that reports synthesized from multiple inspection agents can give better accuracy in some cases.}
\label{fig-visual}
\end{figure*}

\textbf{(Inspection agent)} 
A major burden for experimentalists has been the need to wait for experimental results and decide on the next step based on them. In many scenarios, experimentalists must analyze figures to assess the success of an experiment and determine the next steps. In the \textit{k-agents} framework, we introduce inspection agents that have the knowledge needed to evaluate the results of each experiment. These agents will be called after the execution of each experiment to analyze its outcomes.
We provide an interface to inject knowledge for each function that produces experiment figures. This feature is implemented as a Python decorator. In the decorator, the users can add instructions for how to analyze the figure. Besides, as illustrated in Figure~\ref{fig-visual}(a), our interface allows users to add example figures to help others understand the new figures.
Based on the knowledge from the interfaces, we construct inspection agents equipped with multimodal LLMs to inspect and analyze new experiment figures and output text-based reports based on their knowledge.

\subsection{Execution agent}

We introduce the execution agent to coordinate knowledge agents and manage complex experiment procedures. This agent takes experiment procedures written in natural language as input, \sx{and driven by a text-only LLM}. Since the selection of the next experiment in a procedure could depend on the result of previous experiments, we design the execution agent to first decompose the procedure into a ``state machin'', in which each state represents a distinct experiment stage of the procedure (See Figure~\ref{fig:framework-overview}(c)). Each stage of the experiment is assigned a single instruction that contains only one experiment or sub-procedure. As \sx{an agent-based state machine}, there is also a transition rule at each stage in natural language that determines how to select the next stage based on the result of the experiment. 

By constructing the \sx{agent-based state machine}, the execution agent transforms the task of executing the procedure into executing the \sx{agent-based state machine}. The execution is illustrated in Figure~\ref{fig:framework-overview}(a). Starting from the initial stage, at each stage, the execution first translates the instruction at the stage into executable code with the assistance of translation agents. To increase the accuracy and efficiency of the translation, the execution agent will activate only translation agents related to the context. The relevance score will be calculated based on the embedding similarity between the context and the description of the translation agents. The execution will select a reasonable response from the translation agents and execute the translated code.
After executing the code, the agent gathers a natural language report from inspection agents and, based on the report and the transition rules, determines the next stage to transition to.
If the reports suggest updating the parameters in the new stage, the execution agent will also pass the update to it. 
This state-machine architecture enables the execution agent to focus on executing each stage of the procedure efficiently without needing to monitor the entire experimental history when deciding on the next experiment.

%% file: 3.results.tex
\section{Results}
\label{sec:result}

In this section, we demonstrate how the \textit{k-agents} framework can be applied to the calibration and operation of a superconducting quantum processor. Superconducting circuits provide a popular physical platform for building quantum information processors. These circuits are fabricated on a chip using nanofabrication techniques.  These circuit components operate under quantum mechanical principles at millikelvin temperatures, forming non-linear quantum resonators which can be excited similarly to atoms using microwave signals. Information can be stored in the quantum states of these resonators, which serve as qubits on the processor \cite{PhysRevA.69.062320,Acharya2023,10.1063/1.5089550,annurev-conmatphys-031119-050605}. To control these qubits, external electronics connected to the chip generate precise microwave signals. These signals must be carefully calibrated to accurately set the operational parameters, which are crucial for executing logical quantum gates. Typically, the parameters include the shape of the pulse (the duration and amplitude at a certain time $t$) and the initial phase and the frequency of the pulse. For a two-qubit gate, multiple pulses may be required to send at the same time to generate the entanglement \cite{PhysRevA.93.060302,PRXQuantum.5.020338,10.1063/1.5089550}.

\begin{figure*}
    \centering
    \includegraphics[width=\linewidth]{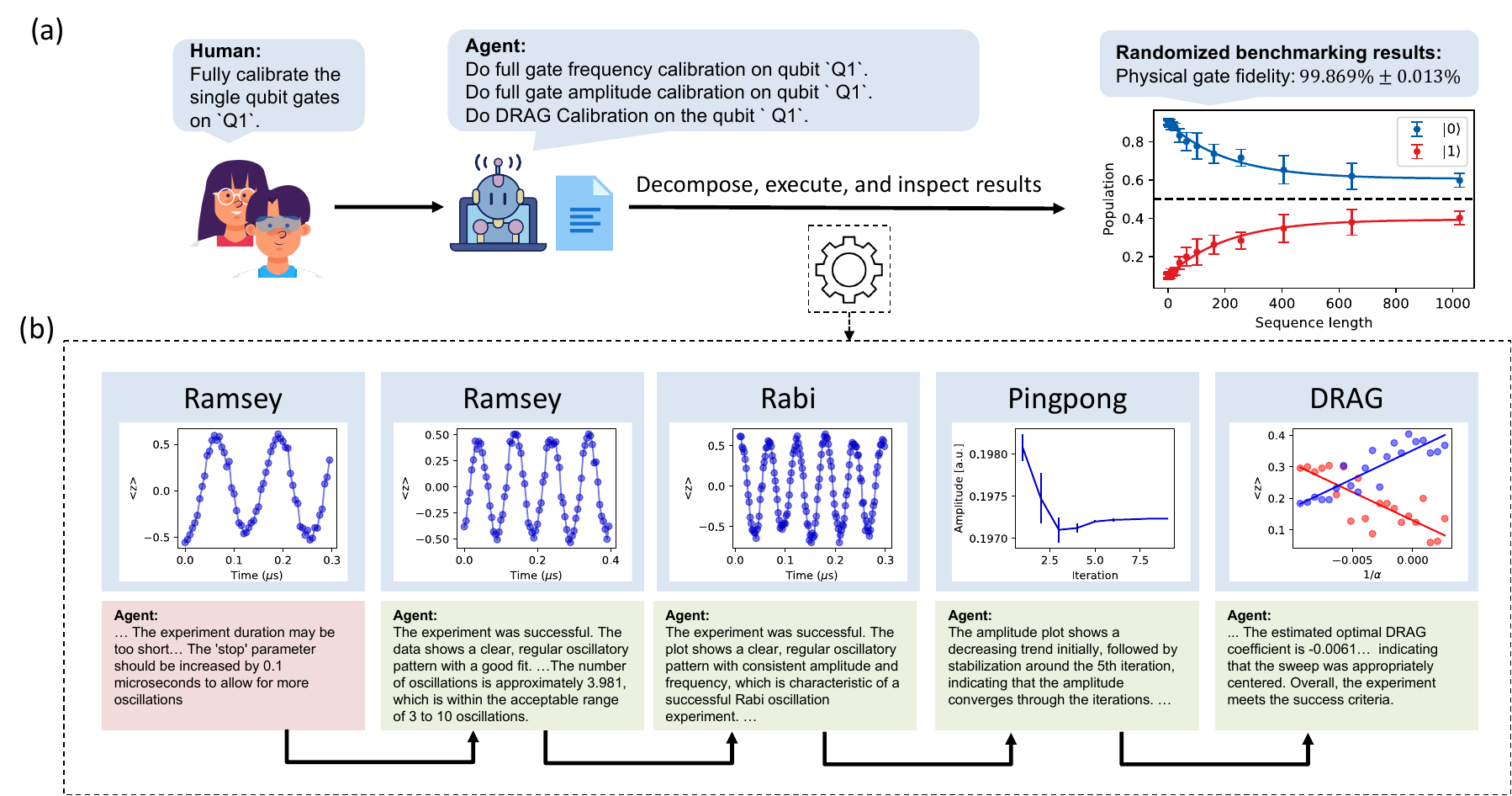}
    \caption{Automated single qubit parameter recalibration automated by \textit{k-agents}. The human scientist instructs the \textit{k-agents} framework for calibrating the single qubit gate parameters. Based on the reference documents provided in natural language, \textit{k-agents} successfully breaks the calibration into four steps, followed by the randomized benchmarking experiment to characterize the single-qubit gate fidelity. In this example, \textit{k-agents} found the first Ramsey experiment had not collected enough oscillations to estimate the frequency. Therefore, it repeats this experiment by increasing the experiment time. The following experiments all passed the success criteria of the \textit{k-agents}. In the end, the \textit{k-agents} implements the randomized benchmarking and reports the gate fidelity.}
    \label{fig:single_qubit_calibration}
\end{figure*}

A significant challenge in superconducting circuit research is identifying these optimal operation parameters. Although automated scripts can determine some parameters, they still require customization by scientists based on different hardware setups. Moreover, programmatically validating the correctness of the result of these scripts is usually difficult to implement. Scientists must continuously monitor experimental status and adjust the calibration scripts to optimize operational parameters. Moreover, these parameters drift over time and occasionally require recalibration. This process is time-consuming and repetitive, often not yielding direct results for physical research, yet it is essential for conducting superconducting circuit research. This calibration process is complex and labour-intensive, leading to significant challenges for scaling quantum computing systems. These difficulties are not limited to the superconducting circuit platform, but also extend to ion traps \cite{maksymov2021optimal,romaszko2020engineering,bruzewicz2019trapped}, spin qubits \cite{botzem2018tuning,burkard2023semiconductor,chatterjee2021semiconductor,stano2022review}, and Rydberg atom systems \cite{wu2021concise,adams2019rydberg,cong2022hardware}. Moreover, the techniques developed for these tasks have broader implications beyond quantum computing, which has great potential to accelerate scientific and technological research in general. 

The above challenges fit the targeting scenario of the \textit{k-agents} framework. We adapt our existing control software stack to the \textit{k-agents} framework, which enables monitoring experimental results, recommending further parameter adjustment, and executing automated calibrations and experiments. To demonstrate its capabilities and assess effectiveness, we conducted three experiments. Before moving on to the hardware demonstration, we first benchmarked the performance of \textit{k-agents} in the context of superconducting qubits. We adopted the \textit{LeeQ} software \cite{LeeQRepo}, originally used to control superconducting quantum processors in the lab, to work with the \textit{k-agents} system and provided it with additional multimodal information to determine the success or failure of the experiments (See Appendix \ref{app:built-in_experiments} for more detail). First, we benchmarked the translation process using different LLM models, comparing the typical RAG method with our approach. Our results show that the GPT-4o model has the best performance and archives an accuracy of $97\%$, which we consider sufficient for practical application (see Figure \ref{fig:interpret}(b)). In practice, when code generation fails, producing unusable code or syntax errors, we prompt the agent to retry, further improving the success rate. We also evaluated the inspection agents to determine if they could accurately identify the success or failure of experiments. We found that the best performance is achieved when using inspection agents constructed with example figures (See Figure \ref{fig-visual}(b)).

\begin{figure*}
    \centering
    \includegraphics[width=\linewidth]{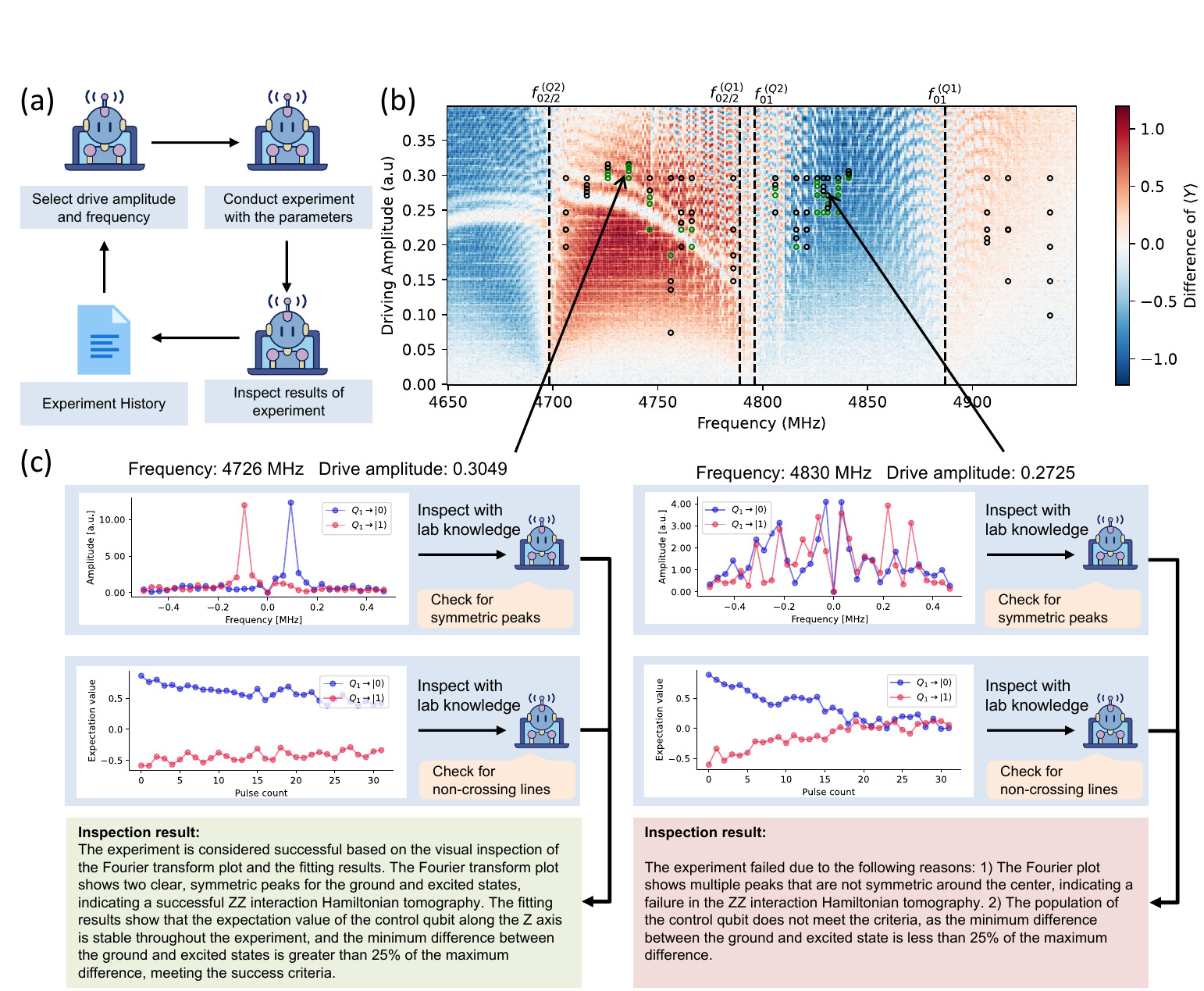}
    \caption{Two-qubit gate parameter discovery driven by \textit{k-agents}. (a) The workflow of the discovery process, where \textit{k-agents} continuously read the history of previous experiments and suggest the next parameter set likely to succeed. (b) The set of discovered parameters and the corresponding parameter search space. The background indicates the ZZ interaction strengths, visualized from data collected during an overnight experiment, which the agent has no prior knowledge of. The green circles denote locations where the agent identified an acceptable parameter set, while the black regions indicate failed attempts. For more details on the search for siZZle gate parameters, refer to Appendix \ref{app:two_qubit_calib}. (c) Examples of figures observed by the agent, along with its response, illustrate a typical success case and a failure case, respectively.}
    \label{fig:two_qubit_calibration}
\end{figure*}

Following the successful benchmarking results, we designed the experiments to further evaluate the performance of \textit{k-agents} on real hardware. First, we demonstrate fully automated recalibration of single-qubit parameters. Second, we showcase  the automated discovery and calibration of two-qubit gates on our platform. Third, we highlighted the AI agent's ability to generate quantum states based on natural language descriptions, including the creation of a GHZ state across three qubits, utilizing previous calibrations.

Our experimental platform is a 16-qubit superconducting quantum processor configured in a four-by-four square lattice connectivity layout, \sx{the characterization details of this device has been reported in \cite{2505.22276}}. The qubits employed are standard fixed frequency transmon qubits with coaxial geometry \cite{Spring2022, Rahamim2017Double-sided, 2501.15059}. For the demonstration of our framework, we selected a subset of \sx{adjacent} 3 qubits. \sx{In superconducting qubit systems, two-qubit gates are usually have to be implemented between adjacent qubits, and the crosstalk and noise are more significant. Our demonstration using an adjacent 3-qubit subset is consistent with current hardware constraints and design practices. Although it involves only a few qubits, this setup reflects the typical conditions of academic labs, where small-scale experiments suffice for exploring quantum phenomena and improving qubit performance.} For further details about the hardware setup, please refer to Appendix \ref{app:Hardwrae}.

In the first experiment, we demonstrate automated calibration and benchmarking of a single qubit with \textit{k-agents}. To facilitate the calibration process, we provided the agent with a document detailing the general procedure for single-qubit calibration in natural language. At the start of the experiment, we manually adjusted the initial parameters to deviate from their optimal values. Based on the document, the execution agent first decomposes the procedure into two distinct stages: calibration and benchmarking. The calibration stage was further broken down into multiple steps, in which the agent sequentially adjusted the qubit's frequency, amplitude, and DRAG parameters. If a failure occurs during the procedure, the step will be re-tried. After several unsuccessful attempts, it reverts to the previous stage to try again. This process is driven by the agent to emulate the typical behavior of a human scientist who implements the calibration process. We observed that the agent is able to perform transition correctly between stages based on the experiment result and find the acceptable parameters. At the benchmarking stage, the agent successfully performed a randomized benchmarking on the single qubit and validated its fidelity (See Figure \ref{fig:single_qubit_calibration}).

\begin{figure*}
    \centering
    \includegraphics[width=0.8\linewidth]{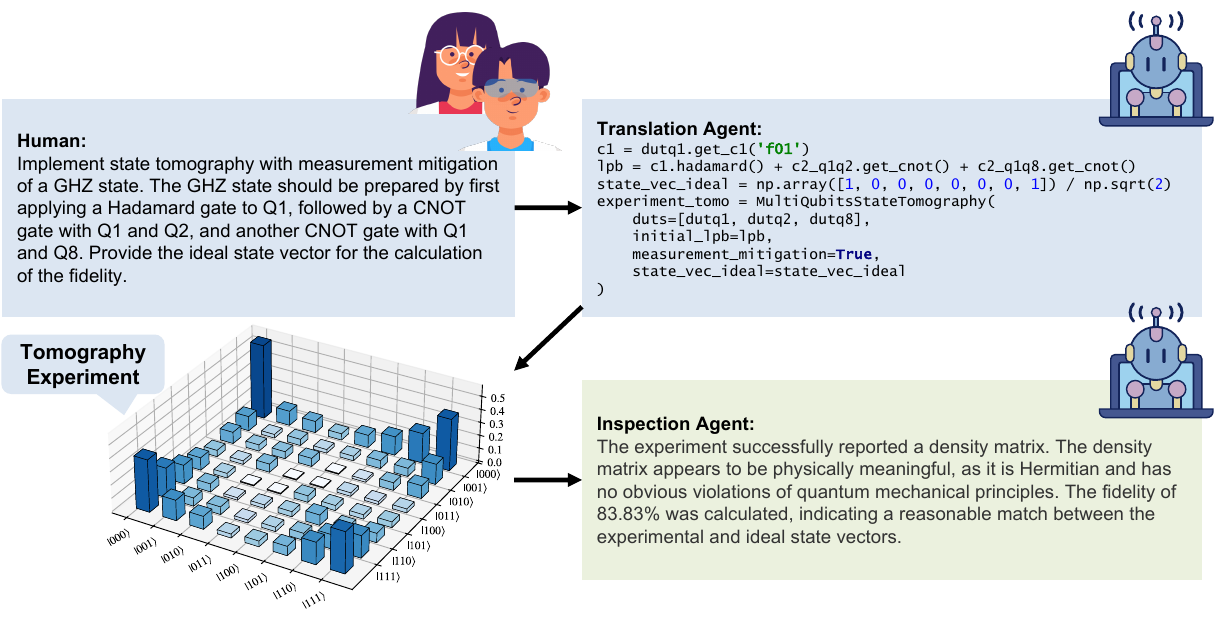}
    \caption{Automation of the GHZ state generation and tomography with \textit{k-agents}. First, the instructions are provided by human scientists. Then, the code is generated by the translation agent to implement the experiment. The density matrix plot of the generated GHZ state is constructed from the experiment result. Finally, a report in natural language is generated by an inspection agent.}
    \label{fig:entanglement-generation}
\end{figure*}

Following the single-qubit calibration experiment, we demonstrated the automated discovery of two-qubit gate parameters. Specifically, we focused on the siZZle  (Stark-induced ZZ by level excursions) gate \cite{PhysRevLett.127.200502, PhysRevResearch.4.023040,wei2021quantum,PhysRevLett.129.060501}, which creates entanglement between two fixed-frequency superconducting qubits. By driving both qubits off-resonance at the same frequency simultaneously, the ZZ term in the qubits’ Hamiltonian is altered compared to when the qubits are undriven. Leveraging this effect, we constructed a pulse sequence with calibrated amplitude and frequency to adjust the ZZ interaction strength, followed by fine-tuning the pulse duration to achieve precise qubit entanglement. The most challenging part of the process was the search for optimal driving amplitude and frequency, while the pulse duration calibration was relatively straightforward, accomplished programmatically by measuring the ZZ interaction rate during the drive. In this experiment, the AI agent autonomously identified a set of working parameters for the siZZle gate.

For a human scientist, searching these parameters typically requires starting with empirical knowledge and measuring the ZZ interaction strength at specific driving frequencies and amplitude. The search process then proceeds by adjusting the frequency and amplitude based on the experimental outcomes. When an experiment succeeds, we often increase the pulse amplitude to enhance the ZZ interaction rate or adjust the driving frequency closer to the qubit’s transition frequency. Conversely, when an experiment fails, we either decrease the amplitude or move the frequency away from the transition frequency. There are three regions that need to be explored: below both qubits' transition frequencies, between the two, and above both qubits' transition frequencies. We inject this empirical knowledge of parameter selection into two LLM-based agents which are accessible to the experiment history for what parameters have been tried. By wraping these agents into two special experiments, the execution agent can call them to get the next frequency and amplitude to try. The parameter searching can then be formulated as an alternate calling to these special experiments and the experiments for testing the proposed parameters.

For each parameter pair (frequency and amplitude), we measured the ZZ interaction rate by performing Hamiltonian tomography on the ZZ term using two methods: continuous-time tomography and repeated gate tomography. The latter method, which accounts for the effects of the pulse's rising and falling edges, yielded more accurate ZZ interaction measurements per pulse.

The most challenging aspect was determining the success of each experiment. To address this, we plotted the Fourier transform of the Hamiltonian tomography results and let an inspection agent decide whether the experiment was successful, where the agent is equipped with a few-shot visual knowledge about the success criteria. Furthermore, we monitored the state of the control qubit, plotting its expectation value of $\braket{Z}$, and tasked another inspection agent with ensuring that it remained stable during the driving process, avoiding any unwanted excitation.

We ran the experiment for 3 hours, limiting the execution agent to carry out 100 experiments to search for parameters and test a maximum of 20 frequencies. By the end of the experiment (See Figure~\ref{fig:two_qubit_calibration}), the agent successfully identified an optimal set of parameters (frequency 4726MHz and amplitude 0.3049).

Finally, we demonstrate the ability of the AI agent to carry out experiments based on natural language instructions. After successfully calibrating the gates, we instructed the execution agent to perform state tomography for a GHZ state. Additionally, we requested the agent to perform process tomography for the two-qubit gate parameters it had calibrated. The agent successfully implemented this experiment and reported the state fidelity (See Figure \ref{fig:entanglement-generation}).

%% file: 4.conclusion.tex
\section{Discussion}

In this study, we introduce the \textit{k-agents} framework for creating laboratory-related knowledge agents and demonstrate its capability to automate experiments. We applied the system in a real-world setting, successfully demonstrating the ability of \textit{k-agents} by performing single-qubit and two-qubit gate calibrations, as well as generating and characterizing a GHZ state on a quantum processor based on human instructions. The \textit{k-agents} succeeded in orchestrating the experiments, interpreting the multimodal experimental results, and guiding the execution of the experiment with closed-loop feedback. These achievements suggest that our system is a valuable tool for research groups working with superconducting quantum processors, with the potential for broader applications in research automation across other fields.
Additionally, we also applied two benchmark experiments on \textit{k-agents}. We found our agent-based translation of natural language instructions outperforms standard RAG approaches in terms of accuracy. By our benchmark of the inspection agents, we showed that current multimodal LLMs have the ability to analyze experiment figures based on human instructions, especially when provided with example figures. 

\sx{
While methods such as rule-based systems, reinforcement learning, and Bayesian optimization have been used for calibrating quantum devices, our approach is fundamentally different. First the \textit{k-agents} framework is designed to automate entire experimental procedures, rather than focusing on the tuning of specific parameters. Optimization methods like reinforcement learning and Bayesian optimization typically require extensive manual tuning, carefully crafted reward functions, and are limited to narrow, predefined tasks. In contrast, our framework provides a more general, interpretable, and adaptable solution, particularly well-suited for dynamic and evolving quantum laboratory environments. Moreover, unlike traditional rule-based systems that depend on rigid, hand-coded logic, the \textit{k-agents} framework uses multimodal prompts (natural language and images) and LLMs to enable more flexible and efficient automation.
}

\sx{One of the strengths of the \textit{k-agents} framework is its approach to knowledge representation. It requires no more effort than maintaining standard laboratory documentation and fits naturally into routine protocol management. While we acknowledge that adopting any new framework involves some onboarding and training, the setup effort is comparable to preparing laboratory manuals or protocols for new PhD students or junior staff. Once standard documents are already exist—which is common practice in most laboratories—our agents can be configured with minimal additional work. }

\sx{While our framework shares conceptual similarities with prior agent systems like ChemCrow~\cite{m2024augmenting} and Voyager~\cite{wang2023voyager}, it introduces several important distinctions that enhance agent-based AI for laboratory automation. First, our framework is specifically tailored for quantum computing laboratories, where experiment automation involves breaking down long-horizon workflows and dynamically adjusting parameterized experiments by analyzing previous results. In contrast, ChemCrow targets open-ended challenges in general chemistry, focusing on relatively straightforward, short-horizon tasks that require less iterative reasoning and fewer re-executions. Second, unlike Voyager, which utilizes large internet-scale corpora about Minecraft that has already trained into large language models, our system operates effectively without depending on such extensive external data, making it better suited for laboratory environments. Additionally, Voyager develops agent skills through simulated environments—a resource not always available in laboratory settings—whereas our \textit{k-agents} framework emphasizes practical application by building on existing laboratory codebases. These distinctions reflect our commitment to improving the practicality and scalability of agent-based AI for real-world scientific experimentation.
}

However, the \textit{k-agents} framework has some potential weaknesses that make full automation challenging. First, our method relies on clean and well-formatted codebases and procedure documents. However, such structured knowledge may not always exist, and transforming existing knowledge could take considerable human effort. Future research on structuralizing unstructured documents and codebases is still needed to completely automate knowledge translation. A possible direction of research could involve enabling the framework to generate structured code and procedures based on unstructured project files and human inputs.
Secondly, using LLM, the execution of \textit{k-agents} exhibits higher latency than traditional automation methods that do not require LLM-based planning and translation of knowledge at runtime. We note that this problem can be well mitigated by caching the response of LLMs and using the same response whenever the input is the same.

\sx{In addition, our current implementation of execution agents within the \sx{agent-based state machine} does not support fine-grained, interruptible execution; it only allows for termination of the entire program in response to external signals or emergency conditions. However, we acknowledge that in quantum systems, it would be beneficial to provide an interactive mechanism for human scientists to monitor and intervene in the experiment's progress. Addressing this limitation is a priority for future development, and we plan to explore the integration of real-time interrupt mechanisms, hardware-level hooks, and human-in-the-loop safety protocols.}

We monitor the cost of the \textit{k-agents} system. During the parameter search for the two-qubit gate, we observed that the LLM used 1,373,207 input tokens (including image tokens) and 168,039 output tokens in 3 hours, costing less than 5 US dollars. We believe that there is potential for further optimization to reduce these costs. Regarding time efficiency, despite network delays, LLMs currently take longer than humans to perform inspections. \sx{The inspection agent utilizes multimodal LLMs and requires a few seconds to evaluate each experimental result, with an inference time comparable to that of the standard LLM agents without the multimodal ability.} However, LLMs tend to be more efficient than humans in generating code. Therefore, overall efficiency is comparable.

The use of superconducting quantum processors simplifies some challenges, such as safety control and error tolerance. In our system, the risk of the agent executing code that causes significant damage or unacceptable costs is very low. If it fails, we can simply allow it to try again. However, this may not be the case in other applications. Prior research \cite{tom2024self, wattoo2024rabit, steering2024} has focused heavily on safety control, and this is an area we plan to explore further in the future.

%% file: 2.methodology.tex
\section{Method}

\subsection{Details on the knowledge agents developed}

\subsubsection{Definition of knowledge agents}

The concept of knowledge agents is defined based on the work proposed by Zhang et al. \cite{zhang2024theory}, in which an agent's learning ability is defined as its ability to internalize knowledge given related inputs. A knowledge agent can then be interpreted as an agent with the ability to learn (i.e., acquire knowledge) and to apply the learned knowledge to respond to other agents (i.e. transfer knowledge). A bare LLM can be regarded as a knowledge agent as it is able to absorb knowledge from its training data. However, it is difficult for LLMs to internalize laboratory documents and codes as their knowledge because of the current pitfalls in model editing \cite{li2024unveiling}. 
Therefore, we regard bare LLMs as weak knowledge agents compared to the knowledge agents we proposed, which are good at accepting knowledge from laboratory documentation.

\subsubsection{Translation Agents} 

In \textit{k-agents}, we included two types of translation agents: \textit{code translation agent}, which translates instructions into code, and \textit{procedure translation agent}, which translates instructions into experiment procedures that may contain multiple institutions. Each translation agent is designed to only process instructions that are very close to the description of an existing experiment. However, by recursively translating the instructions, the whole system can achieve higher generalizability.

\textbf{(Initialization)}  
Each code translation agent is directly constructed from a class of experiments. Similarly, procedure translation agents are also constructed from the structured documents of procedures mentioned above. The document translation agent will try to translate the instructions that resemble the title of the experiment item into code that invokes an execution agent to execute the corresponding procedure.

\textbf{(Activation)} 
Many translation agents might exist based on the complexity of the experimental system. However, not all translation agents are needed to translate a certain instruction. Therefore, we designed to use embedding similarity and only activate the agents related to the instruction to be translated. When generating the agent, we use LLMs to generate a series of natural language instructions $\{I\}$ that the agent should translate. We further calculate the embeddings $\{\vec{E}\}$ of these instructions. 
When there is an incoming instruction with embedding $\vec{E_I}$, the score $S$ of a translation agent is calculated by the maximal inner product between $E_I$ and the vectors in $\{\vec{E}\}$.
\begin{equation}
\label{equ:similarity}
  S = \max_{\vec{E_i}\in \{\vec{E}\}} \vec{E_I} \cdot \vec{E_i}^{\dagger}  
\end{equation}

\textbf{(Code translation)} When activated, the code translation agent attempts to translate the instruction into executable code. The Python class signature for the experiment and the instruction will be passed to an LLM to generate a code translation based on the class. We employ two strategies to improve the translation accuracy throughout this process. First, we implement the chain of thought (CoT) \cite{wei2022chain} strategy to improve performance. Before generating code, the LLM is first asked to provide a paragraph of analysis on how to approach the translation. This helps guide the LLM in retrieving key points from the context and making useful inferences. Next, we adopt a self-reflection strategy to improve the accuracy. The agent will determine whether the instruction should be translated using the experiment class based on the previous analysis. This step selects out cases where the class is not suitable for the instruction, helping to avoid hallucinations. If the agent concludes that the instruction should be translated into code using the class, it proceeds to generate a candidate code snippet. This candidate code is then sent to the execution agent for comparison with other potential candidates.

\textbf{(Procedure translation)}
Our procedure translation agents are designed similarly to code translation agents. If an agent deems an incoming instruction matches the procedure it holds, it will try to rewrite the procedure title to suit the instruction. A code snippet that calls an execution agent to execute the rewritten title will then be generated and sent to the execution agent as a code candidate.

\textbf{(Generalizability)}
In both types of translations, the agents generate new code and new instructions based on the input instruction and therefore generalize the knowledge they have. However, this generalizability is well-controlled as translations that deviate too much from the agent's knowledge will be decided as improper by the strategy we described. This limited generalizability mitigates the negative effect of the LLM's knowledge from the public corpus that is not suitable for a private experimental setup.

\subsection{Details on execution agents}

In \textit{k-agents}, the execution of experiments is controlled by the execution agents. The execution process involves decomposing the incoming procedure into experiment stages, translating the instruction at each stage into code by translation agents, executing the generated code, and analyzing the experiment results to determine the next step. 

\textbf{(Instruction decomposition)} When handling experiment procedures, we ask the execution agent to decompose the instructions into experimental stages and make \sx{an agent-based state machine}. Each stage in the \sx{agent-based state machine} contains a single step of the instructions that only includes one experiment or sub-procedure to be translated by the translation agents. In addition, an experiment procedure might have a complex control flow. For example, selecting the next instruction to execute might depend on the result of the previous instructions. Therefore, at each stage, we also attach a prompt on how to select the next stage given the result of this stage.

The stage generation is done roughly in two calls to an LLM. First, we employ the LLM to extract a list of independent experiments from the procedure. This step removes all the sentences controlling the procedure's progression and generates the instructions at each stage. Then, we add indices to the stages, such as {\sf Stage 1 and Stage 2}, so their indices can be used to refer to them when generating the transition rules. We also add two special stages: {\sf COMPLETE} and {\sf FAILED} to the list of stages. Finally, we provide the LLM with the list of stages and the original procedure, asking it to attach the transition rule for each experiment.

\textbf{(\sx{Translation candidate selection})}
\begin{algorithm}
\caption{Translation with self-reflection}
\label{alg:trans}
\begin{enumerate}
    \item Set the group of translation agents $\{A\}$, the number $N_k$ for the number of activated agents and the number $N_{\max}$ for the maximal number of activated agents.
    \item Input the instruction $I$ to be translated.
    \item Calculate the score of each agent in $\{A\}$ by Equation~\ref{equ:similarity}.
    \item For each agent $A$ whose score is ranked top $N_k$ in $\{A\}$, do the following
    \begin{enumerate}
        \item Using LLM, check whether the instruction can be translated by the knowledge held by $A$. If so, generate the code $C$ that implements the instruction. Else, output nothing.
        \item Add $C$ to the set of code candidates.
    \end{enumerate}
    \item If the set of code candidates $\{C\}$ is empty, increase the number of $N_k$ by $2$ and do the following: If $N_k<N_{\max}$, go back to step 2. Else, the algorithm fails.
    \item If $\{C\}$ is non-empty, add the instruction and the set of code candidates $\{C\}$ to the context of LLM and generate the code $C^*$ for $I$.
    \item Output $C^*$ as the result of translation.
\end{enumerate}
\end{algorithm}
When executing a stage, the execution agent activates the translation agents and gathers candidate code from them. Then, the execution agent synthesizes the code to execute on the basis of the candidates it receives.  
\sx{Specifically, all the candidate code will be input into a LLM with a prompt that ask the LLM to generate an analysis and the final code for the translation.}
This synthesis process has two merits. First, there might be two experiments with similar descriptions. The translation agents for them are likely to generate candidate code together. By providing them as candidates, we make it possible for LLM to analyze which solution is more suitable. Further, as the activating mechanism might not be perfect, there might not be any answer from the translation agents because the desired translation agent is not activated. In this case, we design the execution agent to try to activate more translation agents until a proper candidate appears. We summarize the translation process in Algorithm~\ref{alg:trans}. When applying the algorithm in Section~\ref{sec:result}, we set the $N_k=3$ and $N_{\max}=9$.

\textbf{(Execution)}
Starting from the first stage, the execution agent executes the instruction in the current stage by executing the translated code with the help of the translation agents, as described in the previous section. After execution, the inspection agent generates a report of the experiment in natural language. Based on the report, the execution agent chooses the next state to transition to.
Finally, after reaching the {\sf COMPLETE} or the {\sf FAILED} state, the execution agent will generate a report of the entire execution. \sx{We summarize the execution process in Algorithm~\ref{alg:exec}.} 

We note that the execution agent also maintains a variable table. At the end of each experiment, variables might be injected from the experiment for the use of other experiments. The injection of variables can be programmed by the users when they implement the class for the experiment.

\begin{algorithm}
\caption{Execution agent workflow}
\label{alg:exec}
\begin{enumerate}
    \item Input the experiment procedure $P$.
    \item Using LLM, decompose $P$ into experiment stages (states) $\{S\}$ and transition rules $\{T\}$ at each stage. The decomposition also specifies an initial state $S_0$ as the current state $S^*$. Two final states {\sf COMPLETE} and {\sf FAILED} are also in the $\{S\}$.
    \item At the current experiment stage $S^*$, do the followings
    \begin{enumerate}
        \item Translate the instruction contained in $S^*$ into code with the Algorithm~\ref{alg:trans}.
        \item Execute the translated code.
        \item Activate available inspection agents in the executed experiment and summarize the results from them with LLM, giving a summarized report $R$. Add $S^*$ and $R$ to the experiment history
        \item Using LLM, decide the next stage to transition to, based on the transition rule $T^*$ of $S^*$ and the report $R$.
        \item If the next stage is {\sf COMPLETE} or {\sf FAILED}, add them to the experiment history and break the loop (i.e., go to step 4). Else, set $S^*$ to be the next stage and go to step 3(a).
    \end{enumerate}
    \item Using LLM, summarize the experiment history and produce a report $R$ for $P$.
\end{enumerate}
\end{algorithm}

\subsection{Quantum processor hardware}

The quantum processor used in this experiment consists of 16 coaxmon qubits arranged in a square lattice configuration. Coaxmon qubits, a variant of transmon qubits with coaxial geometry and off-chip wiring, have demonstrated high coherence, low cross-talk, and stable higher excited states~\cite{Spring2022, Rahamim2017Double-sided, PhysRevLett.133.120802, Cao2024}.

To enable the execution agents to conduct experiments, we integrated the \textit{k-agents} framework with the \textit{LeeQ} framework \cite{LeeQRepo}, customized software used for controlling superconducting quantum processors in the lab. The \textit{LeeQ} framework, which has been independently used by human scientists, manages the composition, compilation, and optimization of quantum gates into executable instructions. These processed microwave sequences are then sent to the QubiC system \cite{2309.10333}, an electronic system that generates and processes microwave signals and connects to the quantum processor.

In practice, human scientists write code that predefines experiments and reuse it in different scenarios by modifying the arguments of these experiments. Typically, scientists manually adjust the parameters of these experiments until they achieve their desired outcome. We adapted these predefined experiments to be compatible with and indexable by \textit{k-agents}. In total, the \textit{LeeQ} framework includes more than 40 built-in experiments, of which we selected 17 for indexing by the \textit{k-agents}. For further details on the experiments, please refer to Appendix \ref{app:built-in_experiments}.

%% file: appendix.tex
\onecolumngrid

\newtcolorbox{demobox}[2]{breakable,colback=#2!5!white,colframe=#2!75!black,fonttitle=\bfseries,title=#1}
\lstdefinestyle{base}{
  breaklines=true,
  basicstyle=\fontsize{7}{7}\selectfont\ttfamily
}
\definecolor{teal}{RGB}{50, 220, 220} 

\newtcolorbox{conversationbox1}[1][]{breakable,colframe=teal!5!black,
  colback=teal!75!white, width=\textwidth, fonttitle=\bfseries, title=#1}

\newtcolorbox{conversationbox}[1][]{breakable,colback=teal!5!white,colframe=teal!75!black,fonttitle=\bfseries,title=#1}
\lstdefinestyle{base}{
  breaklines=true,
  basicstyle=\fontsize{7}{7}\selectfont\ttfamily
}

\newcommand{\cvline}[2]{
  \textbf{#1}: #2
}

\newcommand{\dashedline}{
  \begin{center}
    \tikz \draw[dashed] (0,0) -- (\linewidth,0);
  \end{center}
}

\input{app0_knowledge_agents_details}

\input{app1_knowledge_agents_implementations}
\input{app4_benchmarking}
\input{app2_single_qubit_experiments}
\input{app3_two_qubit_experiment}

\input{app5_device_setup}

%% file: app0_knowledge_agents_details.tex
\section{Details on knowledge representation}
\subsection{Representing knowledge of available experiments}
 
In \textit{k-agents}, laboratory experiments (and actions) are formulated as sub-classes of an abstract class {\sf Experiment}. The abstract class provides a few interfaces to put the documentation and code of the experiment. We use the following example to demonstrate how the interfaces are used.

\begin{demobox}{An example of experiment class}{teal}
\begin{lstlisting}[style=base,language=Python]
class SimpleRamseyMultilevel(Experiment):

    _experiment_result_analysis_instructions = """
    The Ramsey experiment is a quantum mechanics experiment that involves the measurement of oscillations in the quantum state of a qubit. Typically, a successful Ramsey experiment will show a clear, regular oscillatory pattern with an amplitude greater than 0.2. If less than approximately 3 oscillations are observed, the experiment requires an increase in the time of the experiment. If more than 10 oscillations are observed, the experiment requires a decrease in the time of the experiment. The frequency of the oscillations should be less than the expected offset value.
    """
    
    def run(self,
            qubit,
            collection_name: str = 'f01',
            mprim_index: int = 0,
            initial_lpb: Optional[Any] = None,
            start: float = 0.0,
            stop: float = 1.0,
            step: float = 0.005,
            set_offset: float = 10.0,
            update: bool = True) -> None:
        """
        Ramsey experiment for estimating the qubit frequency or T2 Ramsey.
        Parameters:
            qubit: The qubit on which the experiment is performed.
            collection_name: The name of the frequency collection (e.g., 'f01').
            mprim_index: The index of the measurement primitive.
            initial_lpb: Initial set of commands, if any.
            start: The start time for the sweep.
            stop: The stop time for the sweep.
            step: The step size for the frequency sweep.
            set_offset: The frequency offset.
            update: Whether to update parameters after the experiment.
        """
        ...

    @visual_inspection("""
        Here is a plot of data from a quantum mechanics experiment involving Ramsey oscillations. Can you analyze whether this plot shows a successful experiment or a failed one? Please consider the following aspects in your analysis:
        1. Clarity of Oscillation: Describe if the data points show a clear, regular oscillatory pattern.
        2. Fit Quality: Evaluate whether the fit line closely follows the data points throughout the plot.
        3. Data Spread: Assess if the data points are tightly clustered around the fit line or if they are widely dispersed.
        4. Amplitude and Frequency: Note any inconsistencies in the amplitude and frequency of the oscillations. For a perfect experiment, the amplitude should be around 1, and the frequency should be close to the expected value. If the amplitude is smaller than 0.2, the experiment is likely considered failed.
        5. Overall Pattern: Provide a general assessment of the plot based on the typical characteristics of successful Ramsey oscillation experiments.
        Here is an example of a successful experiment:
        Image("sucecss.jpg")
        Here is an example of a failed experiment:
        Image("fail.jpg")
        """)
    def plot(self):
        ...

    @text_inspection
    def fitting(self):
        ... 
        return {
            "success": True,
            "analysis": "The experiment is successful based on the fitting results."
        }
\end{lstlisting}
\end{demobox}

The {\sf run} method of the class holds the lower-level code that carries out the experiment. The docstring of this method holds the description of the experiment and the required arguments. This information will be used when the corresponding code translation agent tries to interpret instructions into code with this class.

The results of many experiments are more intuitive if they are represented in figures. However, interpreting these figures requires domain-specific knowledge, which is beyond the scope of most multimodal LLMs. For this, we provide a function decorator {\sf visual\_inspection} to represent the knowledge of how to analyze the figure. 
The decorator accepts a prompt as a parameter and uses it to generate a visual inspection agent that keeps the prompt as its knowledge. When the experiment is finished, and the experiment results are available in the experiment object, visual inspection agents will be called with the newly produced figures. The visual inspection agent will put the figure into a multimodal LLM along with the prompt to produce the report. We additionally note that our decorator accepts few-shot prompting. All the patterns {\sf Image("*")} in the parameter, where {\sf *} is a path to an image, will be automatically detected and replaced by the corresponding image located at the path.

Besides visual methods to analyze experiment results, we also provide an interface to produce inspection agents in a more general form. For example, an experiment's result can also be analyzed by numerical methods such as fittings. In these cases, the \textit{k-agents} framework encourages the users to implement methods that produce reports in texts for the reference of execution agents.
The users can use the decorator {\sf text\_inspection} to mark the methods that produce a report for inspecting experiment results. Similarly, the decorator also produces inspection agents that will be called when the experiment is finished.

There might exist many inspection agents in an experiment class, and the reports they produce may be too long for the execution agent to use directly. The class variable {\sf \_experiment\_result\_analysis\_instructions} provides an interface to inject the knowledge of how to summarize the reports from the inspection agents. Execution agents use this knowledge to summarize the reports and decide the next stage to transition to.

\subsection{Representing knowledge of experiment procedures}

The knowledge of a laboratory can also exist in experiment procedures, which describe how to decompose an instruction into a procedure that may contain many dependent instructions. To help our agents extract this knowledge more efficiently, we propose the following structured format for them.

\begin{demobox}{An example of structured experiment procedure}{teal}
\begin{lstlisting}[style=base]
# Complete Calibrating Single Qubit `dut`

## Background

The process of recalibrating a single qubit is crucial to maintaining optimal quantum computation performance. This recalibration process involves three key steps: frequency calibration, amplitude calibration, and DRAG parameter calibration. Each step is aimed at refining different aspects of the qubit's operational parameters to ensure accurate and reliable qubit control. The sequence of these steps is critical, as each builds on the stability achieved in the previous step.

## Steps

- Do frequency Calibration on `dut`
- Do amplitude Calibration on `dut`
- Do DRAG Calibration on `dut`

## Results

- Whether the experiments are successful
\end{lstlisting}
\end{demobox}

In our format, each experiment procedure is organized in the Markdown format. Each procedure starts from a first-level header, which starts with {\sf \#} and the title of the experiment. Then, our format accepts three sub-sections: {\sf Background}, {\sf Steps} and {\sf Results}, in which the {\sf Steps} section is necessary. Each sub-section should start from a second-level header as shown in the example.

The {\sf Background} section is provided to place important background knowledge of the procedure. It can be included in the prompts of the agents to clarify the usage of the procedure.

A procedure translation agent who carries the above information will try to match instructions that are similar to the title of the procedure and modify the procedure in the {\sf Steps} section to implement the incoming instruction. 

Finally, knowledge in the {\sf Results} section is used to summarize the results of the execution of the procedure and generate a report for the reference of execution agents.

%% file: app1_knowledge_agents_implementations.tex
\section{Details on implementing agents
 \label{app:agents}}

We discuss the details of the implementation of the agents in the \textit{k-agents} framework. We especially select important prompts that are used to implement the key functionalities of the agents. We omit the inspection agents as their structure is relatively simple, as we have discussed in previous sections.

\subsection{Code translation agent}

When constructing a code translation agent from an experiment class, a few instructions that match the class will be generated with the following prompt.

\begin{demobox}{Prompt for generating possible instructions}{teal}
\begin{lstlisting}[style=base]
You are trying to produce imperative sentences that would invoke the execution of action `{exp_name}` based on its documentation.
<docstring>
{doc_string}
</docstring>
<example>
Here are a few of examples of imperative sentences:
- Run the calibration experiment with duts=`duts` and start=0.0
- Carry out a calibration on `duts` 
- Please execute the Ramsey experiment
- Do the Drag experiment.
</example>
<instruction>
You should output a JSON dict. The keys should be string of indices of the sentences and the values should be the sentences. 
Each sentence should be complete and independent. Name of the experiment should be transformed to natural language and be mentioned.
The sentences should be imperative and should be based on the documentation.
You should output 4 sentences.
</instruction>
\end{lstlisting}
\end{demobox}

The instructions will be converted to a list of feature vectors (embeddings) that represent their semantic meaning using an embedding model. These vectors will be used to calculate the matching score of the experiment given a set of instructions to be translated. The score is calculated as the maximum value of the inner product between the embeddings of the instructions to be translated and the generated instructions. 

After being activated, the code translation agent will try to generate candidate code that implements the instruction. The prompt for generation is listed below.

\begin{demobox}{Prompt for generating candidate code}{teal}
\begin{lstlisting}[style=base]
You are trying to call an experiment to fill the code_to_complete in Python. The description of the task is written in the slot.
<experiment>
{self.get_exp_description()}
</experiment>
<code_to_complete>
# [slot: {instruction}]
</code_to_complete>
<available_variables>
{available_variables}
</available_variables>
<requirements>
You should output a JSON dict. The keys should be
- "experiment_name_in_slot" (string): The name of the experiment extracted from the slot.
- "analysis" : The brief analysis of the relation between the experiment and the code_to_complete. You should notice that the code_to_complete might be irrelevant to the experiment. You should be careful not to assume additional information. The experiment should considered irrelevant if it contains extra keywords or irrelevant information.
- "applicable": A boolean whether the experiment you hold is suitable for implementing the task. 
- "code": A code snippet that is helpful for filling the slot. The last line of the snippet must be in the format: `experiment_<name> = {self.exp_cls.__name__}(argument1,argument2, ...)`. No import statements are needed.
- "explanation": A detailed explanation of what the code snippet does based solely on the documentation.
- "suitable": A boolean whether the code snippet matches the task based on the documentation.
</requirements>
\end{lstlisting}
\end{demobox}

\subsection{Procedure translation agent}
\label{appendix:procedure}
When initializing procedure translation agents, similar to code translation agents, a few variants of the procedure's title will be generated for their activation.

When activated, a procedure translation agent first checks whether the given instruction to translate matches the procedure title it holds. Then, the agent tries to rewrite the given instruction into a format that is closer to the procedure's title it knows. In the meantime, as many experiment procedures require inputs, the agent also generates a map that maps the available variables in the context to the required inputs of the procedure.

\begin{demobox}{Prompt for rewriting instructions based on knowledge of an experiment procedure}{teal}
\begin{lstlisting}[style=base]
You are trying to rewrite the following instruction based on your knowledge to suit more general parameters.
<input_instruction>
{instruction}
</input_instruction>
The following variables names are defined in the context and you can directly use the variables names.
<available_variables>
{available_variables}
</available_variables>
By your knowledge, there is an existing instruction that is available
<knowledge>
<instruction>
{procedure_title}
</instruction>
</knowledge>
<requirements>
You are required to output a JSON dict with the following keys
- "parameter_specification" (string): The specification of the parameters of the input_instruction 
- "analysis" (string): An analysis of the relation between the input_instruction and your knowledge. You should notice that your knowledge is likely to be improper if the experiment name contains different keywords. However, the parameter_specification of the experiment can be different.
- "proper" (bool): Whether the input_instruction can be regarded as a proper instance of the experiment in your knowledge.
- "rewritten_instruction" (string): The input_instruction rewritten in a way based on the instruction in your knowledge. You should not change the parameter specification of the input_instruction.
- "parameter_mapping" (dict): A mapping from the parameters in the input_instruction to the parameters in the rewrote_instruction. The keys should be the parameters in the rewrote_instruction and the values should be the parameters in the input_instruction or the value from the description.
- "annotation" (string): A concise annotation that describe how you are confident that the rewritten_instruction is correct.
</requirements>
\end{lstlisting}
\end{demobox}

After the rewriting, the procedure translation agent wraps the rewritten instruction into a new experiment that runs the instruction with a new execution agent. The code that calls the new experiment is sent to the execution agent as a code candidate.

\subsection{Execution agent (Stage generation)}

Given an experiment procedure, an execution agent will decompose it into stages. Specifically, the decomposition is implemented by a pipeline of LLM calls. The core step in the pipeline is to extract the stages with all the other information dropped. 

\begin{demobox}{Prompt for stage extraction}{teal}

\begin{lstlisting}[style=base]
<experiment_description>
{description}
</experiment_description>
<objective>
Your objective is to decompose the experiment description into standalone instruction.
Each instruction should include an experiment. 
The instructions should make a minimal modification to the original description.
You should not make any inferences or interpret the description. 
You are encouraged to copy the description as is.
You should output as few instructions as possible. You must not expand the instructions.
The instructions must not contain any information about what to do next after the instruction, such as a change of parameter and go to fail.
</requirements>
<example>
For example, if a piece of description is:
"Run experiment A with frequency=10. If failed, retry 3 times."
You should change it into:
"Run experiment A with frequency=10."
<output_format>
You are required to output a JSON dict with a single key "instructions", which contains a list of instructions. Each instruction should be represented as a string.
</output_format>
\end{lstlisting}

\end{demobox}

After the extraction, the generated stages are input to LLM again to generate the corresponding transition rules.
We also add additional steps to extract the numbers in the instructions and save them as a key-value map in each stage.
This allows the execution agent to change these numbers in a state transition when needed. 

\subsection{Execution agent (Stage transition)}

After executing the experiment in a stage, the execution agent decides the next stage to transition to based on the summary of the reports from the inspection agents. The prompt for deciding the next stage includes the summary, the information of the current stage, including how many times it has been executed, and transition rules. \sx{A maximum retry limit has been included in the prompt to prevent the model from entering an endless loop.}

\begin{demobox}{Prompt for stage transition}{teal}

\begin{lstlisting}[style=base]
You are analyzing experiment result from current stage and use the rule of transition to determine the next stage of the experiment.

<current_stage>
{current_stage.label}:{current_stage.description}
The current stage has been executed {current_stage.n_executed} times.
There are {current_stage.n_failed} failed attempts and {current_stage.n_success} successful attempts.
</current_stage>

Here are the results from the experiments. Note that results must be consistent to indicate the validity. 
Otherwise they are both invalid.
<experiment_reports>
{result_prompt}
</experiment_reports>

<rule_of_transition>
{rules}
</rule_of_transition>

<requirements>
Return your decision in JSON format With the following keys:
"analysis" (string): an analysis of the results and the rule of transition to determine the next stage.
"next" (string): the name of the next stage.
</requirements>
\end{lstlisting}

\end{demobox}

After deciding on the next stage, we also include additional steps to modify the variables in the next stage because the inspection agent might propose modifying the parameters. This step is especially important when an instruction needs to be retried with a different set of parameters.

%% file: app4_benchmarking.tex
\section{Benchmarking our methods\label{app:benchmark}}

\subsection{Benchmarking translation}

To test the performance of translating instructions into code, we manually make a test set that consists of 80 pairs of experiments and instructions. \sx{This dataset is for benchmarking the performance and it is not required to use our framework in practice.} For example, for the {\sf NormalisedRabi} experiment, which calibrates the driving frequency of a single qubit, we set the following instructions in the test set.
\begin{demobox}{Examples of the instructions for {\sf NormalisedRabi}}{teal}
\begin{lstlisting}[style=base]
- Run rabi experiment to calibrate single qubit gate driving amplitudes
- Measure Rabi oscillations to determine single qubit gate driving amplitudes
- Implement Rabi experiment to find pi pulse duration
- Calibrate the driving amplitudes for single qubit gate by Rabi
- Determine single qubit gate parameter using Rabi experiment
- Run Rabi experiment with default parameters on the single qubit `dut`
- Single qubit gate amplitudes estimation using Rabi experiments
- Do Rabi experiment to measure single qubit drive amplitudes
- Run Rabi experiment on single qubit `dut` with amp=0.3
- Calibrate single qubit drive amplitudes using Rabi experiment'
\end{lstlisting}
\end{demobox}

In \textit{k-agents}, the translation is done by calling the related translation agents, and the execution agent picks one of the code candidates from the translation agents.
\sx{Specifically, this selection is implemented by inputting all the code candidates, together with the instruction to translate, to an LLM for analysis (See Algorithm \ref{alg:trans}).}
To implement the standard RAG implementation of this process, after retrieving the related translation agents, we extract the experiment classes from them and use the signature and documentation of these classes to generate a translation of the instruction \sx{(See Algorithm \ref{alg:transrag} for detail)}. \sx{The same process of embedding-based retrieving ensures that the increase in performance is caused by our agent-based reflexion and selection process.}

\begin{algorithm}
\caption{\sx{Translation with standard RAG}}
\label{alg:transrag}
\begin{enumerate}
    \item Set the group of translation agents $\{A\}$, the number $N_k$ for the number of activated agents and the number $N_{\max}$ for the maximal number of activated agents.
    \item Input the instruction $I$ to be translated.
    \item Calculate the score of each agent in $\{A\}$ by Equation~\ref{equ:similarity}.
    \item For each agent $A$ whose score is ranked top $N_k$ in $\{A\}$, add the experiment description it carries to  $\{C\}$.
    \item Input the instruction and the experiment descriptions in $\{C\}$ to LLM and generate the code $C^*$ for $I$.
    \item Output $C^*$ as the result of translation.
\end{enumerate}
\end{algorithm}

For a fair comparison, we only included code translation agents in the benchmark as the information as the prompt for code generation with procedures is different from that with experiment classes (See Section~\ref{appendix:procedure}). 
Finally, we include 17 experiment classes in the benchmark. When triggering the translation agents, we select the 2 agents with the highest score. If there is no available code candidate, we increase the number of agents to trigger by 2. Accordingly, we set 2 as the number of experiment classes retrieved in the RAG method being compared. 
\sx{
After the instructions are translated into code, we check the correctness of the code by analyzing its abstract syntax tree.
A search for a function call to the correct experiment will be carried out.
If such a function call is not found, the translation is marked as incorrect.
If a function call to the correct experiment exists, the format of the call will be further checked to ensure the call is valid.
Translations that pass this check will be marked as correct.
We note that we only check the format of the function and do not verify whether the values of the inputs are correct.
}

We additionally note that we use the {\sf text-embedding-3-large} model from OpenAI as the embedding model throughout all the experiments, even when the LLM we use is not from OpenAI. We use the same embedding model to control the variables and compare differences just in the performance of LLMs.

\subsection{Comparison with long-prompt apporach}

\sx{Many large language models (LLMs) now support long inputs, enabling them to process entire codebses. 
However, aside from the significant cost implications, prior studies have shown that providing long inputs can impair LLM performance \cite{liu2023lost, wang2024appbench, yao2024tau}. 
For instance, tool-use performance may degrade substantially when too many tools are included in the prompt.
This implies that systems like ChemCrow\cite{m2024augmenting} and CoScientists\cite{boiko2023autonomous}—both of which present all the available tools to the model at once-may face scalability issues. To address this, and in addition to the benchmark introduced in the previous section, we conduct another benchmark to compare our agent-based translation method with long prompt-based methods.}

\begin{figure}[!h]
    \centering
    \includegraphics[width=0.4\linewidth]{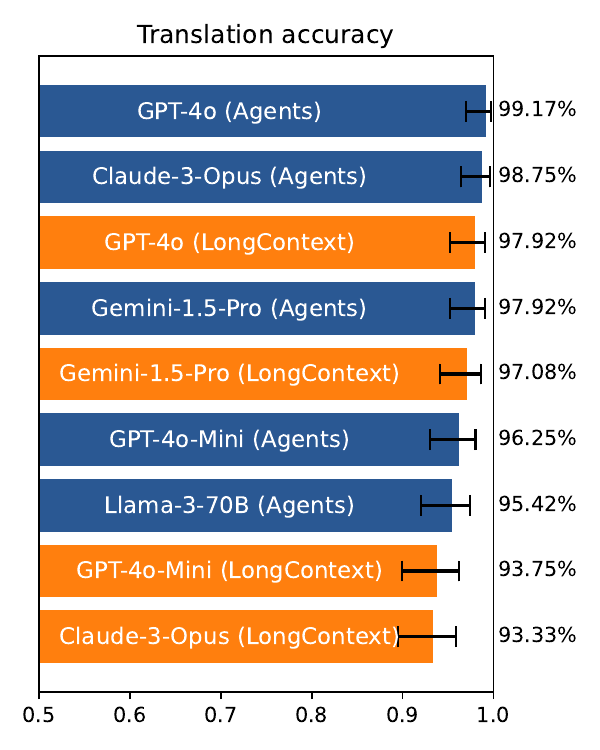}
    \caption{\sx{Comparison of translation success rates using a long context versus our agent-based approach. Long context were found to significantly reduce translation accuracy under certain conditions, such as in the case of Claude 3 Opus, highlighting the advantage of our method.}}
    \label{fig:longcontext-comparison}
\end{figure}

\sx{
In the benchmark, to evaluate the ability of models to handle long inputs, we inserted the descriptions of each of the 17 experiments twice with a random order, resulting in inputs totaling around 11,000 tokens—exceeding the input limits of many LLMs such as GPT-4 (first release) and LLaMA-3-70B. 
For a fair comparison, we set the number of agents recalled during each translation to nine, meaning that nine translation agents were activated whenever an instruction needed to be translated. 
The results, shown in Figure~\ref{fig:longcontext-comparison}, indicate that long-context methods can perform well in many cases; for instance, both GPT-4o and Gemini 1.5 Pro achieved high translation accuracy. However, this high performance is not consistent across all models. For example, Claude 3 Opus exhibited a notable decline in instruction-following ability with longer inputs, leading to poor translation results. 
In contrast, even when only a subset of translation agents is activated, these agents can still outperform their long-context counterparts.
Our result shows that activiating more translation agents can provide better accuracy than inputting all the tools to LLMs.
}

\subsection{Benchmarking visual inspection}

Here, we describe the method for benchmarking the performance of inspection agents to assess their ability to determine the success of an experiment. The agent has three methods for evaluating success: (1) by reading only the fitting report generated by human-written code, (2) by analyzing only the figures generated from the experimental data, or (3) by first analyzing the figures and generating a report, then combining this report with the fitting results to assess success. These methods are referred to as ``Fitting'', ``Visual'', and ``Combined'', respectively.

It is challenging to precisely define success criteria using natural language for the inspection of plots. To address this, we employ a few-shot prompting approach. This involves providing the agent with example images of a successful experiment and a failed experiment, followed by a new image. The agent is asked to determine whether the new image more closely resembles the successful or failed example, with some guidance from natural language descriptions.

The benchmarking dataset consists of four typical experimental results representing common types of plots in quantum computing experiments. To evaluate the agents, we specifically crafted a synthetic dataset where it is difficult to assess success based solely on fitting results—highlighting the need for language model assistance. Each experiment produces 100 successful and 100 failed examples, and the agent is tasked with determining whether each experiment is successful or not. Detailed prompts and reference images are provided below.

\paragraph{Rabi experiment} The rabi experiment represents a set of examples where we expect to observe a set of oscillations. To observe these oscillations, we do a Fourier transform of the input data and plot its spectrum for the LLMs to determine its correctness. Benchmark one sample of this experiment costs on average 6,056 input tokens and 1,220 output tokens.

\begin{demobox}{Examples of the instructions for {\sf NormalisedRabi}}{teal}
\begin{lstlisting}[style=base]
    Analyze this quantum mechanics Rabi oscillation experiment plot in the Fourier frequency domain. A successful experiment should have a significant peak in the figure.
    
    For example, the following Image is a successful Rabi oscillation experiment plot:
\end{lstlisting}
\begin{center}
\includegraphics[width=0.3\textwidth]{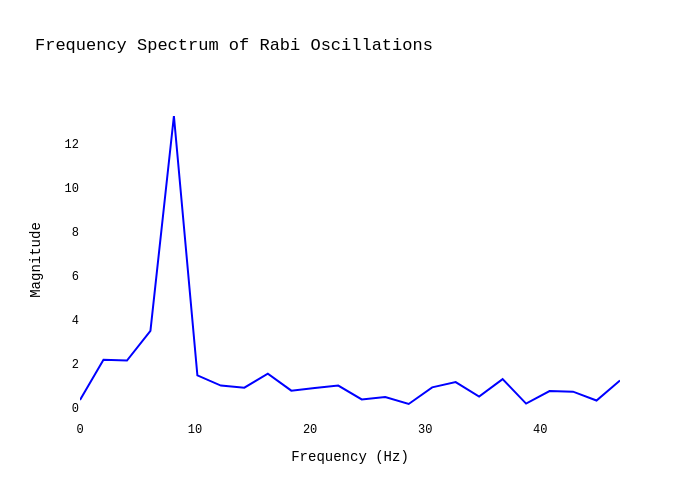}
\end{center}
\begin{lstlisting}[style=base]
    the following Image is a failure case for the Rabi experiment:
\end{lstlisting}
\begin{center}    
\includegraphics[width=0.3\textwidth]{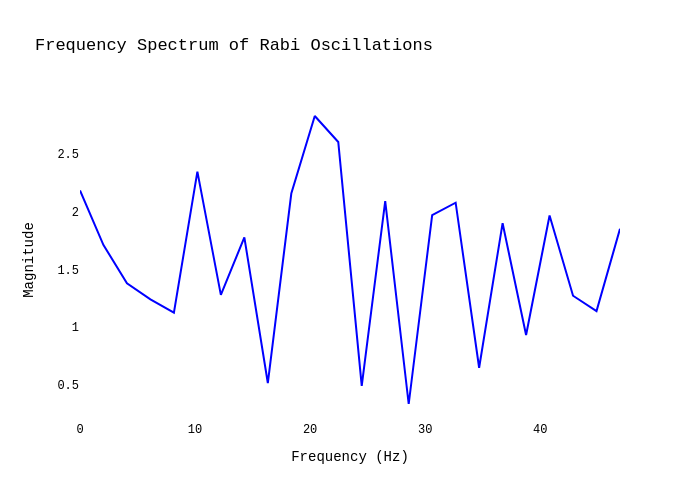}
\end{center}

\end{demobox}

\paragraph{Resonator spectroscopy} The resonator spectroscopy experiment involves looking for a resonator feature in a 1D space. The failure example observes noise and cannot find a clear resonator. Benchmark one sample of this experiment costs on average 6,886 input tokens and 1,365 output tokens.

\begin{demobox}{Examples of the instructions for {\sf ResonatorSpectroscopy}}{teal}
\begin{lstlisting}[style=base]
Analyze the resonator spectroscopy magnitude plot to determine if it exhibits characteristics indicative of a resonator. 
Specifically, look for a sharp dip or peak in the plot, which would signal the presence of a resonator. 
The rest of the plot should remain relatively stable, without significant fluctuations.
Provide a detailed analysis of the magnitude and frequency data. Identifying a resonator indicates a successful experiment.
For example, the following Image is a successful experiment plot:
\end{lstlisting}
\begin{center}    
\includegraphics[width=0.3\textwidth]{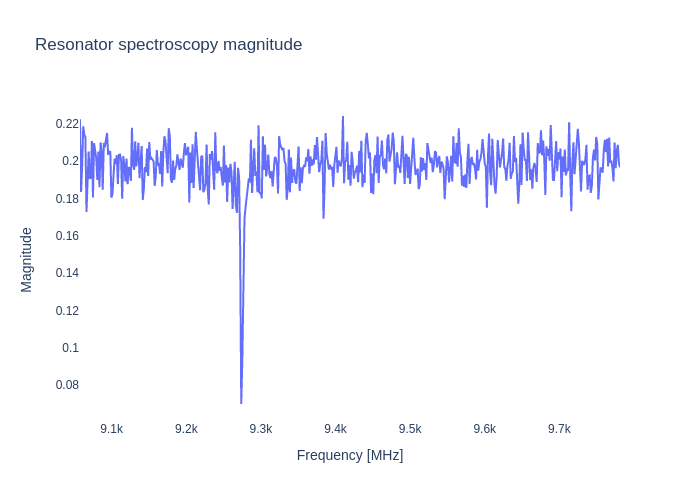}
\end{center}
\begin{lstlisting}[style=base]
the following Image is a failure case for the experiment: 
\end{lstlisting}
\begin{center}    
\includegraphics[width=0.3\textwidth]{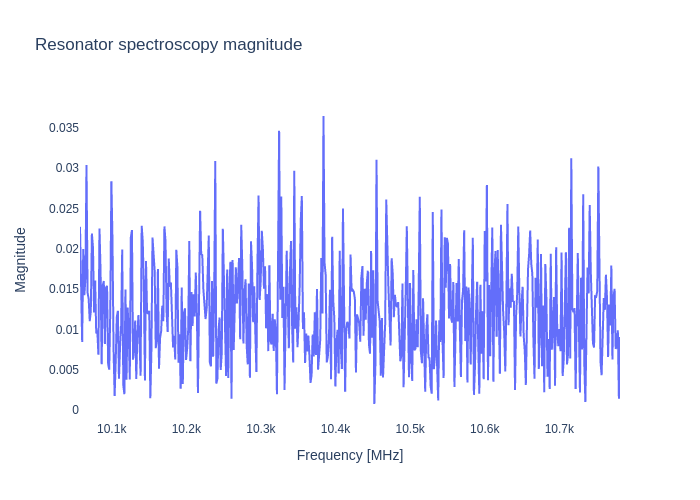}
\end{center}
\end{demobox}

\paragraph{Qubit Readout with Gaussian Mixture model} The experiment visualizes the integrated readout signal on the IQ plane. Typically, the effectiveness of a Gaussian mixture model can be assessed using a spherical Gaussian mixture model (where the covariance matrix is diagonal) by examining the signal-to-noise ratio (SNR), which is the distance between the Gaussian centers divided by the average standard deviation of the two distributions. However, this approach becomes less effective when there are more than two clusters in the response signal, which can occur if the resonator is overdriven or the driving frequency excites higher energy states. In such cases, using the SNR as a metric for model accuracy is not straightforward. In this synthetic dataset, we introduce scenarios where higher excited states may exhibit small leakage or significant population. Describing such complex criteria in natural language also poses challenges. Benchmark one sample of this experiment costs on average 7,840 input tokens and 1,857 output tokens.

\begin{demobox}{Examples of the instructions for {\sf Gaussian Mixture Model readout}}{teal}
\begin{lstlisting}[style=base]
Analyze a plot of collected signal data to determine experiment success:
1. Identify clusters: The signal represents hidden system states, with each state generating a 2D Gaussian distribution (spherical blobs).
2. Count and evaluate distributions:
    - Treat partially overlapped clusters with two visible density centers as separate distributions.
    - Consider elliptical distributions with only one visible density center as a single distribution.
    - Compare densities of observed distributions.
    - If three or more distributions are present, but only two have major density, consider only the two high-density distributions and ignore the low-density ones.
3. Experiment outcome:
    - Success: Exactly two major distributions observed (after accounting for density).
     - Failure: Any other outcome (e.g., one distribution, or more than two major distributions).
For example, the following Image is a successful experiment plot:
\end{lstlisting}
\begin{center}    
\includegraphics[width=0.3\textwidth]{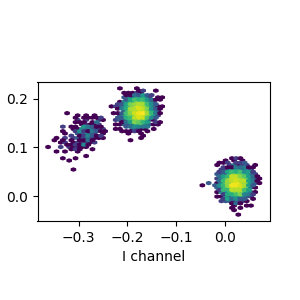}
\end{center}
\begin{lstlisting}[style=base]
the following Image is a failure case for the experiment: 
\end{lstlisting}
\begin{center}    
\includegraphics[width=0.3\textwidth]{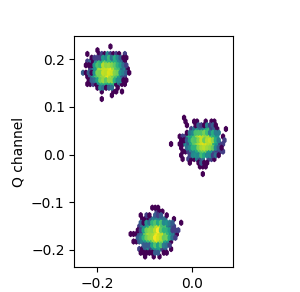}
\end{center}
\end{demobox}

\paragraph{DRAG experiment} The goal of the DRAG experiment calibration is to identify the intersection point of two lines. We synthesize a dataset where the fitting process will always predict a crossing point between the blue and red lines, though the data may be too noisy to be reliable. The language model is designed to assess whether the input data from the experiment is valid and whether the fitting result can be trusted. Benchmark one sample of this experiment costs on average 7,899 input tokens and 1,980 output tokens. 

\begin{demobox}{Examples of the instructions for {\sf DRAG}}{teal}
\begin{lstlisting}[style=base]
Analyze the scatter plot with blue and red data points and trend lines:
    1.Compare the slopes of the trend lines.
    2.Assess how well data points fit their trend lines, noting outliers or patterns.
    3.Evaluate data point distribution along the DRAG coefficient axis.
    4.Determine if trend lines accurately represent their datasets.
    5.Compare trends between the two datasets.
    6.Estimate the fitting residuals.
Success criteria:
    1.Distinct trends for each color
    2.Appropriate line fitting, with the blue and red lines has significant difference in distribution.
    3.Lines intersect near the plot's center region, small shifts away from the center is acceptable.
    4. Residuals are within acceptable range.
If criteria aren't met, mark the experiment as failed and suggest a new range for the sweep.
For example, the following Image is a successful experiment plot:
\end{lstlisting}
\begin{center}    
\includegraphics[width=0.3\textwidth]{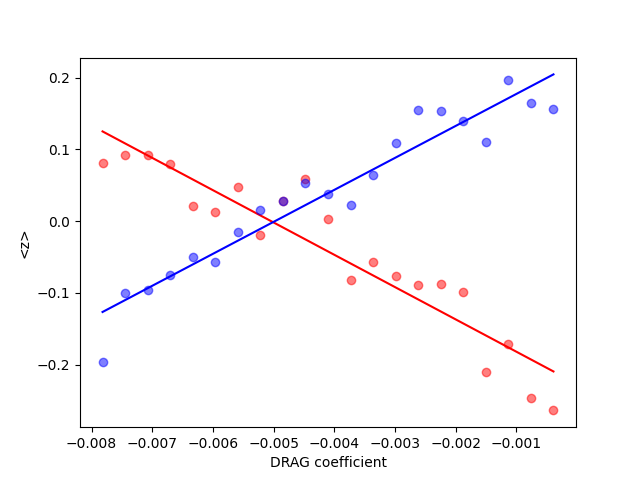}
\end{center}
\begin{lstlisting}[style=base]
the following Image is a failure case for the experiment: 
\end{lstlisting}
\begin{center}    
\includegraphics[width=0.3\textwidth]{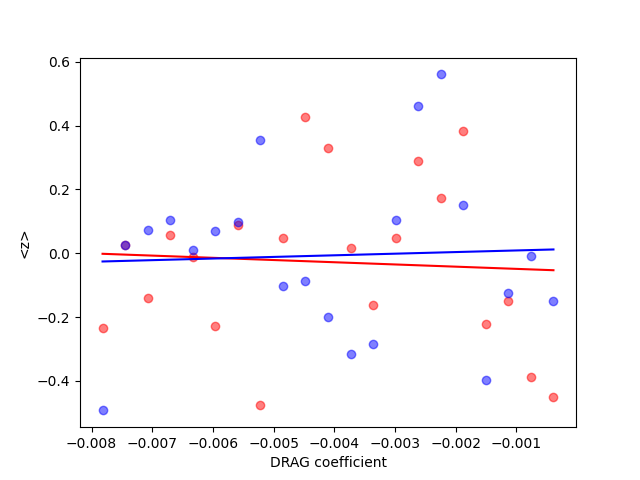}
\end{center}
\end{demobox}

We additionally note that the {\sf Llama-3-70B+LLaVA-v1.6-34B} setup in the figure uses the Llama 3 70B model \cite{dubey2024llama} when there is no image involved. When inspecting a figure, we adopt an LLaVA 1.6 model \cite{liu2024visual, liu2024improved} using the {\sf Nous-Hermes-2-Yi-34B} model \cite{NousHermes2} as its base model. As LLaVA 1.6 does not support including multiple images in the conversation, we only demonstrate its performance with the zero-shot prompt for visual inspection.

%% file: app2_single_qubit_experiments.tex
\section{Built-in experiments\label{app:built-in_experiments}}

Here we detail the implementation of the built-in experiments for single qubit calibration and provide details on how the language model gets prompted. These experiments are standard for superconducting quantum processor calibration, please refer to \cite{10.1063/1.5089550} for more details. For the two-qubit gate related built-in experiments, please refer to appendix \ref{app:two_qubit_calib}.

\paragraph{Ramsey experiment} The Ramsey experiment measures the qubit frequency and coherence time of a qubit by applying two $\pi/2$ pulses separated by a free evolution period, during which the qubit's phase oscillation reflects the offset between the driving frequency and the qubit frequency.

\begin{conversationbox}[Ramsey experiment example]
  \cvline{Execution agent (input)}{Conduct the Ramsey experiment with parameters "frequency offset=0.1 MHz", "stop time=35 microseconds", "step size=0.5 microseconds".}
  \dashedline
\cvline{Code translation agent (output)}{}
\begin{lstlisting}[style=base,language=Python]
  experiment_ramsey= SimpleRamseyMultilevel(dut=dut, set_offset=0.1, stop=35, step=0.5)
\end{lstlisting}
  \dashedline
  \cvline{Experiment plot (input)}{\\
  \begin{center}
  \includegraphics[width=.7\linewidth]
  {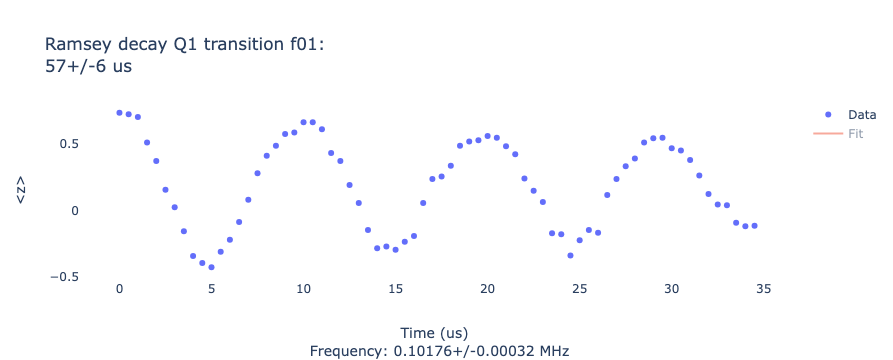}       
  \end{center}
}
  
\cvline{Fitting (input)}{The Ramsey experiment for qubit Q1 has been analyzed. The expected offset was set to 0.100 MHz, and the measured oscillation is 0.102+/-0.000 MHz. Oscillation amplitude is 0.599+/-0.019. The number of oscillations is 3.562+/-0.011.}
  \dashedline
  
  \cvline{Visual inspection agent (output)}{The plot shows data from a quantum mechanics experiment involving Ramsey oscillations. The data points exhibit a clear, regular oscillatory pattern, indicating successful oscillations. The fit line closely follows the data points throughout the plot, suggesting a good fit quality. The data points are relatively tightly clustered around the fit line, with some minor dispersion. The amplitude of the oscillations is around 0.8, which is close to the ideal value of 1, and the frequency is approximately 0.10176 MHz, which is within the expected range. Overall, the plot demonstrates the typical characteristics of a successful Ramsey oscillation experiment, with clear oscillations, a good fit, and consistent amplitude and frequency.}

  \cvline{Final analysis (output)}{The Ramsey experiment for qubit Q1 was successful. The plot shows a clear, regular oscillatory pattern with a good fit. The amplitude of the oscillations is 0.599, which is slightly lower than the ideal value of 1 but still within an acceptable range. The frequency of the oscillations is 0.102 MHz, which is very close to the expected offset of 0.100 MHz. The number of oscillations observed is 3.562, which is within the acceptable range (greater than 3 and less than 10).}

  \dashedline
  
  \cvline{Execution Agent (output)}{The results from the experiment indicate that the Ramsey experiment was successful. The plot shows a clear oscillatory pattern with a good fit, and the final analysis confirms that the frequency and amplitude are within acceptable ranges. The experiment success flag is also set to True, meaning the experiment was successful. Therefore, based on the rules, the next state should be 'Complete'.
}
  
\end{conversationbox}

\paragraph{Rabi experiment} The Rabi experiment involves driving a qubit with an external electromagnetic pulse on resonance to the qubit frequency, to induce transitions between its quantum states. The oscillation speed of the experiment can be used to infer the rough optimal single-qubit gate pulse amplitude.

\begin{conversationbox}[Rabi experiment example]
  \cvline{Execution agent (input)}{Conduct the Rabi experiment to determine the Rabi rate for rough amplitude calibration.}
  
  \dashedline
  \cvline{Code translation agent (output)}{}
\begin{lstlisting}[style=base,language=Python]
  rabi_experiment = NormalisedRabi(dut_qubit=dut, amp=0.2, start=0.01, stop=0.3, step=0.002, fit=True, collection_name='f01', mprim_index=0, pulse_discretization=True, update=True)
\end{lstlisting}
  
  \dashedline
  \cvline{Experiment plot (input)}{\\
  \begin{center}      
  \includegraphics[width=.7\linewidth]
  {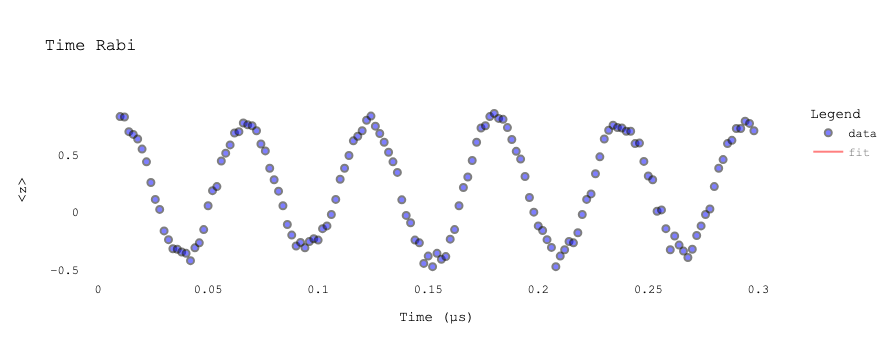} 
    \end{center}

  }
  
\cvline{Fitting (input)}{The fitting result of the Rabi oscillation suggests the amplitude of 0.5893195663534404, the frequency of 17.602131636706986, the phase of 1.5439919053018119. The offset of 0.2087721909927606. The suggested new driving amplitude is 0.19815270673482663.From the fitting results, the plot should exhibit 5.104618174645026 oscillations.}
  \dashedline
  
  \cvline{Visual inspection agent (output)}{The plot shows data points that exhibit a clear, regular oscillatory pattern, indicative of Rabi oscillations. The amplitude of the oscillations appears consistent, with peaks and troughs occurring at regular intervals. The frequency of the oscillations also seems stable throughout the time range. Overall, the plot demonstrates the typical characteristics of successful Rabi oscillation experiments, with well-defined oscillations and consistent amplitude and frequency.}

  \cvline{Final analysis (output)}{The experiment was successful. The plot shows a clear, regular oscillatory pattern with consistent amplitude and frequency, which is characteristic of a successful Rabi oscillation experiment. The fitting results suggest an amplitude of 0.5893, a frequency of 17.6021, and a phase of 1.5440. The plot should exhibit approximately 5.1 oscillations, which is within the acceptable range of 5 oscillations.}

  \dashedline
  
  \cvline{Execution Agent (output)}{Transitioning to the next stage Stage2 with the following description:
Conduct the Pingpong experiment.
The experiment was successful based on the results provided. The plot shows a clear, regular oscillatory pattern with consistent amplitude and frequency, which is characteristic of a successful Rabi oscillation experiment. The fitting results are consistent with the plot, and the final analysis confirms the success of the experiment. Since the experiment was successful, the next stage should be Stage2 (Pingpong Experiment) as per the rules.}
  
\end{conversationbox}

\paragraph{Pingpong experiment} The ping-pong experiment repeatedly applies sequences of quantum gates to estimate the precise amplitude of the single-qubit drive pulse.

\begin{conversationbox}[Pingpong experiment example]
  \cvline{Execution agent (input)}{Conduct the Pingpong experiment.}
  
  \dashedline
  \cvline{Code translation agent (output)}{}

\begin{lstlisting}[style=base,language=Python]
  experiment_pingpong = AmpPingpongCalibrationSingleQubitMultilevel(dut=dut, iteration=9, points=10, mprim_index=0, repeated_gate='X', flip_other=False, ai_inspection=True)
\end{lstlisting}
  \dashedline
  \cvline{Experiment plot (input)}{\\
  \begin{center}
\includegraphics[width=.5\linewidth]
  {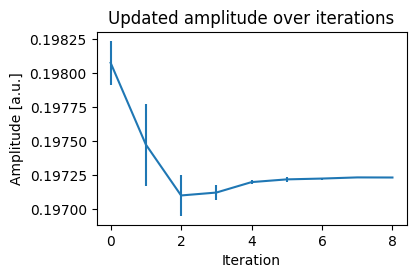} 
        
  \end{center}
  }
  \dashedline
  
  \cvline{Visual inspection agent (output)}{The amplitude plot shows a decreasing trend initially, followed by a stabilization around the 5th iteration. This indicates that the amplitude converges through the iterations, as the values become relatively constant after the initial decrease.}

  \cvline{Final analysis (output)}{The amplitude plot shows a decreasing trend initially, followed by stabilization around the 5th iteration, indicating that the amplitude converges through the iterations. This suggests that the experiment was successful in achieving fine calibration of the single qubit pulses.}

  \dashedline
  
  \cvline{Execution Agent (output)}{The results from both the plot and Final analysis indicate that the amplitude converges and stabilizes around the 5th iteration, which suggests that the experiment was successful in achieving fine calibration of the single qubit pulses. Additionally, the 'Experiment success' result is marked as True, confirming that the experiment was successful. Therefore, based on the rules, the next stage should be 'Complete' as the experiment has been successful.
}
  
\end{conversationbox}

\paragraph{DRAG calibration experiment} The DRAG (Derivative Removal by Adiabatic Gate) calibration experiment we used in this experiment reduces phase errors during gate operations by sweeping the DRAG parameter of the pulse.

\begin{conversationbox}[DRAG experiment example]
  \cvline{Execution agent (input)}{Conduct the DRAG calibration on the single qubit gate.}
  
  \dashedline
  \cvline{Code translation agent (output)}{} 

\begin{lstlisting}[style=base,language=Python]
experiment_drag = DragCalibrationSingleQubitMultilevel(dut=dut, collection_name='f01', mprim_index=0, initial_lpb=None, N=1, num=21)
\end{lstlisting}

  \dashedline
  \cvline{Experiment plot (input)}{\\
\begin{center}
  \includegraphics[width=.4\linewidth]
  {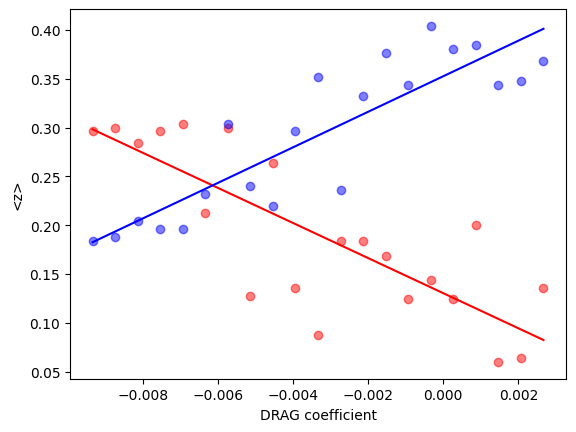}  
\end{center}
}
  
\cvline{Fitting (input)}{Sweep start: -0.009333333333333334 Sweep stop: 0.0026666666666666666 The fitting results are as follows: The estimated optimal DRAG coefficient is -0.0061392677674084704 The estimated optimal DRAG coefficient falls within the central half of the sweep.Residual average for Xp: 0.0023451666048237476, Residual average for Xm: 0.0012794699649556793}
  \dashedline
  
  \cvline{Visual inspection agent (output)}{1. The slope of the blue trend line is positive, indicating an increasing trend, while the slope of the red trend line is negative, indicating a decreasing trend. 2. The data points for both colors generally fit their respective trend lines, though there are a few outliers, particularly among the red data points. 3. The data points are distributed along the DRAG coefficient axis, with blue points more concentrated on the right side and red points more concentrated on the left side. 4. The trend lines appear to accurately represent their datasets, capturing the overall direction of the data points. 5. The trends between the two datasets are distinct, with one increasing and the other decreasing. 6. The fitting residuals seem to be within an acceptable range, though there are some deviations, particularly among the red data points.}

  \cvline{Final analysis (output)}{The experiment shows distinct trends between the two datasets, with the blue trend line increasing and the red trend line decreasing. The fitting residuals are within an acceptable range, although there are some deviations, particularly among the red data points. The estimated optimal DRAG coefficient is -0.0061, which falls within the central half of the sweep range, indicating that the sweep was appropriately centered. Overall, the experiment meets the success criteria.}

  \dashedline
  
  \cvline{Execution Agent (output)}{The experiment has been successful based on the results provided. The final analysis confirms that the experiment meets the success criteria, with the estimated optimal DRAG coefficient falling within the central half of the sweep range and the residuals being within an acceptable range. Additionally, the 'Experiment success' result is marked as 'True', indicating no need for further retries or adjustments.}
  
\end{conversationbox}

\paragraph{Randomized benchmarking experiment} The randomized benchmarking experiment evaluates the error rates of quantum gates by applying a sequence of random gate operations followed by their inverse, measuring how well the system returns to its initial state~\cite{Emerson_2005, Unitary2designRB, InterleavedRB}. 

\begin{conversationbox}[Randomized benchmarking example]
  \cvline{Execution agent (input)}{Measure the single qubit gate fidelity using randomized benchmarking.}
  
  \dashedline
  \cvline{Code translation agent (output)}{}
  
\begin{lstlisting}[style=base,language=Python]  
  experiment_rb = SingleQubitRandomizedBenchmarking(dut=dut, collection_name='f01', seq_length=1024, kinds=10, cliff_set='XY', pi_half_only=False, mprim_index=0, seed=42)
\end{lstlisting}
  
  \dashedline
  \cvline{Experiment plot (input)}{\\
  \begin{center}
      
  \includegraphics[width=.4\linewidth]
  {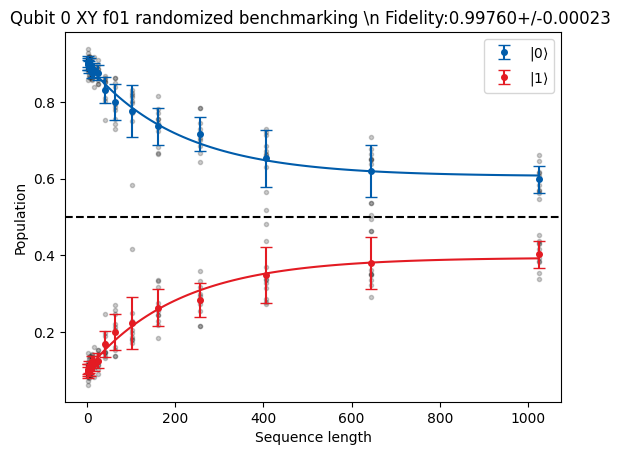} 
  \end{center}
}
  
\cvline{Fitting (input)}{The randomized benchmarking experiment suggest the infidelity per clifford is 0.00240+/-0.00023 and the infidelity per gate is 0.00131+/-0.00013.}
  \dashedline
  
  \cvline{Visual inspection agent (output)}{The plot shows the population of the $\ket{0}$ state (blue) and $\ket{1}$ state (red) as a function of sequence length in a randomized benchmarking experiment. The $\ket{0}$ state exhibits a clear exponential decay, indicating that the experiment is successful. The decay rate appears to be neither too fast nor too slow, suggesting that the sequence length is appropriate.}

  \cvline{Final analysis (output)}{The experiment shows a clear exponential decay in the population of the $\ket{0}$ state, indicating that the experiment is successful. The infidelity per Clifford and per gate are within reasonable physical limits, suggesting that the experiment was conducted properly.
}

  \dashedline
  
  \cvline{Execution Agent (output)}{The results from the experiment indicate a successful measurement. The plot shows a clear exponential decay in the population of the $\ket{0}$ state, which is consistent with the expected behavior in a randomized benchmarking experiment. The infidelity per Clifford and per gate are within reasonable physical limits, and there are no suggested parameter updates. Additionally, the final analysis confirms that the experiment was successful, and the 'Experiment success' flag is set to True. Therefore, the fidelity measurement is successful, and the next stage should be 'Complete'.

}
  
\end{conversationbox}

%% file: app3_two_qubit_experiment.tex
\section{Two-qubit siZZle gate \label{app:two_qubit_calib}}

The siZZle gate (Stark-induced ZZ by level excursions) is one method to implement two-qubit interactions on fixed-frequency transmon qubits \cite{PhysRevLett.127.200502, PhysRevResearch.4.023040,wei2021quantum,PhysRevLett.129.060501}. The siZZle gate off-resonantly drives two qubits, inducing a shift in the energy levels of the two-qubit system, which modifies the ZZ interaction. An approximate formula for the ZZ interaction rate, denoted as $\tilde{\nu}_{ZZ}$, is \cite{wei2021quantum}:

\begin{equation} \tilde{\nu}_{ZZ} = \nu_{ZZ,s} + \frac{2J \alpha_0 \alpha_1 \Omega_0 \Omega_1 \cos(\phi_0 - \phi_1)}{\Delta_{0,d} \Delta_{1,d} (\Delta_{0,d} + \alpha_0)(\Delta_{1,d} + \alpha_1)} \end{equation}

where $\nu_{ZZ,s}$ is the static ZZ interaction rate without any applied drives. $\alpha_0$ and $\alpha_1$ represent the anharmonicity of the two qubits. $\Delta_{0,d}$ and $\Delta_{1,d}$ are the detunings of the driving frequencies relative to the transition frequencies of the two qubits. $\phi_0 - \phi_1$ is the relative phase between the two drives applied to the qubits. $J$ denotes the capacitive coupling strength between the qubits, while $\Omega_0$ and $\Omega_1$ are the driving amplitudes of the pulses applied to each qubit. 

To calibrate a siZZle gate, the objective is to find the optimal method to drive the two qubits such that the accumulation of the $ZZ$ interaction over time is precisely $\pi/4$. This involves adjusting the driving frequency, which determines $\Delta_{0,d}$ and $\Delta_{1,d}$, the driving amplitude (which sets $\Omega_0$ and $\Omega_1$), and the driving time. Since the ZZ interaction is proportional to $\Omega_0 \Omega_1$, we typically choose $\Omega_0 = \Omega_1$ in experiments to maximize the ZZ interaction. However, this does not imply that the amplitudes are exactly the same, as variations in cable attenuations and microwave component performance can cause discrepancies. Therefore, to reproduce the above behavior, we request the agent modify the drive amplitude of the first qubit, while our code infers the amplitude for the second qubit based on previous calibration results of single-qubit gates. In the formula, the relative phase of the two drives, $\phi_0 - \phi_1$, is also significant. Due to the experimental setup, the phase reaching the qubits may differ from the phase difference sent by the electronics. In our experiment, we found that our phase was close to the optimal value, so we did not optimize this parameter.

\begin{figure}
    \centering
    \includegraphics[width=\linewidth]{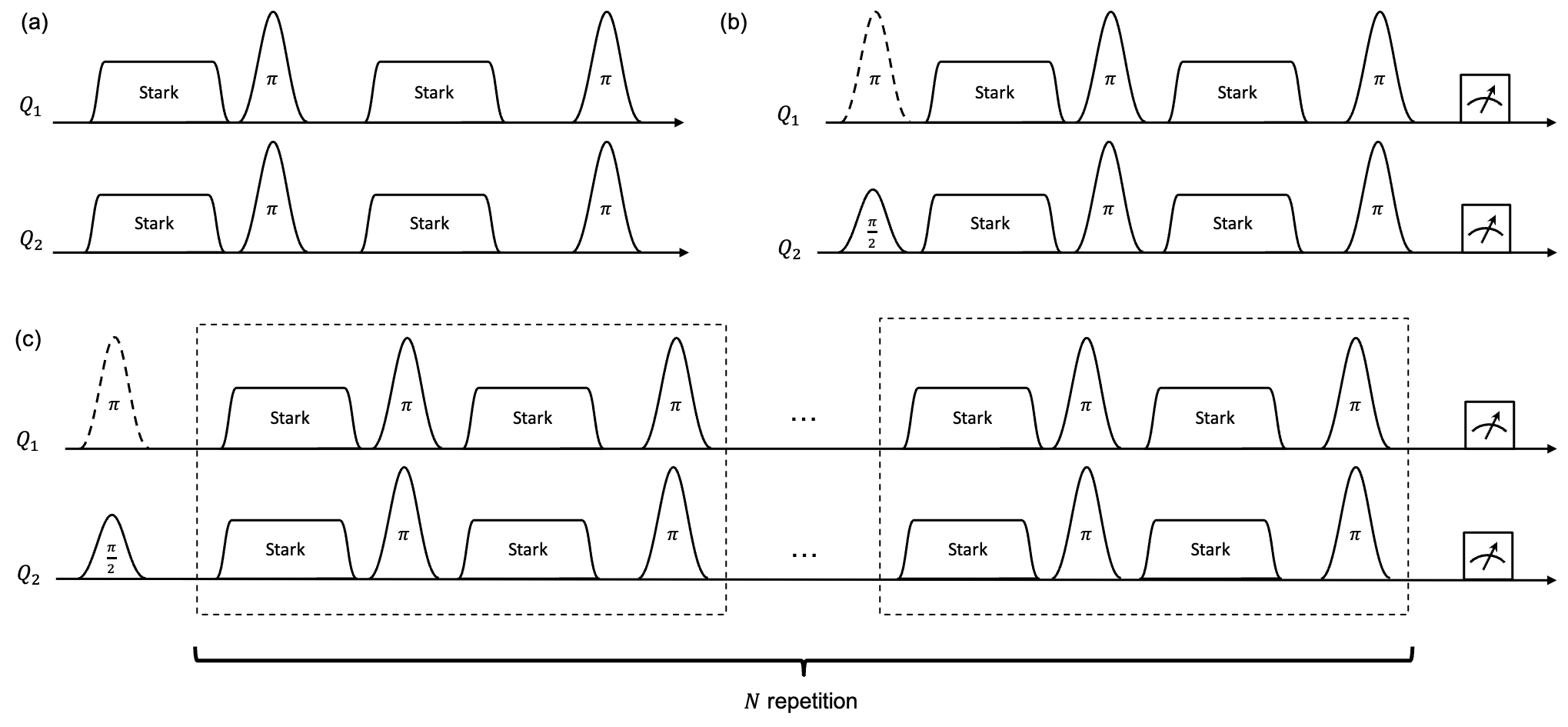}
    \caption{Pulse sequences for calibrating the siZZle gate. (a) The pulse sequence used to implement a single siZZle gate consists of two off-resonance drives to induce a Stark shift (the Stark drive), followed by single-qubit $\pi$ pulses inserted between and at the end of the drive. These $\pi$ pulses are used to cancel the single qubit phase accumulated on individual qubits. (b) The experimental pulse sequence for Hamiltonian tomography to measure the ZZ interaction strength. In this experiment, the pulse width is varied to observe the oscillatory behavior in the phase of the second qubit, which allows inference of the ZZ interaction strength. The dashed line on the pulse for $Q_1$ indicates that the experiment is performed both with and without preparing $Q_1$ in the excited state. (c) Repeated gate tomography experiment. Unlike Hamiltonian tomography, this experiment uses discrete pulses rather than continuously sweeping the width of the Stark pulse. This approach offers more precise calibration of the accumulated ZZ interaction over one repetition block, which can then be used to implement an entangling gate.}
    \label{fig:pulse_sequences_for_two_qubits_exps}
\end{figure}

The pulse sequence for implementing a siZZle gate is shown in figure \ref{fig:pulse_sequences_for_two_qubits_exps}(a). Typically, in the manual calibration approach, a human scientist selects a pair of driving frequency and amplitude of the Stark drive and then observes the stability of the quantum system. The stability is observed by implementing the Hamiltoniain tomography experiment followed by repeated gate tomography experiment, shown in figure \ref{fig:pulse_sequences_for_two_qubits_exps}(b) and (c), respectively. If the chosen parameters are not optimal, the oscillation of the expectation value of the quantum state tends to be chaotic and noisy. The scientist then reviews the experimental results and, based on empirical knowledge, suggests a new set of parameters to test. This iterative process is often the most time-consuming aspect of tuning a two-qubit gate. Once a stable and clear result is observed—indicating that the quantum system's dynamics are behaving as expected, an automated code can be employed to determine the necessary time duration, which is the final parameter required to execute the gate.

To provide the human evaluator with an intuitive understanding of the agent's exploration behavior, we perform a grid scan over the domain of frequency and amplitude, gathering information related to the ZZ interaction strength. While the ideal approach would be directly measuring the interaction strength at each point, this is impractical and too time-consuming. Instead, we employ the pulse sequence shown in figure \ref{fig:pulse_sequences_for_two_qubits_exps}(b), where the duration of the Stark drive pulse is fixed at $1 \mu s$, and we measure the difference in the expectation value when $Q_1$ is prepared to the ground and excited state. The entire experiment takes approximately 12 hours to generate the background figure.

An example of the search iteration on the real hardware is shown as follows. The inspection agent only inspects the control qubit state plot and the Fourier transform of the phase oscillation. The other plots are for humans to evaluate the experiment result.

\begin{conversationbox}[siZZle gate parameter search]
  \cvline{Parameter agent}{Since no experiments have been run yet, we should start by trying a frequency between the two qubits, which is around 4858 MHz. We can start with a gentle amplitude of 0.2 for the control qubit, which is close to its single qubit gate amplitude. The other parameters are set based on the given rules. params: \{'frequency': 4858.0, 'amp\_control': 0.2, 'amp\_target': 0.0, 'rise': 0.015, 'width': 0.1, 'phase\_diff': 0.0, 'zz\_interaction\_positive': True\}
}\\

  \cvline{Translation agent}
\begin{lstlisting}[style=base,language=Python]    
  experiment_conditional_stark_shift_continuous = ConditionalStarkShiftContinuous(qubits=duts, frequency=4858.0, amp_control=0.2, amp_target=0.09943319110052375, rise=0.015, start=0, stop=20, sweep_points=40, phase_diff=0.0, echo=True)
\end{lstlisting}

  \cvline{Experiment plots}{\\
  \begin{center}
  \includegraphics[width=.8\linewidth]{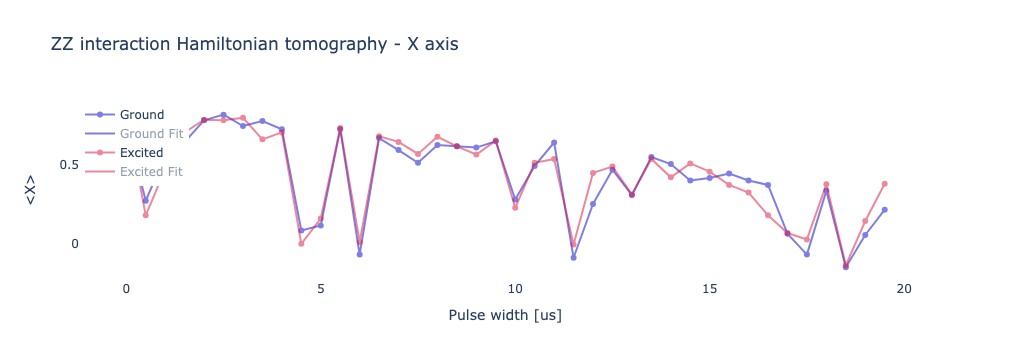} 
  
  \includegraphics[width=.8\linewidth]{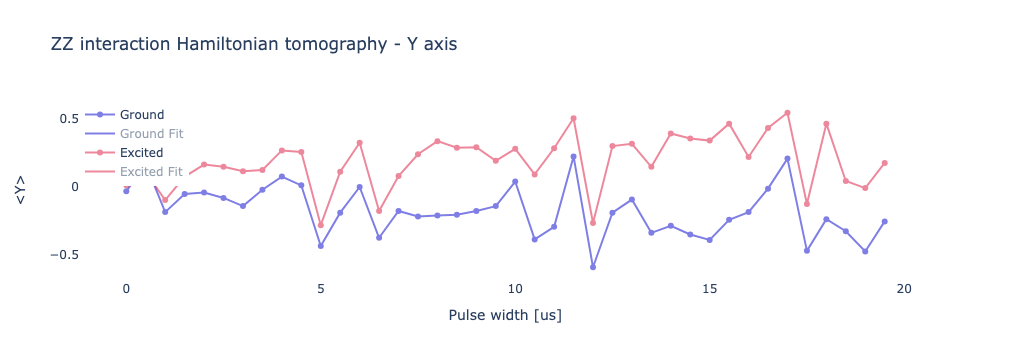} 
  
  \includegraphics[width=.8\linewidth]{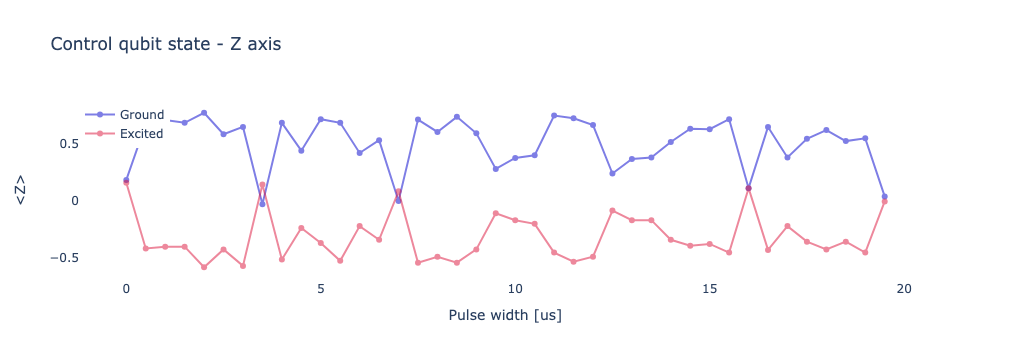} 
  
  \includegraphics[width=.8\linewidth]{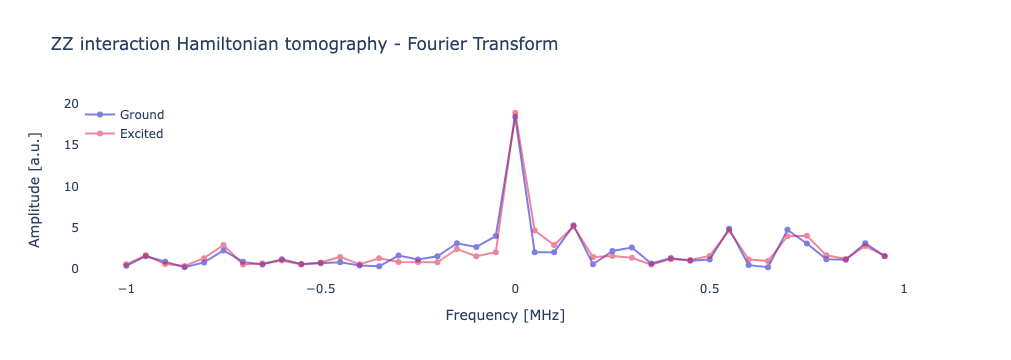}
  \end{center}
  
  }

  \cvline{Inspection agent}{The experiment failed due to the Fourier transform plot showing no clear symmetric peaks around the center frequency for the ground and excited states, indicating unsuccessful measurement of IZ and ZZ interactions.}

  \dashedline

  \cvline{Parameter agent}{The previous experiment failed because the frequency was too close to the qubits' frequencies. We should try a frequency further away from the qubits. Let's start with a frequency below the lowest qubit frequency and a gentle control amplitude. }
  
\begin{lstlisting}[style=base,language=Python] 
params: {'frequency': 4800.0, 'amp_control': 0.2, 'amp_target': 0.09776666514167257, 'rise': 0.015, 'width': 1.1730133547479822, 'phase_diff': 0.0, 'zz_interaction_positive': True}
\end{lstlisting}
  \dashedline
  \cvline{Translation agent:}{}

\begin{lstlisting}[style=base,language=Python] 
  experiment_conditional_stark_shift_continuous = ConditionalStarkShiftContinuous(qubits=duts, frequency=4858.0, amp_control=0.2, amp_target=0.09943319110052375, rise=0.015, start=0, stop=20, sweep\_points=40, phase_diff=0.0, echo=True)
\end{lstlisting}

  \cvline{Experiment plots}{\\
  \begin{center}
\includegraphics[width=.8\linewidth]{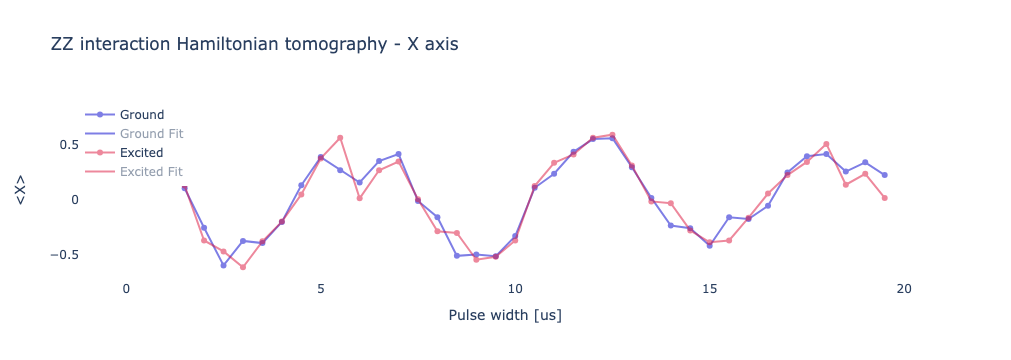} 
  
  \includegraphics[width=.8\linewidth]{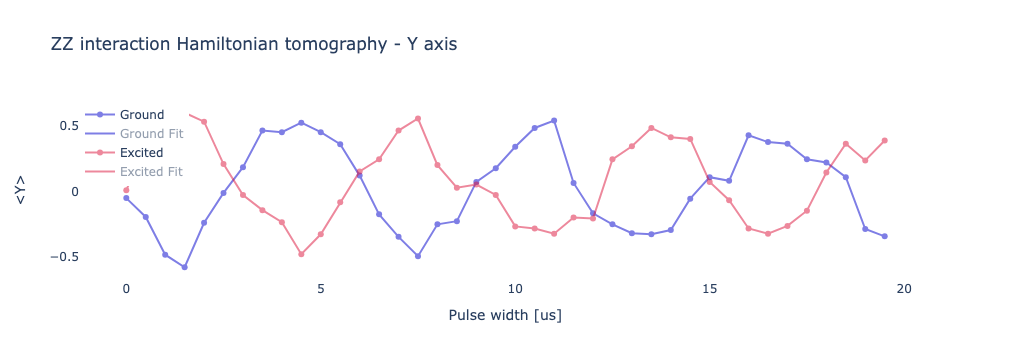} 
  
  \includegraphics[width=.8\linewidth]{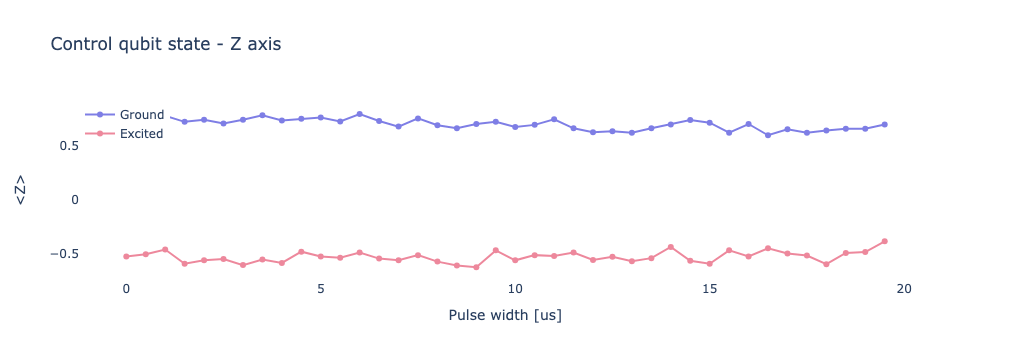} 
  
  \includegraphics[width=.8\linewidth]{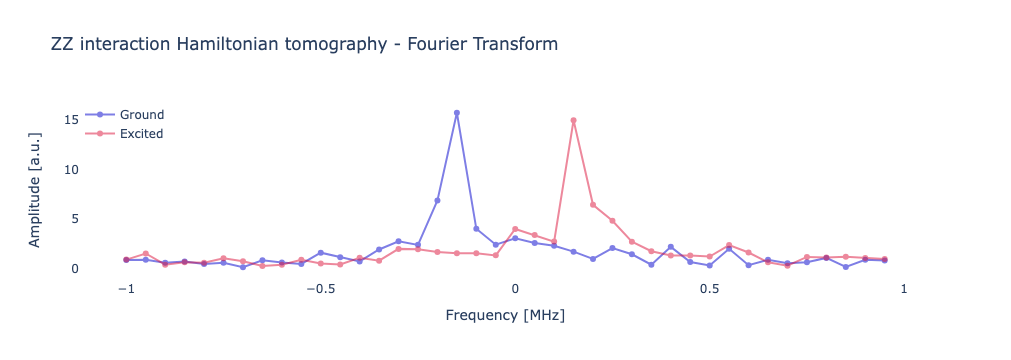}

  \end{center}
}

  \cvline{Inspection agent}{The experiment was successful based on the clear, symmetric peaks in the Fourier plot, stable control qubit expectation value, and reasonable fitting results showing a ZZ interaction strength of 0.16543846027754651 MHz.}

  \cvline{Translation agent}{}
  
\begin{lstlisting}[style=base,language=Python] 

  experiment_result = ConditionalStarkShiftRepeatedGate(duts=duts, frequency=4800.0, amp_control=0.2, amp_target=0.09943319110052375, rise=0.015, start_gate_number=0, gate_count=32, width=0.3778425958178225, phase_diff=0.0, echo=True)
  
\end{lstlisting}

  \cvline{Experiment plots}{\\
\begin{center}
      \includegraphics[width=.8\linewidth]{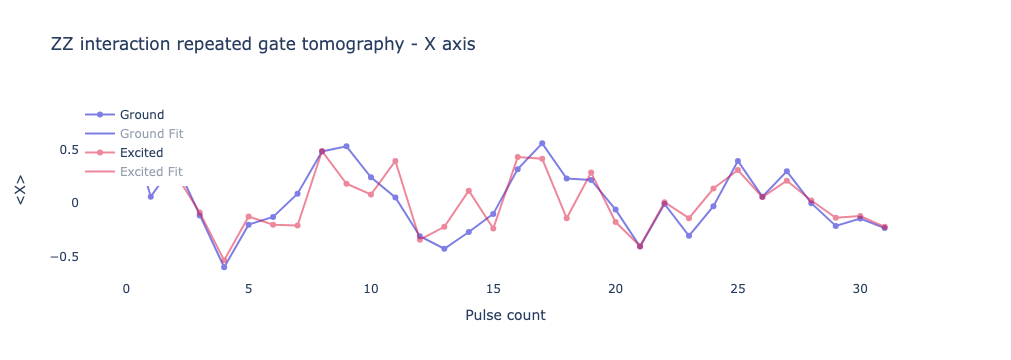} 
  
  \includegraphics[width=.8\linewidth]{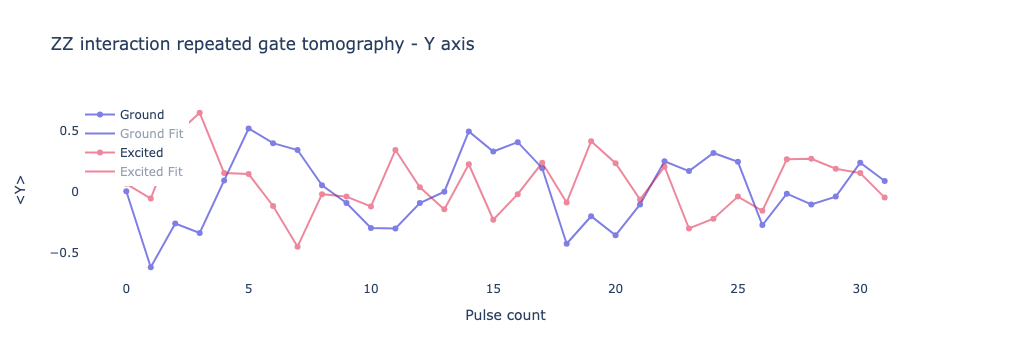} 
  
  \includegraphics[width=.8\linewidth]{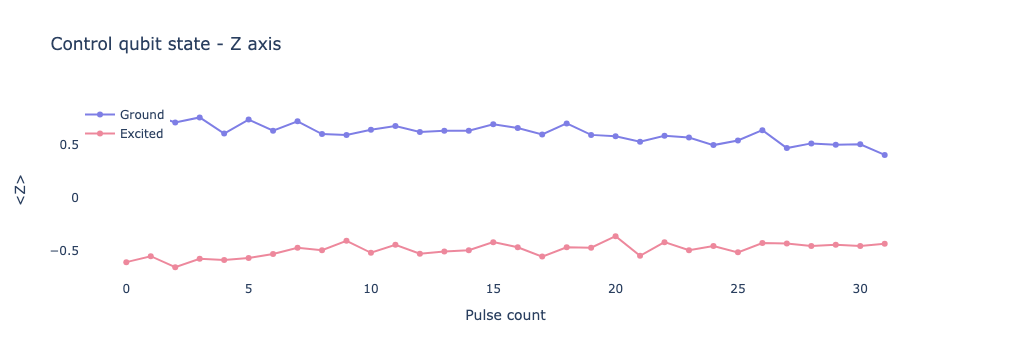} 
  
  \includegraphics[width=.8\linewidth]{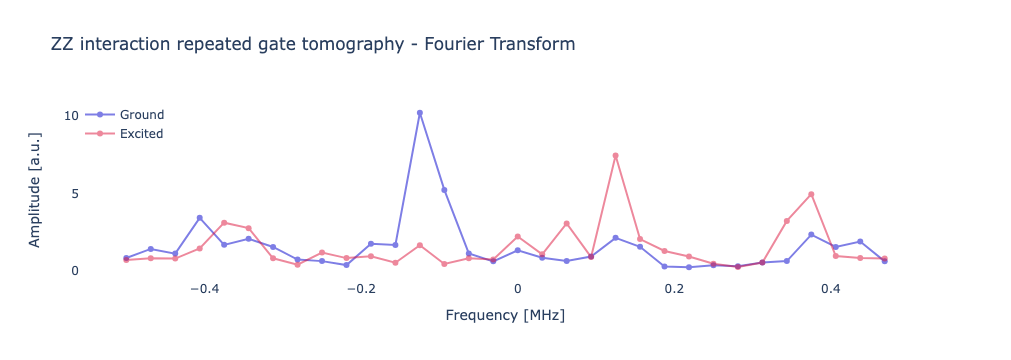}
\end{center}
}

\cvline{Inspection agent}{The data inspection AI reports that the Fourier transform plot shows two clear, symmetric peaks for the ground and excited states, and the expectation value of the control qubit along the Z axis remains stable throughout the experiment. The minimum difference between the ground and excited states is greater than 25\% of the maximum difference, indicating a successful experiment. The fitting results estimate the ZZ interaction strength to be -0.1170390305626309 MHz, which is a reasonable and physical value. Both the data inspection and fitting results suggest that the experiment was successful.}

\end{conversationbox}

%% file: app5_device_setup.tex
\section{Experimental Hardware and Setup}
 \label{app:Hardwrae}

\begin{figure*}[htbp]
    \centering
    \begin{subfigure}[t]{0.5\textwidth}
        \centering
        \includegraphics[height=6cm]{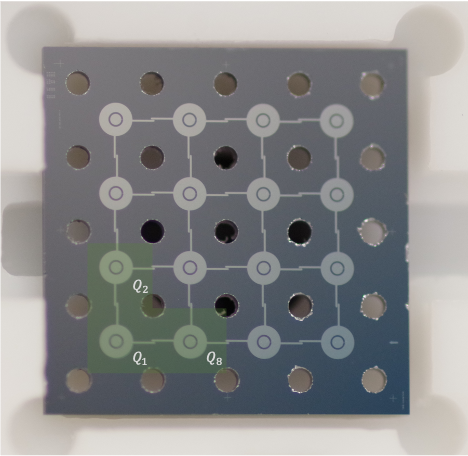}
        \caption{}
    \end{subfigure}%
    \hfill
    \begin{subfigure}[t]{0.5\textwidth}
        \centering
        \includegraphics[height=6cm]{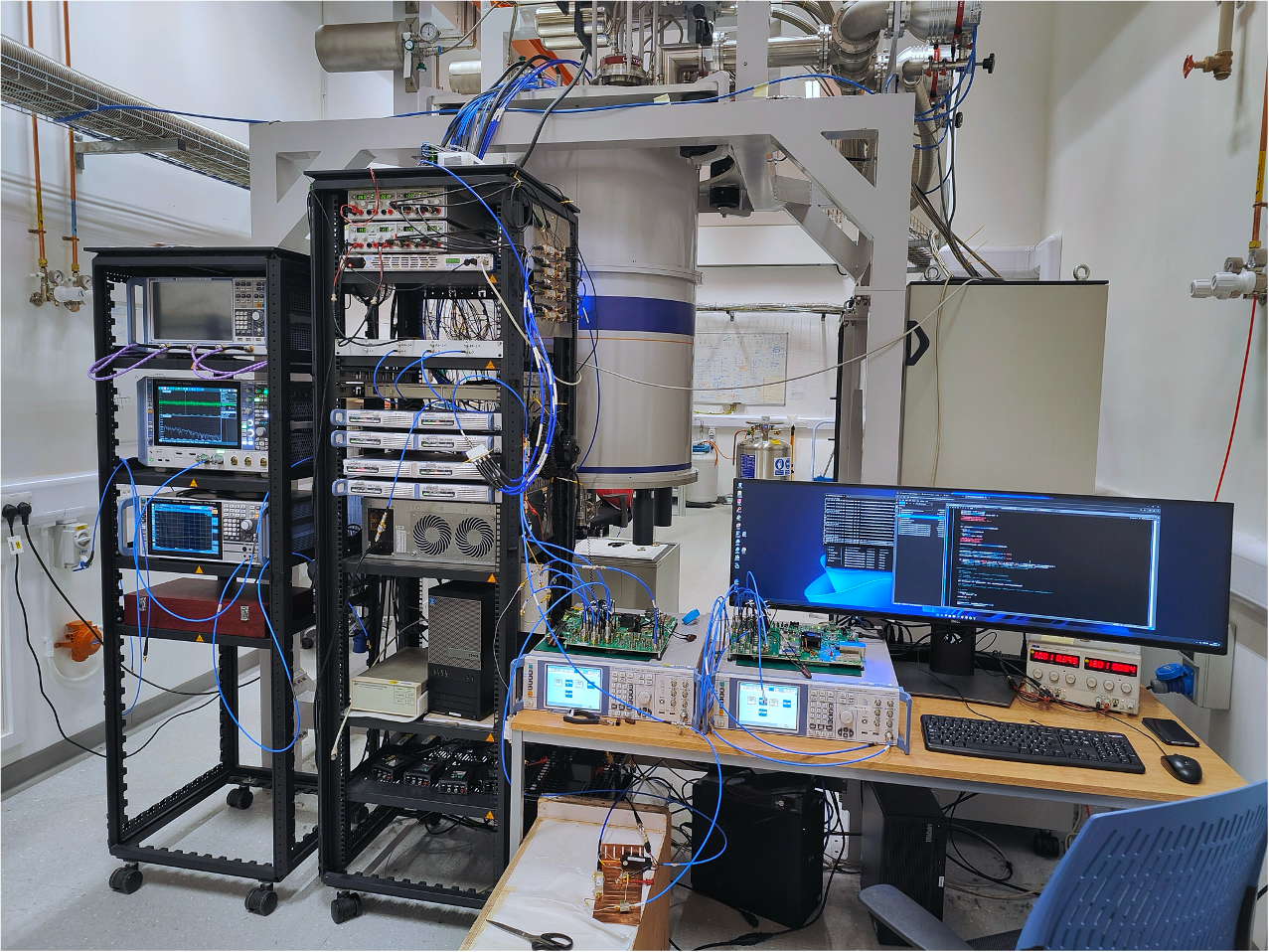}
        \caption{}
    \end{subfigure}
    \caption{(a) The photo of the superconducting quantum processor chip used in this experiment. The three qubits involved in this experiment are highlighted in green color.  (b) The photo of the hardware setup in the lab.}
    \label{fig:hardware_setup}
\end{figure*}

The quantum device contains 16 transmon qubits in 3D-integrated coaxial circuit quantum electrodynamics (cQED) architecture \cite{Spring2022, Rahamim2017Double-sided,2501.15059}. \sx{The details of this device has been reported in \cite{2505.22276}.} Each qubit consists of a Josephson junction (JJ) and a readout resonator positioned on opposite sides of the chip, enabling precise couplings to off-chip coaxial ports for control and readout. The qubits are arranged in a $4 \times 4$ lattice and are statically coupled through capacitive elements, facilitating interactions characterized by exchange energy rates (\( J \)) between each pair of qubits. The layout of the chip, depicted in Figure \ref{fig:hardware_setup}(a), illustrates the positioning of the three qubits used in this work.

The basic characterization of the three selected qubits is detailed in Table \ref{tab:extended_qubit_characterization}. These qubits operate in the straddling regime, that is, detuning frequencies smaller than the average anharmonicity (\(\langle \alpha\rangle = 196.4 \, \text{MHz}\)) across the device. This ensures that most qubits remain within a regime where their interactions can be effectively controlled and utilized to drive two-qubit gates \cite{Emerson_2005, Unitary2designRB, InterleavedRB}. 

\begin{table}[htbp]
    \centering
    \caption{Basic characterization of the three qubits used in this work. $Q_{\text{int}}$ is the internal Q factor of the resonator, $\kappa_{\text{ext}}$ is the external decay rate of the resonator, and $\mathcal{F}$ is single-qubit gate fidelity by randomized benchmarking, where all Clifford gates are decomposed into physical X and Y gates.}
    \scalebox{1.2}{ 
        \begin{tabular}{|c|c|c|c|c|c|c|c|c|c|c|c|c|}
            \hline
            Qubit & \( \omega_{q}/2\pi \) & \( \omega_{r}/2\pi \) & \( Q_{\text{int}} \) & \( \kappa_{\text{ext}}/2\pi \) & \( \chi/2\pi \) & \( \alpha/2\pi \) & \( T_{1} \) & \( T_{2R} \) & \( T_{2E} \) & \( \mathcal{F} \) & \( J_{Q1-Q2} \) & \( J_{Q1-Q8} \) \\
                  & GHz & GHz & $10^{4}$ & MHz & kHz & MHz & \(\mu\)s & \(\mu\)s & \(\mu\)s & $\%$ & \multicolumn{2}{c|}{MHz} \\
            \hline
            \( Q_{1} \) &   4.888   &   9.997   &  11.661   &   2.645   &  -200   &  -197   &   126\,(18)   &  107\,(12)  &  124\,(23)   &  99.993  & \multirow{2}{*}{0.760} &  -  \\ 
            \( Q_{2} \) &   4.795   &   9.386   &  11.785   &   1.337   &  -225   &  -197   &   89\,(13)   &  56\,(15)  &  86\,(12)   &  99.861  & &  \multirow{2}{*}{0.702} \\ 
            \( Q_{8} \) &   4.829   &   9.908   &  5.843    &   2.040   &  -175   &  -197   &  63\,(8)    & 32\,(7)   & 71\,(8)    &  99.944  &  -  &   \\
            \hline
        \end{tabular}
    }
    \label{tab:extended_qubit_characterization}
\end{table}

\begin{figure}[!h]
    \centering
    \includegraphics[width=\linewidth]{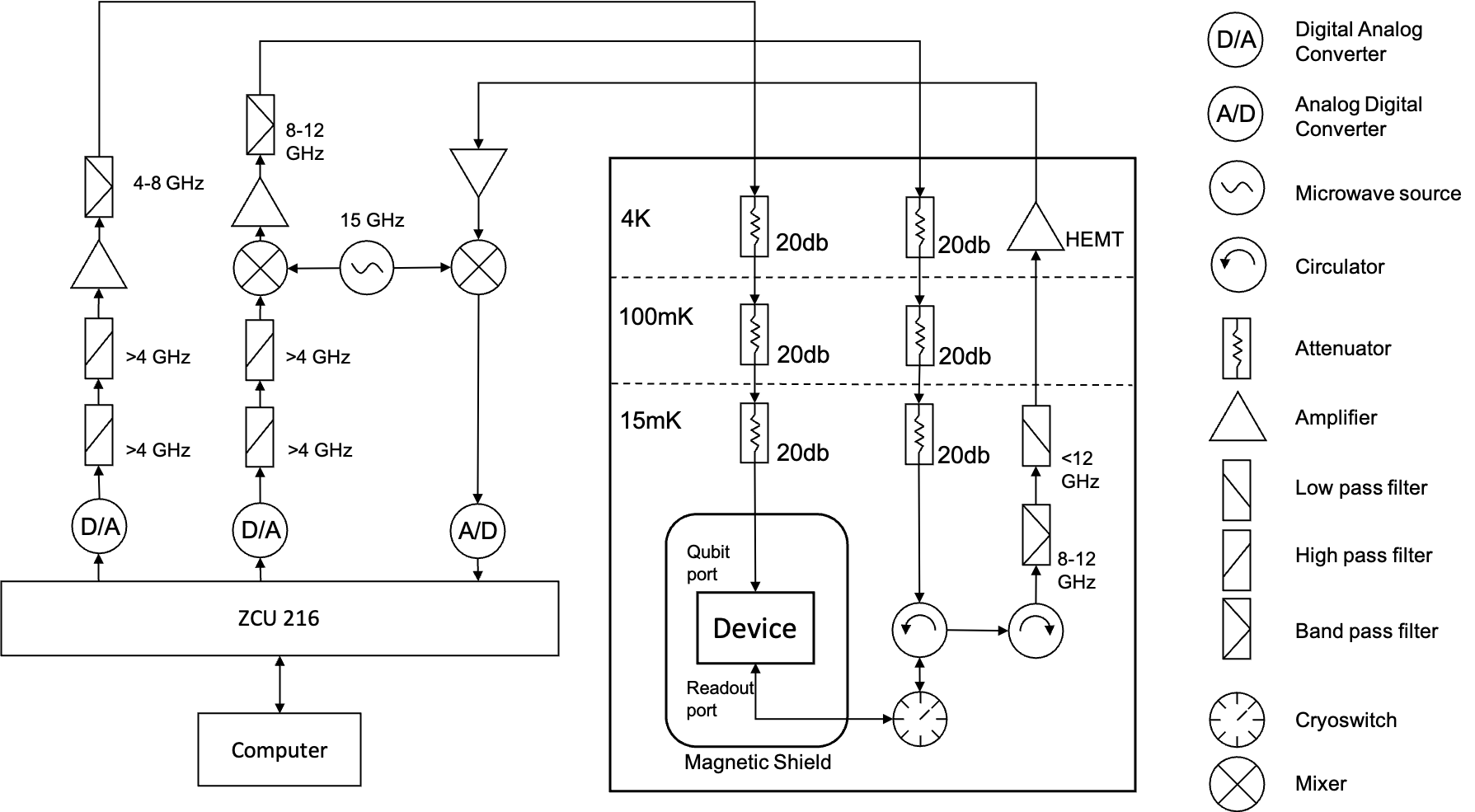}
    \caption{Schematic diagrams of the full experimental setup, including wiring and cryogenic components.}
    \label{fig:fridge_diagram}
\end{figure}

The experimental setup operates at a base temperature of approximately \( 15 \, \text{mK} \) using a \( ^3\text{He}/^4\text{He} \) dilution refrigerator. The control and readout of qubits are facilitated by the wiring configurations shown in Figure \ref{fig:fridge_diagram}. For qubit control, microwave pulses are synthesized directly using 
the QubiC system \cite{2309.10333}. See Figure \ref{fig:hardware_setup}(b) for the lab photo. The pulses are carefully shaped for single and two-qubit gates to align at the desired frequencies. For readout, reflected signals from the resonators undergo amplification and down-conversion and are then captured by Analog-to-Digital Converters (ADCs) connected to the FPGA for measurements and further data analysis.

%% file: main.bbl
\begin{thebibliography}{82}%
\makeatletter
\providecommand \@ifxundefined [1]{%
 \@ifx{#1\undefined}
}%
\providecommand \@ifnum [1]{%
 \ifnum #1\expandafter \@firstoftwo
 \else \expandafter \@secondoftwo
 \fi
}%
\providecommand \@ifx [1]{%
 \ifx #1\expandafter \@firstoftwo
 \else \expandafter \@secondoftwo
 \fi
}%
\providecommand \natexlab [1]{#1}%
\providecommand \enquote  [1]{``#1''}%
\providecommand \bibnamefont  [1]{#1}%
\providecommand \bibfnamefont [1]{#1}%
\providecommand \citenamefont [1]{#1}%
\providecommand \href@noop [0]{\@secondoftwo}%
\providecommand \href [0]{\begingroup \@sanitize@url \@href}%
\providecommand \@href[1]{\@@startlink{#1}\@@href}%
\providecommand \@@href[1]{\endgroup#1\@@endlink}%
\providecommand \@sanitize@url [0]{\catcode `\\12\catcode `\$12\catcode `\&12\catcode `\#12\catcode `\^12\catcode `\_12\catcode `\%12\relax}%
\providecommand \@@startlink[1]{}%
\providecommand \@@endlink[0]{}%
\providecommand \url  [0]{\begingroup\@sanitize@url \@url }%
\providecommand \@url [1]{\endgroup\@href {#1}{\urlprefix }}%
\providecommand \urlprefix  [0]{URL }%
\providecommand \Eprint [0]{\href }%
\providecommand \doibase [0]{http://dx.doi.org/}%
\providecommand \selectlanguage [0]{\@gobble}%
\providecommand \bibinfo  [0]{\@secondoftwo}%
\providecommand \bibfield  [0]{\@secondoftwo}%
\providecommand \translation [1]{[#1]}%
\providecommand \BibitemOpen [0]{}%
\providecommand \bibitemStop [0]{}%
\providecommand \bibitemNoStop [0]{.\EOS\space}%
\providecommand \EOS [0]{\spacefactor3000\relax}%
\providecommand \BibitemShut  [1]{\csname bibitem#1\endcsname}%
\let\auto@bib@innerbib\@empty
\bibitem [{\citenamefont {Tom}\ \emph {et~al.}(2024)\citenamefont {Tom}, \citenamefont {Schmid}, \citenamefont {Baird}, \citenamefont {Cao}, \citenamefont {Darvish}, \citenamefont {Hao}, \citenamefont {Lo}, \citenamefont {Pablo-Garc{\'\i}a}, \citenamefont {Rajaonson}, \citenamefont {Skreta} \emph {et~al.}}]{tom2024self}%
  \BibitemOpen
  \bibfield  {author} {\bibinfo {author} {\bibfnamefont {G.}~\bibnamefont {Tom}}, \bibinfo {author} {\bibfnamefont {S.~P.}\ \bibnamefont {Schmid}}, \bibinfo {author} {\bibfnamefont {S.~G.}\ \bibnamefont {Baird}}, \bibinfo {author} {\bibfnamefont {Y.}~\bibnamefont {Cao}}, \bibinfo {author} {\bibfnamefont {K.}~\bibnamefont {Darvish}}, \bibinfo {author} {\bibfnamefont {H.}~\bibnamefont {Hao}}, \bibinfo {author} {\bibfnamefont {S.}~\bibnamefont {Lo}}, \bibinfo {author} {\bibfnamefont {S.}~\bibnamefont {Pablo-Garc{\'\i}a}}, \bibinfo {author} {\bibfnamefont {E.~M.}\ \bibnamefont {Rajaonson}}, \bibinfo {author} {\bibfnamefont {M.}~\bibnamefont {Skreta}},  \emph {et~al.},\ }\href@noop {} {\bibfield  {journal} {\bibinfo  {journal} {Chemical Reviews}\ }\textbf {\bibinfo {volume} {124}},\ \bibinfo {pages} {9633} (\bibinfo {year} {2024})}\BibitemShut {NoStop}%
\bibitem [{\citenamefont {Sim}\ \emph {et~al.}(2024)\citenamefont {Sim}, \citenamefont {Vakili}, \citenamefont {Strieth-Kalthoff}, \citenamefont {Hao}, \citenamefont {Hickman}, \citenamefont {Miret}, \citenamefont {Pablo-Garc{\'\i}a},\ and\ \citenamefont {Aspuru-Guzik}}]{sim2024chemos}%
  \BibitemOpen
  \bibfield  {author} {\bibinfo {author} {\bibfnamefont {M.}~\bibnamefont {Sim}}, \bibinfo {author} {\bibfnamefont {M.~G.}\ \bibnamefont {Vakili}}, \bibinfo {author} {\bibfnamefont {F.}~\bibnamefont {Strieth-Kalthoff}}, \bibinfo {author} {\bibfnamefont {H.}~\bibnamefont {Hao}}, \bibinfo {author} {\bibfnamefont {R.~J.}\ \bibnamefont {Hickman}}, \bibinfo {author} {\bibfnamefont {S.}~\bibnamefont {Miret}}, \bibinfo {author} {\bibfnamefont {S.}~\bibnamefont {Pablo-Garc{\'\i}a}}, \ and\ \bibinfo {author} {\bibfnamefont {A.}~\bibnamefont {Aspuru-Guzik}},\ }\href@noop {} {\bibfield  {journal} {\bibinfo  {journal} {Matter}\ }\textbf {\bibinfo {volume} {7}},\ \bibinfo {pages} {2959} (\bibinfo {year} {2024})}\BibitemShut {NoStop}%
\bibitem [{\citenamefont {Seifrid}\ \emph {et~al.}(2024)\citenamefont {Seifrid}, \citenamefont {Strieth-Kalthoff}, \citenamefont {Haddadnia}, \citenamefont {Wu}, \citenamefont {Alca}, \citenamefont {Bodo}, \citenamefont {Arellano-Rubach}, \citenamefont {Yoshikawa}, \citenamefont {Skreta}, \citenamefont {Keunen} \emph {et~al.}}]{seifrid2024chemspyd}%
  \BibitemOpen
  \bibfield  {author} {\bibinfo {author} {\bibfnamefont {M.}~\bibnamefont {Seifrid}}, \bibinfo {author} {\bibfnamefont {F.}~\bibnamefont {Strieth-Kalthoff}}, \bibinfo {author} {\bibfnamefont {M.}~\bibnamefont {Haddadnia}}, \bibinfo {author} {\bibfnamefont {T.~C.}\ \bibnamefont {Wu}}, \bibinfo {author} {\bibfnamefont {E.}~\bibnamefont {Alca}}, \bibinfo {author} {\bibfnamefont {L.}~\bibnamefont {Bodo}}, \bibinfo {author} {\bibfnamefont {S.}~\bibnamefont {Arellano-Rubach}}, \bibinfo {author} {\bibfnamefont {N.}~\bibnamefont {Yoshikawa}}, \bibinfo {author} {\bibfnamefont {M.}~\bibnamefont {Skreta}}, \bibinfo {author} {\bibfnamefont {R.}~\bibnamefont {Keunen}},  \emph {et~al.},\ }\href@noop {} {\bibfield  {journal} {\bibinfo  {journal} {Digital Discovery}\ } (\bibinfo {year} {2024})}\BibitemShut {NoStop}%
\bibitem [{\citenamefont {Xu}\ \emph {et~al.}(2023)\citenamefont {Xu}, \citenamefont {Huang}, \citenamefont {Fruitwala}, \citenamefont {Rajagopala}, \citenamefont {Naik}, \citenamefont {Nowrouzi}, \citenamefont {Santiago},\ and\ \citenamefont {Siddiqi}}]{2309.10333}%
  \BibitemOpen
  \bibfield  {author} {\bibinfo {author} {\bibfnamefont {Y.}~\bibnamefont {Xu}}, \bibinfo {author} {\bibfnamefont {G.}~\bibnamefont {Huang}}, \bibinfo {author} {\bibfnamefont {N.}~\bibnamefont {Fruitwala}}, \bibinfo {author} {\bibfnamefont {A.}~\bibnamefont {Rajagopala}}, \bibinfo {author} {\bibfnamefont {R.~K.}\ \bibnamefont {Naik}}, \bibinfo {author} {\bibfnamefont {K.}~\bibnamefont {Nowrouzi}}, \bibinfo {author} {\bibfnamefont {D.~I.}\ \bibnamefont {Santiago}}, \ and\ \bibinfo {author} {\bibfnamefont {I.}~\bibnamefont {Siddiqi}},\ }\href@noop {} {\enquote {\bibinfo {title} {Qubic 2.0: An extensible open-source qubit control system capable of mid-circuit measurement and feed-forward},}\ } (\bibinfo {year} {2023})\BibitemShut {NoStop}%
\bibitem [{\citenamefont {Stein}\ and\ \citenamefont {Gregoire}(2019)}]{Stein2019}%
  \BibitemOpen
  \bibfield  {author} {\bibinfo {author} {\bibfnamefont {H.~S.}\ \bibnamefont {Stein}}\ and\ \bibinfo {author} {\bibfnamefont {J.~M.}\ \bibnamefont {Gregoire}},\ }\href {\doibase 10.1039/c9sc03766g} {\bibfield  {journal} {\bibinfo  {journal} {Chemical Science}\ }\textbf {\bibinfo {volume} {10}},\ \bibinfo {pages} {9640–9649} (\bibinfo {year} {2019})}\BibitemShut {NoStop}%
\bibitem [{\citenamefont {Shi}\ \emph {et~al.}(2021)\citenamefont {Shi}, \citenamefont {Prieto}, \citenamefont {Zepel}, \citenamefont {Grunert},\ and\ \citenamefont {Hein}}]{AutoChem2021}%
  \BibitemOpen
  \bibfield  {author} {\bibinfo {author} {\bibfnamefont {Y.}~\bibnamefont {Shi}}, \bibinfo {author} {\bibfnamefont {P.~L.}\ \bibnamefont {Prieto}}, \bibinfo {author} {\bibfnamefont {T.}~\bibnamefont {Zepel}}, \bibinfo {author} {\bibfnamefont {S.}~\bibnamefont {Grunert}}, \ and\ \bibinfo {author} {\bibfnamefont {J.~E.}\ \bibnamefont {Hein}},\ }\href {\doibase 10.1021/acs.accounts.0c00736} {\bibfield  {journal} {\bibinfo  {journal} {Accounts of Chemical Research}\ }\textbf {\bibinfo {volume} {54}},\ \bibinfo {pages} {546} (\bibinfo {year} {2021})}\BibitemShut {NoStop}%
\bibitem [{\citenamefont {Xie}\ \emph {et~al.}(2023)\citenamefont {Xie}, \citenamefont {Sattari}, \citenamefont {Zhang},\ and\ \citenamefont {Lin}}]{XIE2023101043}%
  \BibitemOpen
  \bibfield  {author} {\bibinfo {author} {\bibfnamefont {Y.}~\bibnamefont {Xie}}, \bibinfo {author} {\bibfnamefont {K.}~\bibnamefont {Sattari}}, \bibinfo {author} {\bibfnamefont {C.}~\bibnamefont {Zhang}}, \ and\ \bibinfo {author} {\bibfnamefont {J.}~\bibnamefont {Lin}},\ }\href {\doibase https://doi.org/10.1016/j.pmatsci.2022.101043} {\bibfield  {journal} {\bibinfo  {journal} {Progress in Materials Science}\ }\textbf {\bibinfo {volume} {132}},\ \bibinfo {pages} {101043} (\bibinfo {year} {2023})}\BibitemShut {NoStop}%
\bibitem [{\citenamefont {Yu}\ \emph {et~al.}(2019)\citenamefont {Yu}, \citenamefont {Chi}, \citenamefont {Wee}, \citenamefont {Rusydi},\ and\ \citenamefont {Breese}}]{10.1063/1.5096007}%
  \BibitemOpen
  \bibfield  {author} {\bibinfo {author} {\bibfnamefont {X.~J.}\ \bibnamefont {Yu}}, \bibinfo {author} {\bibfnamefont {X.}~\bibnamefont {Chi}}, \bibinfo {author} {\bibfnamefont {A.~T.~S.}\ \bibnamefont {Wee}}, \bibinfo {author} {\bibfnamefont {A.}~\bibnamefont {Rusydi}}, \ and\ \bibinfo {author} {\bibfnamefont {M.~B.~H.}\ \bibnamefont {Breese}},\ }\href {\doibase 10.1063/1.5096007} {\bibfield  {journal} {\bibinfo  {journal} {Review of Scientific Instruments}\ }\textbf {\bibinfo {volume} {90}},\ \bibinfo {pages} {103902} (\bibinfo {year} {2019})},\ \Eprint {http://arxiv.org/abs/https://pubs.aip.org/aip/rsi/article-pdf/doi/10.1063/1.5096007/14710054/103902\_1\_online.pdf} {https://pubs.aip.org/aip/rsi/article-pdf/doi/10.1063/1.5096007/14710054/103902\_1\_online.pdf} \BibitemShut {NoStop}%
\bibitem [{\citenamefont {Seifrid}\ \emph {et~al.}(2022)\citenamefont {Seifrid}, \citenamefont {Pollice}, \citenamefont {Aguilar-Granda}, \citenamefont {Morgan~Chan}, \citenamefont {Hotta}, \citenamefont {Ser}, \citenamefont {Vestfrid}, \citenamefont {Wu},\ and\ \citenamefont {Aspuru-Guzik}}]{ChemistryAutoLab2022}%
  \BibitemOpen
  \bibfield  {author} {\bibinfo {author} {\bibfnamefont {M.}~\bibnamefont {Seifrid}}, \bibinfo {author} {\bibfnamefont {R.}~\bibnamefont {Pollice}}, \bibinfo {author} {\bibfnamefont {A.}~\bibnamefont {Aguilar-Granda}}, \bibinfo {author} {\bibfnamefont {Z.}~\bibnamefont {Morgan~Chan}}, \bibinfo {author} {\bibfnamefont {K.}~\bibnamefont {Hotta}}, \bibinfo {author} {\bibfnamefont {C.~T.}\ \bibnamefont {Ser}}, \bibinfo {author} {\bibfnamefont {J.}~\bibnamefont {Vestfrid}}, \bibinfo {author} {\bibfnamefont {T.~C.}\ \bibnamefont {Wu}}, \ and\ \bibinfo {author} {\bibfnamefont {A.}~\bibnamefont {Aspuru-Guzik}},\ }\href {\doibase 10.1021/acs.accounts.2c00220} {\bibfield  {journal} {\bibinfo  {journal} {Accounts of Chemical Research}\ }\textbf {\bibinfo {volume} {55}},\ \bibinfo {pages} {2454} (\bibinfo {year} {2022})}\BibitemShut {NoStop}%
\bibitem [{\citenamefont {Brown}\ \emph {et~al.}(2020)\citenamefont {Brown}, \citenamefont {Mann}, \citenamefont {Ryder}, \citenamefont {Subbiah}, \citenamefont {Kaplan}, \citenamefont {Dhariwal}, \citenamefont {Neelakantan}, \citenamefont {Shyam}, \citenamefont {Sastry}, \citenamefont {Askell} \emph {et~al.}}]{brown2020language}%
  \BibitemOpen
  \bibfield  {author} {\bibinfo {author} {\bibfnamefont {T.}~\bibnamefont {Brown}}, \bibinfo {author} {\bibfnamefont {B.}~\bibnamefont {Mann}}, \bibinfo {author} {\bibfnamefont {N.}~\bibnamefont {Ryder}}, \bibinfo {author} {\bibfnamefont {M.}~\bibnamefont {Subbiah}}, \bibinfo {author} {\bibfnamefont {J.~D.}\ \bibnamefont {Kaplan}}, \bibinfo {author} {\bibfnamefont {P.}~\bibnamefont {Dhariwal}}, \bibinfo {author} {\bibfnamefont {A.}~\bibnamefont {Neelakantan}}, \bibinfo {author} {\bibfnamefont {P.}~\bibnamefont {Shyam}}, \bibinfo {author} {\bibfnamefont {G.}~\bibnamefont {Sastry}}, \bibinfo {author} {\bibfnamefont {A.}~\bibnamefont {Askell}},  \emph {et~al.},\ }\href@noop {} {\bibfield  {journal} {\bibinfo  {journal} {Advances in neural information processing systems}\ }\textbf {\bibinfo {volume} {33}},\ \bibinfo {pages} {1877} (\bibinfo {year} {2020})}\BibitemShut {NoStop}%
\bibitem [{\citenamefont {OpenAI}(2023)}]{openai2023gpt4}%
  \BibitemOpen
  \bibfield  {author} {\bibinfo {author} {\bibnamefont {OpenAI}},\ }\href@noop {} {\enquote {\bibinfo {title} {Gpt-4 technical report},}\ } (\bibinfo {year} {2023}),\ \Eprint {http://arxiv.org/abs/2303.08774} {arXiv:2303.08774 [cs.CL]} \BibitemShut {NoStop}%
\bibitem [{\citenamefont {Team}\ \emph {et~al.}(2023)\citenamefont {Team}, \citenamefont {Anil}, \citenamefont {Borgeaud}, \citenamefont {Wu}, \citenamefont {Alayrac}, \citenamefont {Yu}, \citenamefont {Soricut}, \citenamefont {Schalkwyk}, \citenamefont {Dai}, \citenamefont {Hauth} \emph {et~al.}}]{team2023gemini}%
  \BibitemOpen
  \bibfield  {author} {\bibinfo {author} {\bibfnamefont {G.}~\bibnamefont {Team}}, \bibinfo {author} {\bibfnamefont {R.}~\bibnamefont {Anil}}, \bibinfo {author} {\bibfnamefont {S.}~\bibnamefont {Borgeaud}}, \bibinfo {author} {\bibfnamefont {Y.}~\bibnamefont {Wu}}, \bibinfo {author} {\bibfnamefont {J.-B.}\ \bibnamefont {Alayrac}}, \bibinfo {author} {\bibfnamefont {J.}~\bibnamefont {Yu}}, \bibinfo {author} {\bibfnamefont {R.}~\bibnamefont {Soricut}}, \bibinfo {author} {\bibfnamefont {J.}~\bibnamefont {Schalkwyk}}, \bibinfo {author} {\bibfnamefont {A.~M.}\ \bibnamefont {Dai}}, \bibinfo {author} {\bibfnamefont {A.}~\bibnamefont {Hauth}},  \emph {et~al.},\ }\href@noop {} {\bibfield  {journal} {\bibinfo  {journal} {arXiv preprint arXiv:2312.11805}\ } (\bibinfo {year} {2023})}\BibitemShut {NoStop}%
\bibitem [{\citenamefont {Dubey}\ \emph {et~al.}(2024)\citenamefont {Dubey}, \citenamefont {Jauhri}, \citenamefont {Pandey}, \citenamefont {Kadian}, \citenamefont {Al-Dahle}, \citenamefont {Letman}, \citenamefont {Mathur}, \citenamefont {Schelten}, \citenamefont {Yang}, \citenamefont {Fan} \emph {et~al.}}]{dubey2024llama}%
  \BibitemOpen
  \bibfield  {author} {\bibinfo {author} {\bibfnamefont {A.}~\bibnamefont {Dubey}}, \bibinfo {author} {\bibfnamefont {A.}~\bibnamefont {Jauhri}}, \bibinfo {author} {\bibfnamefont {A.}~\bibnamefont {Pandey}}, \bibinfo {author} {\bibfnamefont {A.}~\bibnamefont {Kadian}}, \bibinfo {author} {\bibfnamefont {A.}~\bibnamefont {Al-Dahle}}, \bibinfo {author} {\bibfnamefont {A.}~\bibnamefont {Letman}}, \bibinfo {author} {\bibfnamefont {A.}~\bibnamefont {Mathur}}, \bibinfo {author} {\bibfnamefont {A.}~\bibnamefont {Schelten}}, \bibinfo {author} {\bibfnamefont {A.}~\bibnamefont {Yang}}, \bibinfo {author} {\bibfnamefont {A.}~\bibnamefont {Fan}},  \emph {et~al.},\ }\href@noop {} {\bibfield  {journal} {\bibinfo  {journal} {arXiv preprint arXiv:2407.21783}\ } (\bibinfo {year} {2024})}\BibitemShut {NoStop}%
\bibitem [{\citenamefont {Liu}\ \emph {et~al.}(2024{\natexlab{a}})\citenamefont {Liu}, \citenamefont {Li}, \citenamefont {Wu},\ and\ \citenamefont {Lee}}]{liu2024visual}%
  \BibitemOpen
  \bibfield  {author} {\bibinfo {author} {\bibfnamefont {H.}~\bibnamefont {Liu}}, \bibinfo {author} {\bibfnamefont {C.}~\bibnamefont {Li}}, \bibinfo {author} {\bibfnamefont {Q.}~\bibnamefont {Wu}}, \ and\ \bibinfo {author} {\bibfnamefont {Y.~J.}\ \bibnamefont {Lee}},\ }\href@noop {} {\bibfield  {journal} {\bibinfo  {journal} {Advances in neural information processing systems}\ }\textbf {\bibinfo {volume} {36}} (\bibinfo {year} {2024}{\natexlab{a}})}\BibitemShut {NoStop}%
\bibitem [{\citenamefont {Liu}\ \emph {et~al.}(2024{\natexlab{b}})\citenamefont {Liu}, \citenamefont {Li}, \citenamefont {Li},\ and\ \citenamefont {Lee}}]{liu2024improved}%
  \BibitemOpen
  \bibfield  {author} {\bibinfo {author} {\bibfnamefont {H.}~\bibnamefont {Liu}}, \bibinfo {author} {\bibfnamefont {C.}~\bibnamefont {Li}}, \bibinfo {author} {\bibfnamefont {Y.}~\bibnamefont {Li}}, \ and\ \bibinfo {author} {\bibfnamefont {Y.~J.}\ \bibnamefont {Lee}},\ }in\ \href@noop {} {\emph {\bibinfo {booktitle} {Proceedings of the IEEE/CVF Conference on Computer Vision and Pattern Recognition}}}\ (\bibinfo {year} {2024})\ pp.\ \bibinfo {pages} {26296--26306}\BibitemShut {NoStop}%
\bibitem [{\citenamefont {Zhang}\ \emph {et~al.}(2024{\natexlab{a}})\citenamefont {Zhang}, \citenamefont {Aronowitz},\ and\ \citenamefont {Aspuru-Guzik}}]{zhang2024theory}%
  \BibitemOpen
  \bibfield  {author} {\bibinfo {author} {\bibfnamefont {Z.}~\bibnamefont {Zhang}}, \bibinfo {author} {\bibfnamefont {S.}~\bibnamefont {Aronowitz}}, \ and\ \bibinfo {author} {\bibfnamefont {A.}~\bibnamefont {Aspuru-Guzik}},\ }\href@noop {} {\bibfield  {journal} {\bibinfo  {journal} {arXiv preprint arXiv:2408.08463}\ } (\bibinfo {year} {2024}{\natexlab{a}})}\BibitemShut {NoStop}%
\bibitem [{\citenamefont {Li}\ \emph {et~al.}(2023{\natexlab{a}})\citenamefont {Li}, \citenamefont {Hammoud}, \citenamefont {Itani}, \citenamefont {Khizbullin},\ and\ \citenamefont {Ghanem}}]{li2023camel}%
  \BibitemOpen
  \bibfield  {author} {\bibinfo {author} {\bibfnamefont {G.}~\bibnamefont {Li}}, \bibinfo {author} {\bibfnamefont {H.}~\bibnamefont {Hammoud}}, \bibinfo {author} {\bibfnamefont {H.}~\bibnamefont {Itani}}, \bibinfo {author} {\bibfnamefont {D.}~\bibnamefont {Khizbullin}}, \ and\ \bibinfo {author} {\bibfnamefont {B.}~\bibnamefont {Ghanem}},\ }\href@noop {} {\bibfield  {journal} {\bibinfo  {journal} {Advances in Neural Information Processing Systems}\ }\textbf {\bibinfo {volume} {36}},\ \bibinfo {pages} {51991} (\bibinfo {year} {2023}{\natexlab{a}})}\BibitemShut {NoStop}%
\bibitem [{\citenamefont {Chan}\ \emph {et~al.}(2023)\citenamefont {Chan}, \citenamefont {Chen}, \citenamefont {Su}, \citenamefont {Yu}, \citenamefont {Xue}, \citenamefont {Zhang}, \citenamefont {Fu},\ and\ \citenamefont {Liu}}]{chan2023chateval}%
  \BibitemOpen
  \bibfield  {author} {\bibinfo {author} {\bibfnamefont {C.-M.}\ \bibnamefont {Chan}}, \bibinfo {author} {\bibfnamefont {W.}~\bibnamefont {Chen}}, \bibinfo {author} {\bibfnamefont {Y.}~\bibnamefont {Su}}, \bibinfo {author} {\bibfnamefont {J.}~\bibnamefont {Yu}}, \bibinfo {author} {\bibfnamefont {W.}~\bibnamefont {Xue}}, \bibinfo {author} {\bibfnamefont {S.}~\bibnamefont {Zhang}}, \bibinfo {author} {\bibfnamefont {J.}~\bibnamefont {Fu}}, \ and\ \bibinfo {author} {\bibfnamefont {Z.}~\bibnamefont {Liu}},\ }\href@noop {} {\bibfield  {journal} {\bibinfo  {journal} {arXiv preprint arXiv:2308.07201}\ } (\bibinfo {year} {2023})}\BibitemShut {NoStop}%
\bibitem [{\citenamefont {Wu}\ \emph {et~al.}(2023)\citenamefont {Wu}, \citenamefont {Bansal}, \citenamefont {Zhang}, \citenamefont {Wu}, \citenamefont {Zhang}, \citenamefont {Zhu}, \citenamefont {Li}, \citenamefont {Jiang}, \citenamefont {Zhang},\ and\ \citenamefont {Wang}}]{wu2023autogen}%
  \BibitemOpen
  \bibfield  {author} {\bibinfo {author} {\bibfnamefont {Q.}~\bibnamefont {Wu}}, \bibinfo {author} {\bibfnamefont {G.}~\bibnamefont {Bansal}}, \bibinfo {author} {\bibfnamefont {J.}~\bibnamefont {Zhang}}, \bibinfo {author} {\bibfnamefont {Y.}~\bibnamefont {Wu}}, \bibinfo {author} {\bibfnamefont {S.}~\bibnamefont {Zhang}}, \bibinfo {author} {\bibfnamefont {E.}~\bibnamefont {Zhu}}, \bibinfo {author} {\bibfnamefont {B.}~\bibnamefont {Li}}, \bibinfo {author} {\bibfnamefont {L.}~\bibnamefont {Jiang}}, \bibinfo {author} {\bibfnamefont {X.}~\bibnamefont {Zhang}}, \ and\ \bibinfo {author} {\bibfnamefont {C.}~\bibnamefont {Wang}},\ }\href@noop {} {\bibfield  {journal} {\bibinfo  {journal} {arXiv preprint arXiv:2308.08155}\ } (\bibinfo {year} {2023})}\BibitemShut {NoStop}%
\bibitem [{\citenamefont {Talebirad}\ and\ \citenamefont {Nadiri}(2023)}]{talebirad2023multi}%
  \BibitemOpen
  \bibfield  {author} {\bibinfo {author} {\bibfnamefont {Y.}~\bibnamefont {Talebirad}}\ and\ \bibinfo {author} {\bibfnamefont {A.}~\bibnamefont {Nadiri}},\ }\href@noop {} {\bibfield  {journal} {\bibinfo  {journal} {arXiv preprint arXiv:2306.03314}\ } (\bibinfo {year} {2023})}\BibitemShut {NoStop}%
\bibitem [{\citenamefont {Zhang}\ \emph {et~al.}(2024{\natexlab{b}})\citenamefont {Zhang}, \citenamefont {Yang}, \citenamefont {Bai}, \citenamefont {Wu}, \citenamefont {Li}, \citenamefont {Li},\ and\ \citenamefont {Wang}}]{zhang2024towards}%
  \BibitemOpen
  \bibfield  {author} {\bibinfo {author} {\bibfnamefont {Y.}~\bibnamefont {Zhang}}, \bibinfo {author} {\bibfnamefont {S.}~\bibnamefont {Yang}}, \bibinfo {author} {\bibfnamefont {C.}~\bibnamefont {Bai}}, \bibinfo {author} {\bibfnamefont {F.}~\bibnamefont {Wu}}, \bibinfo {author} {\bibfnamefont {X.}~\bibnamefont {Li}}, \bibinfo {author} {\bibfnamefont {X.}~\bibnamefont {Li}}, \ and\ \bibinfo {author} {\bibfnamefont {Z.}~\bibnamefont {Wang}},\ }\href@noop {} {\bibfield  {journal} {\bibinfo  {journal} {arXiv preprint arXiv:2405.14314}\ } (\bibinfo {year} {2024}{\natexlab{b}})}\BibitemShut {NoStop}%
\bibitem [{\citenamefont {Guo}\ \emph {et~al.}(2024)\citenamefont {Guo}, \citenamefont {Chen}, \citenamefont {Wang}, \citenamefont {Chang}, \citenamefont {Pei}, \citenamefont {Chawla}, \citenamefont {Wiest},\ and\ \citenamefont {Zhang}}]{guo2024large}%
  \BibitemOpen
  \bibfield  {author} {\bibinfo {author} {\bibfnamefont {T.}~\bibnamefont {Guo}}, \bibinfo {author} {\bibfnamefont {X.}~\bibnamefont {Chen}}, \bibinfo {author} {\bibfnamefont {Y.}~\bibnamefont {Wang}}, \bibinfo {author} {\bibfnamefont {R.}~\bibnamefont {Chang}}, \bibinfo {author} {\bibfnamefont {S.}~\bibnamefont {Pei}}, \bibinfo {author} {\bibfnamefont {N.~V.}\ \bibnamefont {Chawla}}, \bibinfo {author} {\bibfnamefont {O.}~\bibnamefont {Wiest}}, \ and\ \bibinfo {author} {\bibfnamefont {X.}~\bibnamefont {Zhang}},\ }\href@noop {} {\bibfield  {journal} {\bibinfo  {journal} {arXiv preprint arXiv:2402.01680}\ } (\bibinfo {year} {2024})}\BibitemShut {NoStop}%
\bibitem [{\citenamefont {Li}\ \emph {et~al.}(2024{\natexlab{a}})\citenamefont {Li}, \citenamefont {Wang}, \citenamefont {Zeng}, \citenamefont {Wu},\ and\ \citenamefont {Yang}}]{li2024survey}%
  \BibitemOpen
  \bibfield  {author} {\bibinfo {author} {\bibfnamefont {X.}~\bibnamefont {Li}}, \bibinfo {author} {\bibfnamefont {S.}~\bibnamefont {Wang}}, \bibinfo {author} {\bibfnamefont {S.}~\bibnamefont {Zeng}}, \bibinfo {author} {\bibfnamefont {Y.}~\bibnamefont {Wu}}, \ and\ \bibinfo {author} {\bibfnamefont {Y.}~\bibnamefont {Yang}},\ }\href@noop {} {\bibfield  {journal} {\bibinfo  {journal} {Vicinagearth}\ }\textbf {\bibinfo {volume} {1}},\ \bibinfo {pages} {9} (\bibinfo {year} {2024}{\natexlab{a}})}\BibitemShut {NoStop}%
\bibitem [{\citenamefont {Kaplan}\ \emph {et~al.}(2020)\citenamefont {Kaplan}, \citenamefont {McCandlish}, \citenamefont {Henighan}, \citenamefont {Brown}, \citenamefont {Chess}, \citenamefont {Child}, \citenamefont {Gray}, \citenamefont {Radford}, \citenamefont {Wu},\ and\ \citenamefont {Amodei}}]{kaplan2020scaling}%
  \BibitemOpen
  \bibfield  {author} {\bibinfo {author} {\bibfnamefont {J.}~\bibnamefont {Kaplan}}, \bibinfo {author} {\bibfnamefont {S.}~\bibnamefont {McCandlish}}, \bibinfo {author} {\bibfnamefont {T.}~\bibnamefont {Henighan}}, \bibinfo {author} {\bibfnamefont {T.~B.}\ \bibnamefont {Brown}}, \bibinfo {author} {\bibfnamefont {B.}~\bibnamefont {Chess}}, \bibinfo {author} {\bibfnamefont {R.}~\bibnamefont {Child}}, \bibinfo {author} {\bibfnamefont {S.}~\bibnamefont {Gray}}, \bibinfo {author} {\bibfnamefont {A.}~\bibnamefont {Radford}}, \bibinfo {author} {\bibfnamefont {J.}~\bibnamefont {Wu}}, \ and\ \bibinfo {author} {\bibfnamefont {D.}~\bibnamefont {Amodei}},\ }\href@noop {} {\bibfield  {journal} {\bibinfo  {journal} {arXiv preprint arXiv:2001.08361}\ } (\bibinfo {year} {2020})}\BibitemShut {NoStop}%
\bibitem [{\citenamefont {Li}\ \emph {et~al.}(2024{\natexlab{b}})\citenamefont {Li}, \citenamefont {Zhang}, \citenamefont {Yao}, \citenamefont {Wang}, \citenamefont {Chen},\ and\ \citenamefont {Chen}}]{li2024unveiling}%
  \BibitemOpen
  \bibfield  {author} {\bibinfo {author} {\bibfnamefont {Z.}~\bibnamefont {Li}}, \bibinfo {author} {\bibfnamefont {N.}~\bibnamefont {Zhang}}, \bibinfo {author} {\bibfnamefont {Y.}~\bibnamefont {Yao}}, \bibinfo {author} {\bibfnamefont {M.}~\bibnamefont {Wang}}, \bibinfo {author} {\bibfnamefont {X.}~\bibnamefont {Chen}}, \ and\ \bibinfo {author} {\bibfnamefont {H.}~\bibnamefont {Chen}},\ }in\ \href {https://openreview.net/forum?id=fNktD3ib16} {\emph {\bibinfo {booktitle} {The Twelfth International Conference on Learning Representations}}}\ (\bibinfo {year} {2024})\BibitemShut {NoStop}%
\bibitem [{\citenamefont {Hoelscher-Obermaier}\ \emph {et~al.}(2023)\citenamefont {Hoelscher-Obermaier}, \citenamefont {Persson}, \citenamefont {Kran}, \citenamefont {Konstas},\ and\ \citenamefont {Barez}}]{hoelscher2023detecting}%
  \BibitemOpen
  \bibfield  {author} {\bibinfo {author} {\bibfnamefont {J.}~\bibnamefont {Hoelscher-Obermaier}}, \bibinfo {author} {\bibfnamefont {J.}~\bibnamefont {Persson}}, \bibinfo {author} {\bibfnamefont {E.}~\bibnamefont {Kran}}, \bibinfo {author} {\bibfnamefont {I.}~\bibnamefont {Konstas}}, \ and\ \bibinfo {author} {\bibfnamefont {F.}~\bibnamefont {Barez}},\ }\href@noop {} {\bibfield  {journal} {\bibinfo  {journal} {arXiv preprint arXiv:2305.17553}\ } (\bibinfo {year} {2023})}\BibitemShut {NoStop}%
\bibitem [{\citenamefont {Zhang}\ \emph {et~al.}(2024{\natexlab{c}})\citenamefont {Zhang}, \citenamefont {Yao}, \citenamefont {Tian}, \citenamefont {Wang}, \citenamefont {Deng}, \citenamefont {Wang}, \citenamefont {Xi}, \citenamefont {Mao}, \citenamefont {Zhang}, \citenamefont {Ni} \emph {et~al.}}]{zhang2024comprehensive}%
  \BibitemOpen
  \bibfield  {author} {\bibinfo {author} {\bibfnamefont {N.}~\bibnamefont {Zhang}}, \bibinfo {author} {\bibfnamefont {Y.}~\bibnamefont {Yao}}, \bibinfo {author} {\bibfnamefont {B.}~\bibnamefont {Tian}}, \bibinfo {author} {\bibfnamefont {P.}~\bibnamefont {Wang}}, \bibinfo {author} {\bibfnamefont {S.}~\bibnamefont {Deng}}, \bibinfo {author} {\bibfnamefont {M.}~\bibnamefont {Wang}}, \bibinfo {author} {\bibfnamefont {Z.}~\bibnamefont {Xi}}, \bibinfo {author} {\bibfnamefont {S.}~\bibnamefont {Mao}}, \bibinfo {author} {\bibfnamefont {J.}~\bibnamefont {Zhang}}, \bibinfo {author} {\bibfnamefont {Y.}~\bibnamefont {Ni}},  \emph {et~al.},\ }\href@noop {} {\bibfield  {journal} {\bibinfo  {journal} {arXiv preprint arXiv:2401.01286}\ } (\bibinfo {year} {2024}{\natexlab{c}})}\BibitemShut {NoStop}%
\bibitem [{\citenamefont {Lewis}\ \emph {et~al.}(2020)\citenamefont {Lewis}, \citenamefont {Perez}, \citenamefont {Piktus}, \citenamefont {Petroni}, \citenamefont {Karpukhin}, \citenamefont {Goyal}, \citenamefont {K{\"u}ttler}, \citenamefont {Lewis}, \citenamefont {Yih}, \citenamefont {Rockt{\"a}schel} \emph {et~al.}}]{lewis2020retrieval}%
  \BibitemOpen
  \bibfield  {author} {\bibinfo {author} {\bibfnamefont {P.}~\bibnamefont {Lewis}}, \bibinfo {author} {\bibfnamefont {E.}~\bibnamefont {Perez}}, \bibinfo {author} {\bibfnamefont {A.}~\bibnamefont {Piktus}}, \bibinfo {author} {\bibfnamefont {F.}~\bibnamefont {Petroni}}, \bibinfo {author} {\bibfnamefont {V.}~\bibnamefont {Karpukhin}}, \bibinfo {author} {\bibfnamefont {N.}~\bibnamefont {Goyal}}, \bibinfo {author} {\bibfnamefont {H.}~\bibnamefont {K{\"u}ttler}}, \bibinfo {author} {\bibfnamefont {M.}~\bibnamefont {Lewis}}, \bibinfo {author} {\bibfnamefont {W.-t.}\ \bibnamefont {Yih}}, \bibinfo {author} {\bibfnamefont {T.}~\bibnamefont {Rockt{\"a}schel}},  \emph {et~al.},\ }\href@noop {} {\bibfield  {journal} {\bibinfo  {journal} {Advances in Neural Information Processing Systems}\ }\textbf {\bibinfo {volume} {33}},\ \bibinfo {pages} {9459} (\bibinfo {year} {2020})}\BibitemShut {NoStop}%
\bibitem [{\citenamefont {Gao}\ \emph {et~al.}(2023)\citenamefont {Gao}, \citenamefont {Xiong}, \citenamefont {Gao}, \citenamefont {Jia}, \citenamefont {Pan}, \citenamefont {Bi}, \citenamefont {Dai}, \citenamefont {Sun},\ and\ \citenamefont {Wang}}]{gao2023retrieval}%
  \BibitemOpen
  \bibfield  {author} {\bibinfo {author} {\bibfnamefont {Y.}~\bibnamefont {Gao}}, \bibinfo {author} {\bibfnamefont {Y.}~\bibnamefont {Xiong}}, \bibinfo {author} {\bibfnamefont {X.}~\bibnamefont {Gao}}, \bibinfo {author} {\bibfnamefont {K.}~\bibnamefont {Jia}}, \bibinfo {author} {\bibfnamefont {J.}~\bibnamefont {Pan}}, \bibinfo {author} {\bibfnamefont {Y.}~\bibnamefont {Bi}}, \bibinfo {author} {\bibfnamefont {Y.}~\bibnamefont {Dai}}, \bibinfo {author} {\bibfnamefont {J.}~\bibnamefont {Sun}}, \ and\ \bibinfo {author} {\bibfnamefont {H.}~\bibnamefont {Wang}},\ }\href@noop {} {\bibfield  {journal} {\bibinfo  {journal} {arXiv preprint arXiv:2312.10997}\ } (\bibinfo {year} {2023})}\BibitemShut {NoStop}%
\bibitem [{\citenamefont {Karpukhin}\ \emph {et~al.}(2020)\citenamefont {Karpukhin}, \citenamefont {O{\u{g}}uz}, \citenamefont {Min}, \citenamefont {Lewis}, \citenamefont {Wu}, \citenamefont {Edunov}, \citenamefont {Chen},\ and\ \citenamefont {Yih}}]{karpukhin2020dense}%
  \BibitemOpen
  \bibfield  {author} {\bibinfo {author} {\bibfnamefont {V.}~\bibnamefont {Karpukhin}}, \bibinfo {author} {\bibfnamefont {B.}~\bibnamefont {O{\u{g}}uz}}, \bibinfo {author} {\bibfnamefont {S.}~\bibnamefont {Min}}, \bibinfo {author} {\bibfnamefont {P.}~\bibnamefont {Lewis}}, \bibinfo {author} {\bibfnamefont {L.}~\bibnamefont {Wu}}, \bibinfo {author} {\bibfnamefont {S.}~\bibnamefont {Edunov}}, \bibinfo {author} {\bibfnamefont {D.}~\bibnamefont {Chen}}, \ and\ \bibinfo {author} {\bibfnamefont {W.-t.}\ \bibnamefont {Yih}},\ }\href@noop {} {\bibfield  {journal} {\bibinfo  {journal} {arXiv preprint arXiv:2004.04906}\ } (\bibinfo {year} {2020})}\BibitemShut {NoStop}%
\bibitem [{\citenamefont {Franklin}\ and\ \citenamefont {Perovic}(1998)}]{franklin1998experiment}%
  \BibitemOpen
  \bibfield  {author} {\bibinfo {author} {\bibfnamefont {A.}~\bibnamefont {Franklin}}\ and\ \bibinfo {author} {\bibfnamefont {S.}~\bibnamefont {Perovic}},\ }\href@noop {} {\  (\bibinfo {year} {1998})}\BibitemShut {NoStop}%
\bibitem [{\citenamefont {Boiko}\ \emph {et~al.}(2023)\citenamefont {Boiko}, \citenamefont {MacKnight}, \citenamefont {Kline},\ and\ \citenamefont {Gomes}}]{boiko2023autonomous}%
  \BibitemOpen
  \bibfield  {author} {\bibinfo {author} {\bibfnamefont {D.~A.}\ \bibnamefont {Boiko}}, \bibinfo {author} {\bibfnamefont {R.}~\bibnamefont {MacKnight}}, \bibinfo {author} {\bibfnamefont {B.}~\bibnamefont {Kline}}, \ and\ \bibinfo {author} {\bibfnamefont {G.}~\bibnamefont {Gomes}},\ }\href@noop {} {\bibfield  {journal} {\bibinfo  {journal} {Nature}\ }\textbf {\bibinfo {volume} {624}},\ \bibinfo {pages} {570} (\bibinfo {year} {2023})}\BibitemShut {NoStop}%
\bibitem [{\citenamefont {M.~Bran}\ \emph {et~al.}(2024)\citenamefont {M.~Bran}, \citenamefont {Cox}, \citenamefont {Schilter}, \citenamefont {Baldassari}, \citenamefont {White},\ and\ \citenamefont {Schwaller}}]{m2024augmenting}%
  \BibitemOpen
  \bibfield  {author} {\bibinfo {author} {\bibfnamefont {A.}~\bibnamefont {M.~Bran}}, \bibinfo {author} {\bibfnamefont {S.}~\bibnamefont {Cox}}, \bibinfo {author} {\bibfnamefont {O.}~\bibnamefont {Schilter}}, \bibinfo {author} {\bibfnamefont {C.}~\bibnamefont {Baldassari}}, \bibinfo {author} {\bibfnamefont {A.~D.}\ \bibnamefont {White}}, \ and\ \bibinfo {author} {\bibfnamefont {P.}~\bibnamefont {Schwaller}},\ }\href@noop {} {\bibfield  {journal} {\bibinfo  {journal} {Nature Machine Intelligence}\ ,\ \bibinfo {pages} {1}} (\bibinfo {year} {2024})}\BibitemShut {NoStop}%
\bibitem [{\citenamefont {Darvish}\ \emph {et~al.}(2024)\citenamefont {Darvish}, \citenamefont {Skreta}, \citenamefont {Zhao}, \citenamefont {Yoshikawa}, \citenamefont {Som}, \citenamefont {Bogdanovic}, \citenamefont {Cao}, \citenamefont {Hao}, \citenamefont {Xu}, \citenamefont {Aspuru-Guzik} \emph {et~al.}}]{darvish2024organa}%
  \BibitemOpen
  \bibfield  {author} {\bibinfo {author} {\bibfnamefont {K.}~\bibnamefont {Darvish}}, \bibinfo {author} {\bibfnamefont {M.}~\bibnamefont {Skreta}}, \bibinfo {author} {\bibfnamefont {Y.}~\bibnamefont {Zhao}}, \bibinfo {author} {\bibfnamefont {N.}~\bibnamefont {Yoshikawa}}, \bibinfo {author} {\bibfnamefont {S.}~\bibnamefont {Som}}, \bibinfo {author} {\bibfnamefont {M.}~\bibnamefont {Bogdanovic}}, \bibinfo {author} {\bibfnamefont {Y.}~\bibnamefont {Cao}}, \bibinfo {author} {\bibfnamefont {H.}~\bibnamefont {Hao}}, \bibinfo {author} {\bibfnamefont {H.}~\bibnamefont {Xu}}, \bibinfo {author} {\bibfnamefont {A.}~\bibnamefont {Aspuru-Guzik}},  \emph {et~al.},\ }\href@noop {} {\bibfield  {journal} {\bibinfo  {journal} {arXiv preprint arXiv:2401.06949}\ } (\bibinfo {year} {2024})}\BibitemShut {NoStop}%
\bibitem [{\citenamefont {Yoshikawa}\ \emph {et~al.}(2023)\citenamefont {Yoshikawa}, \citenamefont {Skreta}, \citenamefont {Darvish}, \citenamefont {Arellano-Rubach}, \citenamefont {Ji}, \citenamefont {Bj{\o}rn~Kristensen}, \citenamefont {Li}, \citenamefont {Zhao}, \citenamefont {Xu}, \citenamefont {Kuramshin} \emph {et~al.}}]{yoshikawa2023large}%
  \BibitemOpen
  \bibfield  {author} {\bibinfo {author} {\bibfnamefont {N.}~\bibnamefont {Yoshikawa}}, \bibinfo {author} {\bibfnamefont {M.}~\bibnamefont {Skreta}}, \bibinfo {author} {\bibfnamefont {K.}~\bibnamefont {Darvish}}, \bibinfo {author} {\bibfnamefont {S.}~\bibnamefont {Arellano-Rubach}}, \bibinfo {author} {\bibfnamefont {Z.}~\bibnamefont {Ji}}, \bibinfo {author} {\bibfnamefont {L.}~\bibnamefont {Bj{\o}rn~Kristensen}}, \bibinfo {author} {\bibfnamefont {A.~Z.}\ \bibnamefont {Li}}, \bibinfo {author} {\bibfnamefont {Y.}~\bibnamefont {Zhao}}, \bibinfo {author} {\bibfnamefont {H.}~\bibnamefont {Xu}}, \bibinfo {author} {\bibfnamefont {A.}~\bibnamefont {Kuramshin}},  \emph {et~al.},\ }\href@noop {} {\bibfield  {journal} {\bibinfo  {journal} {Autonomous Robots}\ }\textbf {\bibinfo {volume} {47}},\ \bibinfo {pages} {1057} (\bibinfo {year} {2023})}\BibitemShut {NoStop}%
\bibitem [{\citenamefont {Liu}\ \emph {et~al.}(2023)\citenamefont {Liu}, \citenamefont {Lin}, \citenamefont {Hewitt}, \citenamefont {Paranjape}, \citenamefont {Bevilacqua}, \citenamefont {Petroni},\ and\ \citenamefont {Liang}}]{liu2023lost}%
  \BibitemOpen
  \bibfield  {author} {\bibinfo {author} {\bibfnamefont {N.~F.}\ \bibnamefont {Liu}}, \bibinfo {author} {\bibfnamefont {K.}~\bibnamefont {Lin}}, \bibinfo {author} {\bibfnamefont {J.}~\bibnamefont {Hewitt}}, \bibinfo {author} {\bibfnamefont {A.}~\bibnamefont {Paranjape}}, \bibinfo {author} {\bibfnamefont {M.}~\bibnamefont {Bevilacqua}}, \bibinfo {author} {\bibfnamefont {F.}~\bibnamefont {Petroni}}, \ and\ \bibinfo {author} {\bibfnamefont {P.}~\bibnamefont {Liang}},\ }\href@noop {} {\bibfield  {journal} {\bibinfo  {journal} {arXiv preprint arXiv:2307.03172}\ } (\bibinfo {year} {2023})}\BibitemShut {NoStop}%
\bibitem [{\citenamefont {Li}\ \emph {et~al.}(2024{\natexlab{c}})\citenamefont {Li}, \citenamefont {Zhang}, \citenamefont {Do}, \citenamefont {Yue},\ and\ \citenamefont {Chen}}]{li2024long}%
  \BibitemOpen
  \bibfield  {author} {\bibinfo {author} {\bibfnamefont {T.}~\bibnamefont {Li}}, \bibinfo {author} {\bibfnamefont {G.}~\bibnamefont {Zhang}}, \bibinfo {author} {\bibfnamefont {Q.~D.}\ \bibnamefont {Do}}, \bibinfo {author} {\bibfnamefont {X.}~\bibnamefont {Yue}}, \ and\ \bibinfo {author} {\bibfnamefont {W.}~\bibnamefont {Chen}},\ }\href@noop {} {\bibfield  {journal} {\bibinfo  {journal} {arXiv preprint arXiv:2404.02060}\ } (\bibinfo {year} {2024}{\natexlab{c}})}\BibitemShut {NoStop}%
\bibitem [{\citenamefont {Li}\ \emph {et~al.}(2023{\natexlab{b}})\citenamefont {Li}, \citenamefont {Shao}, \citenamefont {Xie}, \citenamefont {Sheng}, \citenamefont {Zheng}, \citenamefont {Gonzalez}, \citenamefont {Stoica}, \citenamefont {Ma},\ and\ \citenamefont {Zhang}}]{li2023long}%
  \BibitemOpen
  \bibfield  {author} {\bibinfo {author} {\bibfnamefont {D.}~\bibnamefont {Li}}, \bibinfo {author} {\bibfnamefont {R.}~\bibnamefont {Shao}}, \bibinfo {author} {\bibfnamefont {A.}~\bibnamefont {Xie}}, \bibinfo {author} {\bibfnamefont {Y.}~\bibnamefont {Sheng}}, \bibinfo {author} {\bibfnamefont {L.}~\bibnamefont {Zheng}}, \bibinfo {author} {\bibfnamefont {J.}~\bibnamefont {Gonzalez}}, \bibinfo {author} {\bibfnamefont {I.}~\bibnamefont {Stoica}}, \bibinfo {author} {\bibfnamefont {X.}~\bibnamefont {Ma}}, \ and\ \bibinfo {author} {\bibfnamefont {H.}~\bibnamefont {Zhang}},\ }in\ \href@noop {} {\emph {\bibinfo {booktitle} {NeurIPS 2023 Workshop on Instruction Tuning and Instruction Following}}}\ (\bibinfo {year} {2023})\BibitemShut {NoStop}%
\bibitem [{\citenamefont {Acharya}\ \emph {et~al.}(2024)\citenamefont {Acharya}, \citenamefont {Aghababaie-Beni}, \citenamefont {Aleiner}, \citenamefont {Andersen}, \citenamefont {Ansmann}, \citenamefont {Arute}, \citenamefont {Arya}, \citenamefont {Asfaw}, \citenamefont {Astrakhantsev}, \citenamefont {Atalaya} \emph {et~al.}}]{acharya2024quantum}%
  \BibitemOpen
  \bibfield  {author} {\bibinfo {author} {\bibfnamefont {R.}~\bibnamefont {Acharya}}, \bibinfo {author} {\bibfnamefont {L.}~\bibnamefont {Aghababaie-Beni}}, \bibinfo {author} {\bibfnamefont {I.}~\bibnamefont {Aleiner}}, \bibinfo {author} {\bibfnamefont {T.~I.}\ \bibnamefont {Andersen}}, \bibinfo {author} {\bibfnamefont {M.}~\bibnamefont {Ansmann}}, \bibinfo {author} {\bibfnamefont {F.}~\bibnamefont {Arute}}, \bibinfo {author} {\bibfnamefont {K.}~\bibnamefont {Arya}}, \bibinfo {author} {\bibfnamefont {A.}~\bibnamefont {Asfaw}}, \bibinfo {author} {\bibfnamefont {N.}~\bibnamefont {Astrakhantsev}}, \bibinfo {author} {\bibfnamefont {J.}~\bibnamefont {Atalaya}},  \emph {et~al.},\ }\href@noop {} {\bibfield  {journal} {\bibinfo  {journal} {arXiv preprint arXiv:2408.13687}\ } (\bibinfo {year} {2024})}\BibitemShut {NoStop}%
\bibitem [{\citenamefont {Reichardt}\ \emph {et~al.}(2024)\citenamefont {Reichardt}, \citenamefont {Aasen}, \citenamefont {Chao}, \citenamefont {Chernoguzov}, \citenamefont {van Dam}, \citenamefont {Gaebler}, \citenamefont {Gresh}, \citenamefont {Lucchetti}, \citenamefont {Mills}, \citenamefont {Moses} \emph {et~al.}}]{reichardt2024demonstration}%
  \BibitemOpen
  \bibfield  {author} {\bibinfo {author} {\bibfnamefont {B.~W.}\ \bibnamefont {Reichardt}}, \bibinfo {author} {\bibfnamefont {D.}~\bibnamefont {Aasen}}, \bibinfo {author} {\bibfnamefont {R.}~\bibnamefont {Chao}}, \bibinfo {author} {\bibfnamefont {A.}~\bibnamefont {Chernoguzov}}, \bibinfo {author} {\bibfnamefont {W.}~\bibnamefont {van Dam}}, \bibinfo {author} {\bibfnamefont {J.~P.}\ \bibnamefont {Gaebler}}, \bibinfo {author} {\bibfnamefont {D.}~\bibnamefont {Gresh}}, \bibinfo {author} {\bibfnamefont {D.}~\bibnamefont {Lucchetti}}, \bibinfo {author} {\bibfnamefont {M.}~\bibnamefont {Mills}}, \bibinfo {author} {\bibfnamefont {S.~A.}\ \bibnamefont {Moses}},  \emph {et~al.},\ }\href@noop {} {\bibfield  {journal} {\bibinfo  {journal} {arXiv preprint arXiv:2409.04628}\ } (\bibinfo {year} {2024})}\BibitemShut {NoStop}%
\bibitem [{\citenamefont {Putterman}\ \emph {et~al.}(2024)\citenamefont {Putterman}, \citenamefont {Noh}, \citenamefont {Hann}, \citenamefont {MacCabe}, \citenamefont {Aghaeimeibodi}, \citenamefont {Patel}, \citenamefont {Lee}, \citenamefont {Jones}, \citenamefont {Moradinejad}, \citenamefont {Rodriguez} \emph {et~al.}}]{putterman2024hardware}%
  \BibitemOpen
  \bibfield  {author} {\bibinfo {author} {\bibfnamefont {H.}~\bibnamefont {Putterman}}, \bibinfo {author} {\bibfnamefont {K.}~\bibnamefont {Noh}}, \bibinfo {author} {\bibfnamefont {C.~T.}\ \bibnamefont {Hann}}, \bibinfo {author} {\bibfnamefont {G.~S.}\ \bibnamefont {MacCabe}}, \bibinfo {author} {\bibfnamefont {S.}~\bibnamefont {Aghaeimeibodi}}, \bibinfo {author} {\bibfnamefont {R.~N.}\ \bibnamefont {Patel}}, \bibinfo {author} {\bibfnamefont {M.}~\bibnamefont {Lee}}, \bibinfo {author} {\bibfnamefont {W.~M.}\ \bibnamefont {Jones}}, \bibinfo {author} {\bibfnamefont {H.}~\bibnamefont {Moradinejad}}, \bibinfo {author} {\bibfnamefont {R.}~\bibnamefont {Rodriguez}},  \emph {et~al.},\ }\href@noop {} {\bibfield  {journal} {\bibinfo  {journal} {arXiv preprint arXiv:2409.13025}\ } (\bibinfo {year} {2024})}\BibitemShut {NoStop}%
\bibitem [{\citenamefont {Brock}\ \emph {et~al.}(2024)\citenamefont {Brock}, \citenamefont {Singh}, \citenamefont {Eickbusch}, \citenamefont {Sivak}, \citenamefont {Ding}, \citenamefont {Frunzio}, \citenamefont {Girvin},\ and\ \citenamefont {Devoret}}]{brock2024quantum}%
  \BibitemOpen
  \bibfield  {author} {\bibinfo {author} {\bibfnamefont {B.~L.}\ \bibnamefont {Brock}}, \bibinfo {author} {\bibfnamefont {S.}~\bibnamefont {Singh}}, \bibinfo {author} {\bibfnamefont {A.}~\bibnamefont {Eickbusch}}, \bibinfo {author} {\bibfnamefont {V.~V.}\ \bibnamefont {Sivak}}, \bibinfo {author} {\bibfnamefont {A.~Z.}\ \bibnamefont {Ding}}, \bibinfo {author} {\bibfnamefont {L.}~\bibnamefont {Frunzio}}, \bibinfo {author} {\bibfnamefont {S.~M.}\ \bibnamefont {Girvin}}, \ and\ \bibinfo {author} {\bibfnamefont {M.~H.}\ \bibnamefont {Devoret}},\ }\href@noop {} {\bibfield  {journal} {\bibinfo  {journal} {arXiv preprint arXiv:2409.15065}\ } (\bibinfo {year} {2024})}\BibitemShut {NoStop}%
\bibitem [{\citenamefont {Mohseni}\ \emph {et~al.}(2024)\citenamefont {Mohseni}, \citenamefont {Scherer}, \citenamefont {Johnson}, \citenamefont {Wertheim}, \citenamefont {Otten}, \citenamefont {Aadit}, \citenamefont {Bresniker}, \citenamefont {Camsari}, \citenamefont {Chapman}, \citenamefont {Chatterjee}, \citenamefont {Dagnew}, \citenamefont {Esposito}, \citenamefont {Fahim}, \citenamefont {Fiorentino}, \citenamefont {Khalid}, \citenamefont {Kong}, \citenamefont {Kulchytskyy}, \citenamefont {Li}, \citenamefont {Lott}, \citenamefont {Markov}, \citenamefont {McDermott}, \citenamefont {Pedretti}, \citenamefont {Gajjar}, \citenamefont {Silva}, \citenamefont {Sorebo}, \citenamefont {Spentzouris}, \citenamefont {Steiner}, \citenamefont {Torosov}, \citenamefont {Venturelli}, \citenamefont {Visser}, \citenamefont {Webb}, \citenamefont {Zhan}, \citenamefont {Cohen}, \citenamefont {Ronagh}, \citenamefont {Ho}, \citenamefont {Beausoleil},\ and\ \citenamefont {Martinis}}]{2411.10406}%
  \BibitemOpen
  \bibfield  {author} {\bibinfo {author} {\bibfnamefont {M.}~\bibnamefont {Mohseni}}, \bibinfo {author} {\bibfnamefont {A.}~\bibnamefont {Scherer}}, \bibinfo {author} {\bibfnamefont {K.~G.}\ \bibnamefont {Johnson}}, \bibinfo {author} {\bibfnamefont {O.}~\bibnamefont {Wertheim}}, \bibinfo {author} {\bibfnamefont {M.}~\bibnamefont {Otten}}, \bibinfo {author} {\bibfnamefont {N.~A.}\ \bibnamefont {Aadit}}, \bibinfo {author} {\bibfnamefont {K.~M.}\ \bibnamefont {Bresniker}}, \bibinfo {author} {\bibfnamefont {K.~Y.}\ \bibnamefont {Camsari}}, \bibinfo {author} {\bibfnamefont {B.}~\bibnamefont {Chapman}}, \bibinfo {author} {\bibfnamefont {S.}~\bibnamefont {Chatterjee}}, \bibinfo {author} {\bibfnamefont {G.~A.}\ \bibnamefont {Dagnew}}, \bibinfo {author} {\bibfnamefont {A.}~\bibnamefont {Esposito}}, \bibinfo {author} {\bibfnamefont {F.}~\bibnamefont {Fahim}}, \bibinfo {author} {\bibfnamefont {M.}~\bibnamefont {Fiorentino}}, \bibinfo {author} {\bibfnamefont {A.}~\bibnamefont {Khalid}}, \bibinfo {author} {\bibfnamefont
  {X.}~\bibnamefont {Kong}}, \bibinfo {author} {\bibfnamefont {B.}~\bibnamefont {Kulchytskyy}}, \bibinfo {author} {\bibfnamefont {R.}~\bibnamefont {Li}}, \bibinfo {author} {\bibfnamefont {P.~A.}\ \bibnamefont {Lott}}, \bibinfo {author} {\bibfnamefont {I.~L.}\ \bibnamefont {Markov}}, \bibinfo {author} {\bibfnamefont {R.~F.}\ \bibnamefont {McDermott}}, \bibinfo {author} {\bibfnamefont {G.}~\bibnamefont {Pedretti}}, \bibinfo {author} {\bibfnamefont {A.}~\bibnamefont {Gajjar}}, \bibinfo {author} {\bibfnamefont {A.}~\bibnamefont {Silva}}, \bibinfo {author} {\bibfnamefont {J.}~\bibnamefont {Sorebo}}, \bibinfo {author} {\bibfnamefont {P.}~\bibnamefont {Spentzouris}}, \bibinfo {author} {\bibfnamefont {Z.}~\bibnamefont {Steiner}}, \bibinfo {author} {\bibfnamefont {B.}~\bibnamefont {Torosov}}, \bibinfo {author} {\bibfnamefont {D.}~\bibnamefont {Venturelli}}, \bibinfo {author} {\bibfnamefont {R.~J.}\ \bibnamefont {Visser}}, \bibinfo {author} {\bibfnamefont {Z.}~\bibnamefont {Webb}}, \bibinfo {author} {\bibfnamefont
  {X.}~\bibnamefont {Zhan}}, \bibinfo {author} {\bibfnamefont {Y.}~\bibnamefont {Cohen}}, \bibinfo {author} {\bibfnamefont {P.}~\bibnamefont {Ronagh}}, \bibinfo {author} {\bibfnamefont {A.}~\bibnamefont {Ho}}, \bibinfo {author} {\bibfnamefont {R.~G.}\ \bibnamefont {Beausoleil}}, \ and\ \bibinfo {author} {\bibfnamefont {J.~M.}\ \bibnamefont {Martinis}},\ }\href@noop {} {\enquote {\bibinfo {title} {How to build a quantum supercomputer: Scaling challenges and opportunities},}\ } (\bibinfo {year} {2024}),\ \Eprint {http://arxiv.org/abs/arXiv:2411.10406} {arXiv:2411.10406} \BibitemShut {NoStop}%
\bibitem [{\citenamefont {Krenn}\ \emph {et~al.}(2022)\citenamefont {Krenn}, \citenamefont {Pollice}, \citenamefont {Guo}, \citenamefont {Aldeghi}, \citenamefont {Cervera-Lierta}, \citenamefont {Friederich}, \citenamefont {dos Passos~Gomes}, \citenamefont {H{\"a}se}, \citenamefont {Jinich}, \citenamefont {Nigam} \emph {et~al.}}]{krenn2022scientific}%
  \BibitemOpen
  \bibfield  {author} {\bibinfo {author} {\bibfnamefont {M.}~\bibnamefont {Krenn}}, \bibinfo {author} {\bibfnamefont {R.}~\bibnamefont {Pollice}}, \bibinfo {author} {\bibfnamefont {S.~Y.}\ \bibnamefont {Guo}}, \bibinfo {author} {\bibfnamefont {M.}~\bibnamefont {Aldeghi}}, \bibinfo {author} {\bibfnamefont {A.}~\bibnamefont {Cervera-Lierta}}, \bibinfo {author} {\bibfnamefont {P.}~\bibnamefont {Friederich}}, \bibinfo {author} {\bibfnamefont {G.}~\bibnamefont {dos Passos~Gomes}}, \bibinfo {author} {\bibfnamefont {F.}~\bibnamefont {H{\"a}se}}, \bibinfo {author} {\bibfnamefont {A.}~\bibnamefont {Jinich}}, \bibinfo {author} {\bibfnamefont {A.}~\bibnamefont {Nigam}},  \emph {et~al.},\ }\href@noop {} {\bibfield  {journal} {\bibinfo  {journal} {Nature Reviews Physics}\ }\textbf {\bibinfo {volume} {4}},\ \bibinfo {pages} {761} (\bibinfo {year} {2022})}\BibitemShut {NoStop}%
\bibitem [{\citenamefont {Yao}\ \emph {et~al.}(2024)\citenamefont {Yao}, \citenamefont {Shinn}, \citenamefont {Razavi},\ and\ \citenamefont {Narasimhan}}]{yao2024tau}%
  \BibitemOpen
  \bibfield  {author} {\bibinfo {author} {\bibfnamefont {S.}~\bibnamefont {Yao}}, \bibinfo {author} {\bibfnamefont {N.}~\bibnamefont {Shinn}}, \bibinfo {author} {\bibfnamefont {P.}~\bibnamefont {Razavi}}, \ and\ \bibinfo {author} {\bibfnamefont {K.}~\bibnamefont {Narasimhan}},\ }\href@noop {} {\bibfield  {journal} {\bibinfo  {journal} {arXiv preprint arXiv:2406.12045}\ } (\bibinfo {year} {2024})}\BibitemShut {NoStop}%
\bibitem [{\citenamefont {Wang}\ \emph {et~al.}(2024)\citenamefont {Wang}, \citenamefont {Wang}, \citenamefont {Xue}, \citenamefont {Xia}, \citenamefont {Cao}, \citenamefont {Liu}, \citenamefont {Pan},\ and\ \citenamefont {Wong}}]{wang2024appbench}%
  \BibitemOpen
  \bibfield  {author} {\bibinfo {author} {\bibfnamefont {H.}~\bibnamefont {Wang}}, \bibinfo {author} {\bibfnamefont {R.}~\bibnamefont {Wang}}, \bibinfo {author} {\bibfnamefont {B.}~\bibnamefont {Xue}}, \bibinfo {author} {\bibfnamefont {H.}~\bibnamefont {Xia}}, \bibinfo {author} {\bibfnamefont {J.}~\bibnamefont {Cao}}, \bibinfo {author} {\bibfnamefont {Z.}~\bibnamefont {Liu}}, \bibinfo {author} {\bibfnamefont {J.~Z.}\ \bibnamefont {Pan}}, \ and\ \bibinfo {author} {\bibfnamefont {K.-F.}\ \bibnamefont {Wong}},\ }\href@noop {} {\bibfield  {journal} {\bibinfo  {journal} {arXiv preprint arXiv:2410.19743}\ } (\bibinfo {year} {2024})}\BibitemShut {NoStop}%
\bibitem [{\citenamefont {Blais}\ \emph {et~al.}(2004)\citenamefont {Blais}, \citenamefont {Huang}, \citenamefont {Wallraff}, \citenamefont {Girvin},\ and\ \citenamefont {Schoelkopf}}]{PhysRevA.69.062320}%
  \BibitemOpen
  \bibfield  {author} {\bibinfo {author} {\bibfnamefont {A.}~\bibnamefont {Blais}}, \bibinfo {author} {\bibfnamefont {R.-S.}\ \bibnamefont {Huang}}, \bibinfo {author} {\bibfnamefont {A.}~\bibnamefont {Wallraff}}, \bibinfo {author} {\bibfnamefont {S.~M.}\ \bibnamefont {Girvin}}, \ and\ \bibinfo {author} {\bibfnamefont {R.~J.}\ \bibnamefont {Schoelkopf}},\ }\href {\doibase 10.1103/PhysRevA.69.062320} {\bibfield  {journal} {\bibinfo  {journal} {Phys. Rev. A}\ }\textbf {\bibinfo {volume} {69}},\ \bibinfo {pages} {062320} (\bibinfo {year} {2004})}\BibitemShut {NoStop}%
\bibitem [{\citenamefont {Acharya}\ \emph {et~al.}(2023)\citenamefont {Acharya}, \citenamefont {Aleiner}, \citenamefont {Allen}, \citenamefont {Andersen}, \citenamefont {Ansmann}, \citenamefont {Arute}, \citenamefont {Arya}, \citenamefont {Asfaw}, \citenamefont {Atalaya}, \citenamefont {Babbush}, \citenamefont {Bacon}, \citenamefont {Bardin}, \citenamefont {Basso}, \citenamefont {Bengtsson}, \citenamefont {Boixo}, \citenamefont {Bortoli}, \citenamefont {Bourassa}, \citenamefont {Bovaird}, \citenamefont {Brill}, \citenamefont {Broughton}, \citenamefont {Buckley}, \citenamefont {Buell}, \citenamefont {Burger}, \citenamefont {Burkett}, \citenamefont {Bushnell}, \citenamefont {Chen}, \citenamefont {Chen}, \citenamefont {Chiaro}, \citenamefont {Cogan}, \citenamefont {Collins}, \citenamefont {Conner}, \citenamefont {Courtney}, \citenamefont {Crook}, \citenamefont {Curtin}, \citenamefont {Debroy}, \citenamefont {Del Toro~Barba}, \citenamefont {Demura}, \citenamefont {Dunsworth}, \citenamefont {Eppens}, \citenamefont
  {Erickson}, \citenamefont {Faoro}, \citenamefont {Farhi}, \citenamefont {Fatemi}, \citenamefont {Flores~Burgos}, \citenamefont {Forati}, \citenamefont {Fowler}, \citenamefont {Foxen}, \citenamefont {Giang}, \citenamefont {Gidney}, \citenamefont {Gilboa}, \citenamefont {Giustina}, \citenamefont {Grajales~Dau}, \citenamefont {Gross}, \citenamefont {Habegger}, \citenamefont {Hamilton}, \citenamefont {Harrigan}, \citenamefont {Harrington}, \citenamefont {Higgott}, \citenamefont {Hilton}, \citenamefont {Hoffmann}, \citenamefont {Hong}, \citenamefont {Huang}, \citenamefont {Huff}, \citenamefont {Huggins}, \citenamefont {Ioffe}, \citenamefont {Isakov}, \citenamefont {Iveland}, \citenamefont {Jeffrey}, \citenamefont {Jiang}, \citenamefont {Jones}, \citenamefont {Juhas}, \citenamefont {Kafri}, \citenamefont {Kechedzhi}, \citenamefont {Kelly}, \citenamefont {Khattar}, \citenamefont {Khezri}, \citenamefont {Kieferov{\'a}}, \citenamefont {Kim}, \citenamefont {Kitaev}, \citenamefont {Klimov}, \citenamefont {Klots},
  \citenamefont {Korotkov}, \citenamefont {Kostritsa}, \citenamefont {Kreikebaum}, \citenamefont {Landhuis}, \citenamefont {Laptev}, \citenamefont {Lau}, \citenamefont {Laws}, \citenamefont {Lee}, \citenamefont {Lee}, \citenamefont {Lester}, \citenamefont {Lill}, \citenamefont {Liu}, \citenamefont {Locharla}, \citenamefont {Lucero}, \citenamefont {Malone}, \citenamefont {Marshall}, \citenamefont {Martin}, \citenamefont {McClean}, \citenamefont {McCourt}, \citenamefont {McEwen}, \citenamefont {Megrant}, \citenamefont {Meurer~Costa}, \citenamefont {Mi}, \citenamefont {Miao}, \citenamefont {Mohseni}, \citenamefont {Montazeri}, \citenamefont {Morvan}, \citenamefont {Mount}, \citenamefont {Mruczkiewicz}, \citenamefont {Naaman}, \citenamefont {Neeley}, \citenamefont {Neill}, \citenamefont {Nersisyan}, \citenamefont {Neven}, \citenamefont {Newman}, \citenamefont {Ng}, \citenamefont {Nguyen}, \citenamefont {Nguyen}, \citenamefont {Niu}, \citenamefont {O'Brien}, \citenamefont {Opremcak}, \citenamefont {Platt},
  \citenamefont {Petukhov}, \citenamefont {Potter}, \citenamefont {Pryadko}, \citenamefont {Quintana}, \citenamefont {Roushan}, \citenamefont {Rubin}, \citenamefont {Saei}, \citenamefont {Sank}, \citenamefont {Sankaragomathi}, \citenamefont {Satzinger}, \citenamefont {Schurkus}, \citenamefont {Schuster}, \citenamefont {Shearn}, \citenamefont {Shorter}, \citenamefont {Shvarts}, \citenamefont {Skruzny}, \citenamefont {Smelyanskiy}, \citenamefont {Smith}, \citenamefont {Sterling}, \citenamefont {Strain}, \citenamefont {Szalay}, \citenamefont {Torres}, \citenamefont {Vidal}, \citenamefont {Villalonga}, \citenamefont {Vollgraff~Heidweiller}, \citenamefont {White}, \citenamefont {Xing}, \citenamefont {Yao}, \citenamefont {Yeh}, \citenamefont {Yoo}, \citenamefont {Young}, \citenamefont {Zalcman}, \citenamefont {Zhang}, \citenamefont {Zhu},\ and\ \citenamefont {AI}}]{Acharya2023}%
  \BibitemOpen
  \bibfield  {author} {\bibinfo {author} {\bibfnamefont {R.}~\bibnamefont {Acharya}}, \bibinfo {author} {\bibfnamefont {I.}~\bibnamefont {Aleiner}}, \bibinfo {author} {\bibfnamefont {R.}~\bibnamefont {Allen}}, \bibinfo {author} {\bibfnamefont {T.~I.}\ \bibnamefont {Andersen}}, \bibinfo {author} {\bibfnamefont {M.}~\bibnamefont {Ansmann}}, \bibinfo {author} {\bibfnamefont {F.}~\bibnamefont {Arute}}, \bibinfo {author} {\bibfnamefont {K.}~\bibnamefont {Arya}}, \bibinfo {author} {\bibfnamefont {A.}~\bibnamefont {Asfaw}}, \bibinfo {author} {\bibfnamefont {J.}~\bibnamefont {Atalaya}}, \bibinfo {author} {\bibfnamefont {R.}~\bibnamefont {Babbush}}, \bibinfo {author} {\bibfnamefont {D.}~\bibnamefont {Bacon}}, \bibinfo {author} {\bibfnamefont {J.~C.}\ \bibnamefont {Bardin}}, \bibinfo {author} {\bibfnamefont {J.}~\bibnamefont {Basso}}, \bibinfo {author} {\bibfnamefont {A.}~\bibnamefont {Bengtsson}}, \bibinfo {author} {\bibfnamefont {S.}~\bibnamefont {Boixo}}, \bibinfo {author} {\bibfnamefont {G.}~\bibnamefont
  {Bortoli}}, \bibinfo {author} {\bibfnamefont {A.}~\bibnamefont {Bourassa}}, \bibinfo {author} {\bibfnamefont {J.}~\bibnamefont {Bovaird}}, \bibinfo {author} {\bibfnamefont {L.}~\bibnamefont {Brill}}, \bibinfo {author} {\bibfnamefont {M.}~\bibnamefont {Broughton}}, \bibinfo {author} {\bibfnamefont {B.~B.}\ \bibnamefont {Buckley}}, \bibinfo {author} {\bibfnamefont {D.~A.}\ \bibnamefont {Buell}}, \bibinfo {author} {\bibfnamefont {T.}~\bibnamefont {Burger}}, \bibinfo {author} {\bibfnamefont {B.}~\bibnamefont {Burkett}}, \bibinfo {author} {\bibfnamefont {N.}~\bibnamefont {Bushnell}}, \bibinfo {author} {\bibfnamefont {Y.}~\bibnamefont {Chen}}, \bibinfo {author} {\bibfnamefont {Z.}~\bibnamefont {Chen}}, \bibinfo {author} {\bibfnamefont {B.}~\bibnamefont {Chiaro}}, \bibinfo {author} {\bibfnamefont {J.}~\bibnamefont {Cogan}}, \bibinfo {author} {\bibfnamefont {R.}~\bibnamefont {Collins}}, \bibinfo {author} {\bibfnamefont {P.}~\bibnamefont {Conner}}, \bibinfo {author} {\bibfnamefont {W.}~\bibnamefont {Courtney}},
  \bibinfo {author} {\bibfnamefont {A.~L.}\ \bibnamefont {Crook}}, \bibinfo {author} {\bibfnamefont {B.}~\bibnamefont {Curtin}}, \bibinfo {author} {\bibfnamefont {D.~M.}\ \bibnamefont {Debroy}}, \bibinfo {author} {\bibfnamefont {A.}~\bibnamefont {Del Toro~Barba}}, \bibinfo {author} {\bibfnamefont {S.}~\bibnamefont {Demura}}, \bibinfo {author} {\bibfnamefont {A.}~\bibnamefont {Dunsworth}}, \bibinfo {author} {\bibfnamefont {D.}~\bibnamefont {Eppens}}, \bibinfo {author} {\bibfnamefont {C.}~\bibnamefont {Erickson}}, \bibinfo {author} {\bibfnamefont {L.}~\bibnamefont {Faoro}}, \bibinfo {author} {\bibfnamefont {E.}~\bibnamefont {Farhi}}, \bibinfo {author} {\bibfnamefont {R.}~\bibnamefont {Fatemi}}, \bibinfo {author} {\bibfnamefont {L.}~\bibnamefont {Flores~Burgos}}, \bibinfo {author} {\bibfnamefont {E.}~\bibnamefont {Forati}}, \bibinfo {author} {\bibfnamefont {A.~G.}\ \bibnamefont {Fowler}}, \bibinfo {author} {\bibfnamefont {B.}~\bibnamefont {Foxen}}, \bibinfo {author} {\bibfnamefont {W.}~\bibnamefont {Giang}},
  \bibinfo {author} {\bibfnamefont {C.}~\bibnamefont {Gidney}}, \bibinfo {author} {\bibfnamefont {D.}~\bibnamefont {Gilboa}}, \bibinfo {author} {\bibfnamefont {M.}~\bibnamefont {Giustina}}, \bibinfo {author} {\bibfnamefont {A.}~\bibnamefont {Grajales~Dau}}, \bibinfo {author} {\bibfnamefont {J.~A.}\ \bibnamefont {Gross}}, \bibinfo {author} {\bibfnamefont {S.}~\bibnamefont {Habegger}}, \bibinfo {author} {\bibfnamefont {M.~C.}\ \bibnamefont {Hamilton}}, \bibinfo {author} {\bibfnamefont {M.~P.}\ \bibnamefont {Harrigan}}, \bibinfo {author} {\bibfnamefont {S.~D.}\ \bibnamefont {Harrington}}, \bibinfo {author} {\bibfnamefont {O.}~\bibnamefont {Higgott}}, \bibinfo {author} {\bibfnamefont {J.}~\bibnamefont {Hilton}}, \bibinfo {author} {\bibfnamefont {M.}~\bibnamefont {Hoffmann}}, \bibinfo {author} {\bibfnamefont {S.}~\bibnamefont {Hong}}, \bibinfo {author} {\bibfnamefont {T.}~\bibnamefont {Huang}}, \bibinfo {author} {\bibfnamefont {A.}~\bibnamefont {Huff}}, \bibinfo {author} {\bibfnamefont {W.~J.}\ \bibnamefont
  {Huggins}}, \bibinfo {author} {\bibfnamefont {L.~B.}\ \bibnamefont {Ioffe}}, \bibinfo {author} {\bibfnamefont {S.~V.}\ \bibnamefont {Isakov}}, \bibinfo {author} {\bibfnamefont {J.}~\bibnamefont {Iveland}}, \bibinfo {author} {\bibfnamefont {E.}~\bibnamefont {Jeffrey}}, \bibinfo {author} {\bibfnamefont {Z.}~\bibnamefont {Jiang}}, \bibinfo {author} {\bibfnamefont {C.}~\bibnamefont {Jones}}, \bibinfo {author} {\bibfnamefont {P.}~\bibnamefont {Juhas}}, \bibinfo {author} {\bibfnamefont {D.}~\bibnamefont {Kafri}}, \bibinfo {author} {\bibfnamefont {K.}~\bibnamefont {Kechedzhi}}, \bibinfo {author} {\bibfnamefont {J.}~\bibnamefont {Kelly}}, \bibinfo {author} {\bibfnamefont {T.}~\bibnamefont {Khattar}}, \bibinfo {author} {\bibfnamefont {M.}~\bibnamefont {Khezri}}, \bibinfo {author} {\bibfnamefont {M.}~\bibnamefont {Kieferov{\'a}}}, \bibinfo {author} {\bibfnamefont {S.}~\bibnamefont {Kim}}, \bibinfo {author} {\bibfnamefont {A.}~\bibnamefont {Kitaev}}, \bibinfo {author} {\bibfnamefont {P.~V.}\ \bibnamefont {Klimov}},
  \bibinfo {author} {\bibfnamefont {A.~R.}\ \bibnamefont {Klots}}, \bibinfo {author} {\bibfnamefont {A.~N.}\ \bibnamefont {Korotkov}}, \bibinfo {author} {\bibfnamefont {F.}~\bibnamefont {Kostritsa}}, \bibinfo {author} {\bibfnamefont {J.~M.}\ \bibnamefont {Kreikebaum}}, \bibinfo {author} {\bibfnamefont {D.}~\bibnamefont {Landhuis}}, \bibinfo {author} {\bibfnamefont {P.}~\bibnamefont {Laptev}}, \bibinfo {author} {\bibfnamefont {K.-M.}\ \bibnamefont {Lau}}, \bibinfo {author} {\bibfnamefont {L.}~\bibnamefont {Laws}}, \bibinfo {author} {\bibfnamefont {J.}~\bibnamefont {Lee}}, \bibinfo {author} {\bibfnamefont {K.}~\bibnamefont {Lee}}, \bibinfo {author} {\bibfnamefont {B.~J.}\ \bibnamefont {Lester}}, \bibinfo {author} {\bibfnamefont {A.}~\bibnamefont {Lill}}, \bibinfo {author} {\bibfnamefont {W.}~\bibnamefont {Liu}}, \bibinfo {author} {\bibfnamefont {A.}~\bibnamefont {Locharla}}, \bibinfo {author} {\bibfnamefont {E.}~\bibnamefont {Lucero}}, \bibinfo {author} {\bibfnamefont {F.~D.}\ \bibnamefont {Malone}}, \bibinfo
  {author} {\bibfnamefont {J.}~\bibnamefont {Marshall}}, \bibinfo {author} {\bibfnamefont {O.}~\bibnamefont {Martin}}, \bibinfo {author} {\bibfnamefont {J.~R.}\ \bibnamefont {McClean}}, \bibinfo {author} {\bibfnamefont {T.}~\bibnamefont {McCourt}}, \bibinfo {author} {\bibfnamefont {M.}~\bibnamefont {McEwen}}, \bibinfo {author} {\bibfnamefont {A.}~\bibnamefont {Megrant}}, \bibinfo {author} {\bibfnamefont {B.}~\bibnamefont {Meurer~Costa}}, \bibinfo {author} {\bibfnamefont {X.}~\bibnamefont {Mi}}, \bibinfo {author} {\bibfnamefont {K.~C.}\ \bibnamefont {Miao}}, \bibinfo {author} {\bibfnamefont {M.}~\bibnamefont {Mohseni}}, \bibinfo {author} {\bibfnamefont {S.}~\bibnamefont {Montazeri}}, \bibinfo {author} {\bibfnamefont {A.}~\bibnamefont {Morvan}}, \bibinfo {author} {\bibfnamefont {E.}~\bibnamefont {Mount}}, \bibinfo {author} {\bibfnamefont {W.}~\bibnamefont {Mruczkiewicz}}, \bibinfo {author} {\bibfnamefont {O.}~\bibnamefont {Naaman}}, \bibinfo {author} {\bibfnamefont {M.}~\bibnamefont {Neeley}}, \bibinfo {author}
  {\bibfnamefont {C.}~\bibnamefont {Neill}}, \bibinfo {author} {\bibfnamefont {A.}~\bibnamefont {Nersisyan}}, \bibinfo {author} {\bibfnamefont {H.}~\bibnamefont {Neven}}, \bibinfo {author} {\bibfnamefont {M.}~\bibnamefont {Newman}}, \bibinfo {author} {\bibfnamefont {J.~H.}\ \bibnamefont {Ng}}, \bibinfo {author} {\bibfnamefont {A.}~\bibnamefont {Nguyen}}, \bibinfo {author} {\bibfnamefont {M.}~\bibnamefont {Nguyen}}, \bibinfo {author} {\bibfnamefont {M.~Y.}\ \bibnamefont {Niu}}, \bibinfo {author} {\bibfnamefont {T.~E.}\ \bibnamefont {O'Brien}}, \bibinfo {author} {\bibfnamefont {A.}~\bibnamefont {Opremcak}}, \bibinfo {author} {\bibfnamefont {J.}~\bibnamefont {Platt}}, \bibinfo {author} {\bibfnamefont {A.}~\bibnamefont {Petukhov}}, \bibinfo {author} {\bibfnamefont {R.}~\bibnamefont {Potter}}, \bibinfo {author} {\bibfnamefont {L.~P.}\ \bibnamefont {Pryadko}}, \bibinfo {author} {\bibfnamefont {C.}~\bibnamefont {Quintana}}, \bibinfo {author} {\bibfnamefont {P.}~\bibnamefont {Roushan}}, \bibinfo {author}
  {\bibfnamefont {N.~C.}\ \bibnamefont {Rubin}}, \bibinfo {author} {\bibfnamefont {N.}~\bibnamefont {Saei}}, \bibinfo {author} {\bibfnamefont {D.}~\bibnamefont {Sank}}, \bibinfo {author} {\bibfnamefont {K.}~\bibnamefont {Sankaragomathi}}, \bibinfo {author} {\bibfnamefont {K.~J.}\ \bibnamefont {Satzinger}}, \bibinfo {author} {\bibfnamefont {H.~F.}\ \bibnamefont {Schurkus}}, \bibinfo {author} {\bibfnamefont {C.}~\bibnamefont {Schuster}}, \bibinfo {author} {\bibfnamefont {M.~J.}\ \bibnamefont {Shearn}}, \bibinfo {author} {\bibfnamefont {A.}~\bibnamefont {Shorter}}, \bibinfo {author} {\bibfnamefont {V.}~\bibnamefont {Shvarts}}, \bibinfo {author} {\bibfnamefont {J.}~\bibnamefont {Skruzny}}, \bibinfo {author} {\bibfnamefont {V.}~\bibnamefont {Smelyanskiy}}, \bibinfo {author} {\bibfnamefont {W.~C.}\ \bibnamefont {Smith}}, \bibinfo {author} {\bibfnamefont {G.}~\bibnamefont {Sterling}}, \bibinfo {author} {\bibfnamefont {D.}~\bibnamefont {Strain}}, \bibinfo {author} {\bibfnamefont {M.}~\bibnamefont {Szalay}}, \bibinfo
  {author} {\bibfnamefont {A.}~\bibnamefont {Torres}}, \bibinfo {author} {\bibfnamefont {G.}~\bibnamefont {Vidal}}, \bibinfo {author} {\bibfnamefont {B.}~\bibnamefont {Villalonga}}, \bibinfo {author} {\bibfnamefont {C.}~\bibnamefont {Vollgraff~Heidweiller}}, \bibinfo {author} {\bibfnamefont {T.}~\bibnamefont {White}}, \bibinfo {author} {\bibfnamefont {C.}~\bibnamefont {Xing}}, \bibinfo {author} {\bibfnamefont {Z.~J.}\ \bibnamefont {Yao}}, \bibinfo {author} {\bibfnamefont {P.}~\bibnamefont {Yeh}}, \bibinfo {author} {\bibfnamefont {J.}~\bibnamefont {Yoo}}, \bibinfo {author} {\bibfnamefont {G.}~\bibnamefont {Young}}, \bibinfo {author} {\bibfnamefont {A.}~\bibnamefont {Zalcman}}, \bibinfo {author} {\bibfnamefont {Y.}~\bibnamefont {Zhang}}, \bibinfo {author} {\bibfnamefont {N.}~\bibnamefont {Zhu}}, \ and\ \bibinfo {author} {\bibfnamefont {G.~Q.}\ \bibnamefont {AI}},\ }\href {\doibase 10.1038/s41586-022-05434-1} {\bibfield  {journal} {\bibinfo  {journal} {Nature}\ }\textbf {\bibinfo {volume} {614}},\ \bibinfo
  {pages} {676} (\bibinfo {year} {2023})}\BibitemShut {NoStop}%
\bibitem [{\citenamefont {Krantz}\ \emph {et~al.}(2019)\citenamefont {Krantz}, \citenamefont {Kjaergaard}, \citenamefont {Yan}, \citenamefont {Orlando}, \citenamefont {Gustavsson},\ and\ \citenamefont {Oliver}}]{10.1063/1.5089550}%
  \BibitemOpen
  \bibfield  {author} {\bibinfo {author} {\bibfnamefont {P.}~\bibnamefont {Krantz}}, \bibinfo {author} {\bibfnamefont {M.}~\bibnamefont {Kjaergaard}}, \bibinfo {author} {\bibfnamefont {F.}~\bibnamefont {Yan}}, \bibinfo {author} {\bibfnamefont {T.~P.}\ \bibnamefont {Orlando}}, \bibinfo {author} {\bibfnamefont {S.}~\bibnamefont {Gustavsson}}, \ and\ \bibinfo {author} {\bibfnamefont {W.~D.}\ \bibnamefont {Oliver}},\ }\href {\doibase 10.1063/1.5089550} {\bibfield  {journal} {\bibinfo  {journal} {Applied Physics Reviews}\ }\textbf {\bibinfo {volume} {6}},\ \bibinfo {pages} {021318} (\bibinfo {year} {2019})},\ \Eprint {http://arxiv.org/abs/https://pubs.aip.org/aip/apr/article-pdf/doi/10.1063/1.5089550/16667201/021318\_1\_online.pdf} {https://pubs.aip.org/aip/apr/article-pdf/doi/10.1063/1.5089550/16667201/021318\_1\_online.pdf} \BibitemShut {NoStop}%
\bibitem [{\citenamefont {Kjaergaard}\ \emph {et~al.}(2020)\citenamefont {Kjaergaard}, \citenamefont {Schwartz}, \citenamefont {Braum{\"u}ller}, \citenamefont {Krantz}, \citenamefont {Wang}, \citenamefont {Gustavsson},\ and\ \citenamefont {Oliver}}]{annurev-conmatphys-031119-050605}%
  \BibitemOpen
  \bibfield  {author} {\bibinfo {author} {\bibfnamefont {M.}~\bibnamefont {Kjaergaard}}, \bibinfo {author} {\bibfnamefont {M.~E.}\ \bibnamefont {Schwartz}}, \bibinfo {author} {\bibfnamefont {J.}~\bibnamefont {Braum{\"u}ller}}, \bibinfo {author} {\bibfnamefont {P.}~\bibnamefont {Krantz}}, \bibinfo {author} {\bibfnamefont {J.~I.-J.}\ \bibnamefont {Wang}}, \bibinfo {author} {\bibfnamefont {S.}~\bibnamefont {Gustavsson}}, \ and\ \bibinfo {author} {\bibfnamefont {W.~D.}\ \bibnamefont {Oliver}},\ }\href {\doibase https://doi.org/10.1146/annurev-conmatphys-031119-050605} {\bibfield  {journal} {\bibinfo  {journal} {Annual Review of Condensed Matter Physics}\ }\textbf {\bibinfo {volume} {11}},\ \bibinfo {pages} {369} (\bibinfo {year} {2020})}\BibitemShut {NoStop}%
\bibitem [{\citenamefont {Sheldon}\ \emph {et~al.}(2016)\citenamefont {Sheldon}, \citenamefont {Magesan}, \citenamefont {Chow},\ and\ \citenamefont {Gambetta}}]{PhysRevA.93.060302}%
  \BibitemOpen
  \bibfield  {author} {\bibinfo {author} {\bibfnamefont {S.}~\bibnamefont {Sheldon}}, \bibinfo {author} {\bibfnamefont {E.}~\bibnamefont {Magesan}}, \bibinfo {author} {\bibfnamefont {J.~M.}\ \bibnamefont {Chow}}, \ and\ \bibinfo {author} {\bibfnamefont {J.~M.}\ \bibnamefont {Gambetta}},\ }\href {\doibase 10.1103/PhysRevA.93.060302} {\bibfield  {journal} {\bibinfo  {journal} {Phys. Rev. A}\ }\textbf {\bibinfo {volume} {93}},\ \bibinfo {pages} {060302} (\bibinfo {year} {2016})}\BibitemShut {NoStop}%
\bibitem [{\citenamefont {Wei}\ \emph {et~al.}(2024)\citenamefont {Wei}, \citenamefont {Lauer}, \citenamefont {Pritchett}, \citenamefont {Shanks}, \citenamefont {McKay},\ and\ \citenamefont {Javadi-Abhari}}]{PRXQuantum.5.020338}%
  \BibitemOpen
  \bibfield  {author} {\bibinfo {author} {\bibfnamefont {K.~X.}\ \bibnamefont {Wei}}, \bibinfo {author} {\bibfnamefont {I.}~\bibnamefont {Lauer}}, \bibinfo {author} {\bibfnamefont {E.}~\bibnamefont {Pritchett}}, \bibinfo {author} {\bibfnamefont {W.}~\bibnamefont {Shanks}}, \bibinfo {author} {\bibfnamefont {D.~C.}\ \bibnamefont {McKay}}, \ and\ \bibinfo {author} {\bibfnamefont {A.}~\bibnamefont {Javadi-Abhari}},\ }\href {\doibase 10.1103/PRXQuantum.5.020338} {\bibfield  {journal} {\bibinfo  {journal} {PRX Quantum}\ }\textbf {\bibinfo {volume} {5}},\ \bibinfo {pages} {020338} (\bibinfo {year} {2024})}\BibitemShut {NoStop}%
\bibitem [{\citenamefont {Maksymov}\ \emph {et~al.}(2021)\citenamefont {Maksymov}, \citenamefont {Niroula},\ and\ \citenamefont {Nam}}]{maksymov2021optimal}%
  \BibitemOpen
  \bibfield  {author} {\bibinfo {author} {\bibfnamefont {A.}~\bibnamefont {Maksymov}}, \bibinfo {author} {\bibfnamefont {P.}~\bibnamefont {Niroula}}, \ and\ \bibinfo {author} {\bibfnamefont {Y.}~\bibnamefont {Nam}},\ }\href@noop {} {\bibfield  {journal} {\bibinfo  {journal} {Quantum Science and Technology}\ }\textbf {\bibinfo {volume} {6}},\ \bibinfo {pages} {034009} (\bibinfo {year} {2021})}\BibitemShut {NoStop}%
\bibitem [{\citenamefont {Romaszko}\ \emph {et~al.}(2020)\citenamefont {Romaszko}, \citenamefont {Hong}, \citenamefont {Siegele}, \citenamefont {Puddy}, \citenamefont {Lebrun-Gallagher}, \citenamefont {Weidt},\ and\ \citenamefont {Hensinger}}]{romaszko2020engineering}%
  \BibitemOpen
  \bibfield  {author} {\bibinfo {author} {\bibfnamefont {Z.~D.}\ \bibnamefont {Romaszko}}, \bibinfo {author} {\bibfnamefont {S.}~\bibnamefont {Hong}}, \bibinfo {author} {\bibfnamefont {M.}~\bibnamefont {Siegele}}, \bibinfo {author} {\bibfnamefont {R.~K.}\ \bibnamefont {Puddy}}, \bibinfo {author} {\bibfnamefont {F.~R.}\ \bibnamefont {Lebrun-Gallagher}}, \bibinfo {author} {\bibfnamefont {S.}~\bibnamefont {Weidt}}, \ and\ \bibinfo {author} {\bibfnamefont {W.~K.}\ \bibnamefont {Hensinger}},\ }\href@noop {} {\bibfield  {journal} {\bibinfo  {journal} {Nature Reviews Physics}\ }\textbf {\bibinfo {volume} {2}},\ \bibinfo {pages} {285} (\bibinfo {year} {2020})}\BibitemShut {NoStop}%
\bibitem [{\citenamefont {Bruzewicz}\ \emph {et~al.}(2019)\citenamefont {Bruzewicz}, \citenamefont {Chiaverini}, \citenamefont {McConnell},\ and\ \citenamefont {Sage}}]{bruzewicz2019trapped}%
  \BibitemOpen
  \bibfield  {author} {\bibinfo {author} {\bibfnamefont {C.~D.}\ \bibnamefont {Bruzewicz}}, \bibinfo {author} {\bibfnamefont {J.}~\bibnamefont {Chiaverini}}, \bibinfo {author} {\bibfnamefont {R.}~\bibnamefont {McConnell}}, \ and\ \bibinfo {author} {\bibfnamefont {J.~M.}\ \bibnamefont {Sage}},\ }\href@noop {} {\bibfield  {journal} {\bibinfo  {journal} {Applied Physics Reviews}\ }\textbf {\bibinfo {volume} {6}} (\bibinfo {year} {2019})}\BibitemShut {NoStop}%
\bibitem [{\citenamefont {Botzem}\ \emph {et~al.}(2018)\citenamefont {Botzem}, \citenamefont {Shulman}, \citenamefont {Foletti}, \citenamefont {Harvey}, \citenamefont {Dial}, \citenamefont {Bethke}, \citenamefont {Cerfontaine}, \citenamefont {McNeil}, \citenamefont {Mahalu}, \citenamefont {Umansky} \emph {et~al.}}]{botzem2018tuning}%
  \BibitemOpen
  \bibfield  {author} {\bibinfo {author} {\bibfnamefont {T.}~\bibnamefont {Botzem}}, \bibinfo {author} {\bibfnamefont {M.~D.}\ \bibnamefont {Shulman}}, \bibinfo {author} {\bibfnamefont {S.}~\bibnamefont {Foletti}}, \bibinfo {author} {\bibfnamefont {S.~P.}\ \bibnamefont {Harvey}}, \bibinfo {author} {\bibfnamefont {O.~E.}\ \bibnamefont {Dial}}, \bibinfo {author} {\bibfnamefont {P.}~\bibnamefont {Bethke}}, \bibinfo {author} {\bibfnamefont {P.}~\bibnamefont {Cerfontaine}}, \bibinfo {author} {\bibfnamefont {R.~P.}\ \bibnamefont {McNeil}}, \bibinfo {author} {\bibfnamefont {D.}~\bibnamefont {Mahalu}}, \bibinfo {author} {\bibfnamefont {V.}~\bibnamefont {Umansky}},  \emph {et~al.},\ }\href@noop {} {\bibfield  {journal} {\bibinfo  {journal} {Physical Review Applied}\ }\textbf {\bibinfo {volume} {10}},\ \bibinfo {pages} {054026} (\bibinfo {year} {2018})}\BibitemShut {NoStop}%
\bibitem [{\citenamefont {Burkard}\ \emph {et~al.}(2023)\citenamefont {Burkard}, \citenamefont {Ladd}, \citenamefont {Pan}, \citenamefont {Nichol},\ and\ \citenamefont {Petta}}]{burkard2023semiconductor}%
  \BibitemOpen
  \bibfield  {author} {\bibinfo {author} {\bibfnamefont {G.}~\bibnamefont {Burkard}}, \bibinfo {author} {\bibfnamefont {T.~D.}\ \bibnamefont {Ladd}}, \bibinfo {author} {\bibfnamefont {A.}~\bibnamefont {Pan}}, \bibinfo {author} {\bibfnamefont {J.~M.}\ \bibnamefont {Nichol}}, \ and\ \bibinfo {author} {\bibfnamefont {J.~R.}\ \bibnamefont {Petta}},\ }\href@noop {} {\bibfield  {journal} {\bibinfo  {journal} {Reviews of Modern Physics}\ }\textbf {\bibinfo {volume} {95}},\ \bibinfo {pages} {025003} (\bibinfo {year} {2023})}\BibitemShut {NoStop}%
\bibitem [{\citenamefont {Chatterjee}\ \emph {et~al.}(2021)\citenamefont {Chatterjee}, \citenamefont {Stevenson}, \citenamefont {De~Franceschi}, \citenamefont {Morello}, \citenamefont {de~Leon},\ and\ \citenamefont {Kuemmeth}}]{chatterjee2021semiconductor}%
  \BibitemOpen
  \bibfield  {author} {\bibinfo {author} {\bibfnamefont {A.}~\bibnamefont {Chatterjee}}, \bibinfo {author} {\bibfnamefont {P.}~\bibnamefont {Stevenson}}, \bibinfo {author} {\bibfnamefont {S.}~\bibnamefont {De~Franceschi}}, \bibinfo {author} {\bibfnamefont {A.}~\bibnamefont {Morello}}, \bibinfo {author} {\bibfnamefont {N.~P.}\ \bibnamefont {de~Leon}}, \ and\ \bibinfo {author} {\bibfnamefont {F.}~\bibnamefont {Kuemmeth}},\ }\href@noop {} {\bibfield  {journal} {\bibinfo  {journal} {Nature Reviews Physics}\ }\textbf {\bibinfo {volume} {3}},\ \bibinfo {pages} {157} (\bibinfo {year} {2021})}\BibitemShut {NoStop}%
\bibitem [{\citenamefont {Stano}\ and\ \citenamefont {Loss}(2022)}]{stano2022review}%
  \BibitemOpen
  \bibfield  {author} {\bibinfo {author} {\bibfnamefont {P.}~\bibnamefont {Stano}}\ and\ \bibinfo {author} {\bibfnamefont {D.}~\bibnamefont {Loss}},\ }\href@noop {} {\bibfield  {journal} {\bibinfo  {journal} {Nature Reviews Physics}\ }\textbf {\bibinfo {volume} {4}},\ \bibinfo {pages} {672} (\bibinfo {year} {2022})}\BibitemShut {NoStop}%
\bibitem [{\citenamefont {Wu}\ \emph {et~al.}(2021)\citenamefont {Wu}, \citenamefont {Liang}, \citenamefont {Tian}, \citenamefont {Yang}, \citenamefont {Chen}, \citenamefont {Liu}, \citenamefont {Tey},\ and\ \citenamefont {You}}]{wu2021concise}%
  \BibitemOpen
  \bibfield  {author} {\bibinfo {author} {\bibfnamefont {X.}~\bibnamefont {Wu}}, \bibinfo {author} {\bibfnamefont {X.}~\bibnamefont {Liang}}, \bibinfo {author} {\bibfnamefont {Y.}~\bibnamefont {Tian}}, \bibinfo {author} {\bibfnamefont {F.}~\bibnamefont {Yang}}, \bibinfo {author} {\bibfnamefont {C.}~\bibnamefont {Chen}}, \bibinfo {author} {\bibfnamefont {Y.-C.}\ \bibnamefont {Liu}}, \bibinfo {author} {\bibfnamefont {M.~K.}\ \bibnamefont {Tey}}, \ and\ \bibinfo {author} {\bibfnamefont {L.}~\bibnamefont {You}},\ }\href@noop {} {\bibfield  {journal} {\bibinfo  {journal} {Chinese Physics B}\ }\textbf {\bibinfo {volume} {30}},\ \bibinfo {pages} {020305} (\bibinfo {year} {2021})}\BibitemShut {NoStop}%
\bibitem [{\citenamefont {Adams}\ \emph {et~al.}(2019)\citenamefont {Adams}, \citenamefont {Pritchard},\ and\ \citenamefont {Shaffer}}]{adams2019rydberg}%
  \BibitemOpen
  \bibfield  {author} {\bibinfo {author} {\bibfnamefont {C.~S.}\ \bibnamefont {Adams}}, \bibinfo {author} {\bibfnamefont {J.~D.}\ \bibnamefont {Pritchard}}, \ and\ \bibinfo {author} {\bibfnamefont {J.~P.}\ \bibnamefont {Shaffer}},\ }\href@noop {} {\bibfield  {journal} {\bibinfo  {journal} {Journal of Physics B: Atomic, Molecular and Optical Physics}\ }\textbf {\bibinfo {volume} {53}},\ \bibinfo {pages} {012002} (\bibinfo {year} {2019})}\BibitemShut {NoStop}%
\bibitem [{\citenamefont {Cong}\ \emph {et~al.}(2022)\citenamefont {Cong}, \citenamefont {Levine}, \citenamefont {Keesling}, \citenamefont {Bluvstein}, \citenamefont {Wang},\ and\ \citenamefont {Lukin}}]{cong2022hardware}%
  \BibitemOpen
  \bibfield  {author} {\bibinfo {author} {\bibfnamefont {I.}~\bibnamefont {Cong}}, \bibinfo {author} {\bibfnamefont {H.}~\bibnamefont {Levine}}, \bibinfo {author} {\bibfnamefont {A.}~\bibnamefont {Keesling}}, \bibinfo {author} {\bibfnamefont {D.}~\bibnamefont {Bluvstein}}, \bibinfo {author} {\bibfnamefont {S.-T.}\ \bibnamefont {Wang}}, \ and\ \bibinfo {author} {\bibfnamefont {M.~D.}\ \bibnamefont {Lukin}},\ }\href@noop {} {\bibfield  {journal} {\bibinfo  {journal} {Physical Review X}\ }\textbf {\bibinfo {volume} {12}},\ \bibinfo {pages} {021049} (\bibinfo {year} {2022})}\BibitemShut {NoStop}%
\bibitem [{\citenamefont {Cao}\ \emph {et~al.}(2024{\natexlab{a}})\citenamefont {Cao}, \citenamefont {Zhang},\ and\ \citenamefont {Alghadeer}}]{LeeQRepo}%
  \BibitemOpen
  \bibfield  {author} {\bibinfo {author} {\bibfnamefont {S.}~\bibnamefont {Cao}}, \bibinfo {author} {\bibfnamefont {Z.}~\bibnamefont {Zhang}}, \ and\ \bibinfo {author} {\bibfnamefont {M.}~\bibnamefont {Alghadeer}},\ }\href@noop {} {\enquote {\bibinfo {title} {Leeq: a python package for orchestrating quantum computing experiments},}\ }\bibinfo {howpublished} {\url{https://github.com/ShuxiangCao/LeeQ}} (\bibinfo {year} {2024}{\natexlab{a}})\BibitemShut {NoStop}%
\bibitem [{\citenamefont {Alghadeer}\ \emph {et~al.}(2025{\natexlab{a}})\citenamefont {Alghadeer}, \citenamefont {Cao}, \citenamefont {Fasciati}, \citenamefont {Piscitelli}, \citenamefont {Gow}, \citenamefont {Gates}, \citenamefont {Bakr},\ and\ \citenamefont {Leek}}]{2505.22276}%
  \BibitemOpen
  \bibfield  {author} {\bibinfo {author} {\bibfnamefont {M.}~\bibnamefont {Alghadeer}}, \bibinfo {author} {\bibfnamefont {S.}~\bibnamefont {Cao}}, \bibinfo {author} {\bibfnamefont {S.~D.}\ \bibnamefont {Fasciati}}, \bibinfo {author} {\bibfnamefont {M.}~\bibnamefont {Piscitelli}}, \bibinfo {author} {\bibfnamefont {P.~C.}\ \bibnamefont {Gow}}, \bibinfo {author} {\bibfnamefont {J.~C.}\ \bibnamefont {Gates}}, \bibinfo {author} {\bibfnamefont {M.}~\bibnamefont {Bakr}}, \ and\ \bibinfo {author} {\bibfnamefont {P.~J.}\ \bibnamefont {Leek}},\ }\href@noop {} {\enquote {\bibinfo {title} {Low crosstalk in a scalable superconducting quantum lattice},}\ } (\bibinfo {year} {2025}{\natexlab{a}}),\ \Eprint {http://arxiv.org/abs/arXiv:2505.22276} {arXiv:2505.22276} \BibitemShut {NoStop}%
\bibitem [{\citenamefont {Spring}\ \emph {et~al.}(2022)\citenamefont {Spring}, \citenamefont {Cao}, \citenamefont {Tsunoda}, \citenamefont {Campanaro}, \citenamefont {Fasciati}, \citenamefont {Wills}, \citenamefont {Bakr}, \citenamefont {Chidambaram}, \citenamefont {Shteynas}, \citenamefont {Carpenter}, \citenamefont {Gow}, \citenamefont {Gates}, \citenamefont {Vlastakis},\ and\ \citenamefont {Leek}}]{Spring2022}%
  \BibitemOpen
  \bibfield  {author} {\bibinfo {author} {\bibfnamefont {P.~A.}\ \bibnamefont {Spring}}, \bibinfo {author} {\bibfnamefont {S.}~\bibnamefont {Cao}}, \bibinfo {author} {\bibfnamefont {T.}~\bibnamefont {Tsunoda}}, \bibinfo {author} {\bibfnamefont {G.}~\bibnamefont {Campanaro}}, \bibinfo {author} {\bibfnamefont {S.}~\bibnamefont {Fasciati}}, \bibinfo {author} {\bibfnamefont {J.}~\bibnamefont {Wills}}, \bibinfo {author} {\bibfnamefont {M.}~\bibnamefont {Bakr}}, \bibinfo {author} {\bibfnamefont {V.}~\bibnamefont {Chidambaram}}, \bibinfo {author} {\bibfnamefont {B.}~\bibnamefont {Shteynas}}, \bibinfo {author} {\bibfnamefont {L.}~\bibnamefont {Carpenter}}, \bibinfo {author} {\bibfnamefont {P.}~\bibnamefont {Gow}}, \bibinfo {author} {\bibfnamefont {J.}~\bibnamefont {Gates}}, \bibinfo {author} {\bibfnamefont {B.}~\bibnamefont {Vlastakis}}, \ and\ \bibinfo {author} {\bibfnamefont {P.~J.}\ \bibnamefont {Leek}},\ }\href {\doibase 10.1126/sciadv.abl6698} {\bibfield  {journal} {\bibinfo  {journal} {Science Advances}\
  }\textbf {\bibinfo {volume} {8}} (\bibinfo {year} {2022}),\ 10.1126/sciadv.abl6698}\BibitemShut {NoStop}%
\bibitem [{\citenamefont {Rahamim}\ \emph {et~al.}(2017)\citenamefont {Rahamim}, \citenamefont {Behrle}, \citenamefont {Peterer}, \citenamefont {Patterson}, \citenamefont {Spring}, \citenamefont {Tsunoda}, \citenamefont {Manenti}, \citenamefont {Tancredi},\ and\ \citenamefont {Leek}}]{Rahamim2017Double-sided}%
  \BibitemOpen
  \bibfield  {author} {\bibinfo {author} {\bibfnamefont {J.}~\bibnamefont {Rahamim}}, \bibinfo {author} {\bibfnamefont {T.}~\bibnamefont {Behrle}}, \bibinfo {author} {\bibfnamefont {M.~J.}\ \bibnamefont {Peterer}}, \bibinfo {author} {\bibfnamefont {A.}~\bibnamefont {Patterson}}, \bibinfo {author} {\bibfnamefont {P.~A.}\ \bibnamefont {Spring}}, \bibinfo {author} {\bibfnamefont {T.}~\bibnamefont {Tsunoda}}, \bibinfo {author} {\bibfnamefont {R.}~\bibnamefont {Manenti}}, \bibinfo {author} {\bibfnamefont {G.}~\bibnamefont {Tancredi}}, \ and\ \bibinfo {author} {\bibfnamefont {P.~J.}\ \bibnamefont {Leek}},\ }\href {\doibase 10.1063/1.4984299} {\bibfield  {journal} {\bibinfo  {journal} {Applied Physics Letters}\ }\textbf {\bibinfo {volume} {110}},\ \bibinfo {pages} {222602} (\bibinfo {year} {2017})},\ \Eprint {http://arxiv.org/abs/https://doi.org/10.1063/1.4984299} {https://doi.org/10.1063/1.4984299} \BibitemShut {NoStop}%
\bibitem [{\citenamefont {Alghadeer}\ \emph {et~al.}(2025{\natexlab{b}})\citenamefont {Alghadeer}, \citenamefont {Fasciati}, \citenamefont {Cao}, \citenamefont {Piscitelli}, \citenamefont {Spink}, \citenamefont {Hopkinson}, \citenamefont {Danaie}, \citenamefont {Speller}, \citenamefont {Leek},\ and\ \citenamefont {Bakr}}]{2501.15059}%
  \BibitemOpen
  \bibfield  {author} {\bibinfo {author} {\bibfnamefont {M.}~\bibnamefont {Alghadeer}}, \bibinfo {author} {\bibfnamefont {S.~D.}\ \bibnamefont {Fasciati}}, \bibinfo {author} {\bibfnamefont {S.}~\bibnamefont {Cao}}, \bibinfo {author} {\bibfnamefont {M.}~\bibnamefont {Piscitelli}}, \bibinfo {author} {\bibfnamefont {M.~C.}\ \bibnamefont {Spink}}, \bibinfo {author} {\bibfnamefont {D.~G.}\ \bibnamefont {Hopkinson}}, \bibinfo {author} {\bibfnamefont {M.}~\bibnamefont {Danaie}}, \bibinfo {author} {\bibfnamefont {S.~C.}\ \bibnamefont {Speller}}, \bibinfo {author} {\bibfnamefont {P.~J.}\ \bibnamefont {Leek}}, \ and\ \bibinfo {author} {\bibfnamefont {M.}~\bibnamefont {Bakr}},\ }\href@noop {} {\enquote {\bibinfo {title} {Characterization of nanostructural imperfections in superconducting quantum circuits},}\ } (\bibinfo {year} {2025}{\natexlab{b}}),\ \Eprint {http://arxiv.org/abs/arXiv:2501.15059} {arXiv:2501.15059} \BibitemShut {NoStop}%
\bibitem [{\citenamefont {Mitchell}\ \emph {et~al.}(2021)\citenamefont {Mitchell}, \citenamefont {Naik}, \citenamefont {Morvan}, \citenamefont {Hashim}, \citenamefont {Kreikebaum}, \citenamefont {Marinelli}, \citenamefont {Lavrijsen}, \citenamefont {Nowrouzi}, \citenamefont {Santiago},\ and\ \citenamefont {Siddiqi}}]{PhysRevLett.127.200502}%
  \BibitemOpen
  \bibfield  {author} {\bibinfo {author} {\bibfnamefont {B.~K.}\ \bibnamefont {Mitchell}}, \bibinfo {author} {\bibfnamefont {R.~K.}\ \bibnamefont {Naik}}, \bibinfo {author} {\bibfnamefont {A.}~\bibnamefont {Morvan}}, \bibinfo {author} {\bibfnamefont {A.}~\bibnamefont {Hashim}}, \bibinfo {author} {\bibfnamefont {J.~M.}\ \bibnamefont {Kreikebaum}}, \bibinfo {author} {\bibfnamefont {B.}~\bibnamefont {Marinelli}}, \bibinfo {author} {\bibfnamefont {W.}~\bibnamefont {Lavrijsen}}, \bibinfo {author} {\bibfnamefont {K.}~\bibnamefont {Nowrouzi}}, \bibinfo {author} {\bibfnamefont {D.~I.}\ \bibnamefont {Santiago}}, \ and\ \bibinfo {author} {\bibfnamefont {I.}~\bibnamefont {Siddiqi}},\ }\href {\doibase 10.1103/PhysRevLett.127.200502} {\bibfield  {journal} {\bibinfo  {journal} {Phys. Rev. Lett.}\ }\textbf {\bibinfo {volume} {127}},\ \bibinfo {pages} {200502} (\bibinfo {year} {2021})}\BibitemShut {NoStop}%
\bibitem [{\citenamefont {Xiong}\ \emph {et~al.}(2022)\citenamefont {Xiong}, \citenamefont {Ficheux}, \citenamefont {Somoroff}, \citenamefont {Nguyen}, \citenamefont {Dogan}, \citenamefont {Rosenstock}, \citenamefont {Wang}, \citenamefont {Nesterov}, \citenamefont {Vavilov},\ and\ \citenamefont {Manucharyan}}]{PhysRevResearch.4.023040}%
  \BibitemOpen
  \bibfield  {author} {\bibinfo {author} {\bibfnamefont {H.}~\bibnamefont {Xiong}}, \bibinfo {author} {\bibfnamefont {Q.}~\bibnamefont {Ficheux}}, \bibinfo {author} {\bibfnamefont {A.}~\bibnamefont {Somoroff}}, \bibinfo {author} {\bibfnamefont {L.~B.}\ \bibnamefont {Nguyen}}, \bibinfo {author} {\bibfnamefont {E.}~\bibnamefont {Dogan}}, \bibinfo {author} {\bibfnamefont {D.}~\bibnamefont {Rosenstock}}, \bibinfo {author} {\bibfnamefont {C.}~\bibnamefont {Wang}}, \bibinfo {author} {\bibfnamefont {K.~N.}\ \bibnamefont {Nesterov}}, \bibinfo {author} {\bibfnamefont {M.~G.}\ \bibnamefont {Vavilov}}, \ and\ \bibinfo {author} {\bibfnamefont {V.~E.}\ \bibnamefont {Manucharyan}},\ }\href {\doibase 10.1103/PhysRevResearch.4.023040} {\bibfield  {journal} {\bibinfo  {journal} {Phys. Rev. Res.}\ }\textbf {\bibinfo {volume} {4}},\ \bibinfo {pages} {023040} (\bibinfo {year} {2022})}\BibitemShut {NoStop}%
\bibitem [{\citenamefont {Wei}\ \emph {et~al.}(2021)\citenamefont {Wei}, \citenamefont {Magesan}, \citenamefont {Lauer}, \citenamefont {Srinivasan}, \citenamefont {Bogorin}, \citenamefont {Carnevale}, \citenamefont {Keefe}, \citenamefont {Kim}, \citenamefont {Klaus}, \citenamefont {Landers} \emph {et~al.}}]{wei2021quantum}%
  \BibitemOpen
  \bibfield  {author} {\bibinfo {author} {\bibfnamefont {K.}~\bibnamefont {Wei}}, \bibinfo {author} {\bibfnamefont {E.}~\bibnamefont {Magesan}}, \bibinfo {author} {\bibfnamefont {I.}~\bibnamefont {Lauer}}, \bibinfo {author} {\bibfnamefont {S.}~\bibnamefont {Srinivasan}}, \bibinfo {author} {\bibfnamefont {D.}~\bibnamefont {Bogorin}}, \bibinfo {author} {\bibfnamefont {S.}~\bibnamefont {Carnevale}}, \bibinfo {author} {\bibfnamefont {G.}~\bibnamefont {Keefe}}, \bibinfo {author} {\bibfnamefont {Y.}~\bibnamefont {Kim}}, \bibinfo {author} {\bibfnamefont {D.}~\bibnamefont {Klaus}}, \bibinfo {author} {\bibfnamefont {W.}~\bibnamefont {Landers}},  \emph {et~al.},\ }\href@noop {} {\bibfield  {journal} {\bibinfo  {journal} {arXiv preprint arXiv:2106.00675}\ } (\bibinfo {year} {2021})}\BibitemShut {NoStop}%
\bibitem [{\citenamefont {Wei}\ \emph {et~al.}(2022{\natexlab{a}})\citenamefont {Wei}, \citenamefont {Magesan}, \citenamefont {Lauer}, \citenamefont {Srinivasan}, \citenamefont {Bogorin}, \citenamefont {Carnevale}, \citenamefont {Keefe}, \citenamefont {Kim}, \citenamefont {Klaus}, \citenamefont {Landers}, \citenamefont {Sundaresan}, \citenamefont {Wang}, \citenamefont {Zhang}, \citenamefont {Steffen}, \citenamefont {Dial}, \citenamefont {McKay},\ and\ \citenamefont {Kandala}}]{PhysRevLett.129.060501}%
  \BibitemOpen
  \bibfield  {author} {\bibinfo {author} {\bibfnamefont {K.~X.}\ \bibnamefont {Wei}}, \bibinfo {author} {\bibfnamefont {E.}~\bibnamefont {Magesan}}, \bibinfo {author} {\bibfnamefont {I.}~\bibnamefont {Lauer}}, \bibinfo {author} {\bibfnamefont {S.}~\bibnamefont {Srinivasan}}, \bibinfo {author} {\bibfnamefont {D.~F.}\ \bibnamefont {Bogorin}}, \bibinfo {author} {\bibfnamefont {S.}~\bibnamefont {Carnevale}}, \bibinfo {author} {\bibfnamefont {G.~A.}\ \bibnamefont {Keefe}}, \bibinfo {author} {\bibfnamefont {Y.}~\bibnamefont {Kim}}, \bibinfo {author} {\bibfnamefont {D.}~\bibnamefont {Klaus}}, \bibinfo {author} {\bibfnamefont {W.}~\bibnamefont {Landers}}, \bibinfo {author} {\bibfnamefont {N.}~\bibnamefont {Sundaresan}}, \bibinfo {author} {\bibfnamefont {C.}~\bibnamefont {Wang}}, \bibinfo {author} {\bibfnamefont {E.~J.}\ \bibnamefont {Zhang}}, \bibinfo {author} {\bibfnamefont {M.}~\bibnamefont {Steffen}}, \bibinfo {author} {\bibfnamefont {O.~E.}\ \bibnamefont {Dial}}, \bibinfo {author} {\bibfnamefont {D.~C.}\
  \bibnamefont {McKay}}, \ and\ \bibinfo {author} {\bibfnamefont {A.}~\bibnamefont {Kandala}},\ }\href {\doibase 10.1103/PhysRevLett.129.060501} {\bibfield  {journal} {\bibinfo  {journal} {Phys. Rev. Lett.}\ }\textbf {\bibinfo {volume} {129}},\ \bibinfo {pages} {060501} (\bibinfo {year} {2022}{\natexlab{a}})}\BibitemShut {NoStop}%
\bibitem [{\citenamefont {Wang}\ \emph {et~al.}(2023)\citenamefont {Wang}, \citenamefont {Xie}, \citenamefont {Jiang}, \citenamefont {Mandlekar}, \citenamefont {Xiao}, \citenamefont {Zhu}, \citenamefont {Fan},\ and\ \citenamefont {Anandkumar}}]{wang2023voyager}%
  \BibitemOpen
  \bibfield  {author} {\bibinfo {author} {\bibfnamefont {G.}~\bibnamefont {Wang}}, \bibinfo {author} {\bibfnamefont {Y.}~\bibnamefont {Xie}}, \bibinfo {author} {\bibfnamefont {Y.}~\bibnamefont {Jiang}}, \bibinfo {author} {\bibfnamefont {A.}~\bibnamefont {Mandlekar}}, \bibinfo {author} {\bibfnamefont {C.}~\bibnamefont {Xiao}}, \bibinfo {author} {\bibfnamefont {Y.}~\bibnamefont {Zhu}}, \bibinfo {author} {\bibfnamefont {L.}~\bibnamefont {Fan}}, \ and\ \bibinfo {author} {\bibfnamefont {A.}~\bibnamefont {Anandkumar}},\ }\href@noop {} {\bibfield  {journal} {\bibinfo  {journal} {arXiv preprint arXiv:2305.16291}\ } (\bibinfo {year} {2023})}\BibitemShut {NoStop}%
\bibitem [{\citenamefont {Wattoo}\ \emph {et~al.}(2024)\citenamefont {Wattoo}, \citenamefont {Vitis}, \citenamefont {Zhu}, \citenamefont {Depner}, \citenamefont {Zhang}, \citenamefont {Hein}, \citenamefont {Gujarati},\ and\ \citenamefont {Seltzer}}]{wattoo2024rabit}%
  \BibitemOpen
  \bibfield  {author} {\bibinfo {author} {\bibfnamefont {Z.~S.}\ \bibnamefont {Wattoo}}, \bibinfo {author} {\bibfnamefont {P.}~\bibnamefont {Vitis}}, \bibinfo {author} {\bibfnamefont {R.}~\bibnamefont {Zhu}}, \bibinfo {author} {\bibfnamefont {N.}~\bibnamefont {Depner}}, \bibinfo {author} {\bibfnamefont {I.}~\bibnamefont {Zhang}}, \bibinfo {author} {\bibfnamefont {J.}~\bibnamefont {Hein}}, \bibinfo {author} {\bibfnamefont {A.}~\bibnamefont {Gujarati}}, \ and\ \bibinfo {author} {\bibfnamefont {M.}~\bibnamefont {Seltzer}},\ }in\ \href@noop {} {\emph {\bibinfo {booktitle} {2024 54th Annual IEEE/IFIP International Conference on Dependable Systems and Networks (DSN)}}}\ (\bibinfo {organization} {IEEE},\ \bibinfo {year} {2024})\ pp.\ \bibinfo {pages} {353--361}\BibitemShut {NoStop}%
\bibitem [{\citenamefont {Leong}\ \emph {et~al.}(2024)\citenamefont {Leong}, \citenamefont {Griesbach}, \citenamefont {Zhang}, \citenamefont {Darvish}, \citenamefont {Zhao}, \citenamefont {Mandal}, \citenamefont {Zou}, \citenamefont {Hao}, \citenamefont {Bernales},\ and\ \citenamefont {Aspuru-Guzik}}]{steering2024}%
  \BibitemOpen
  \bibfield  {author} {\bibinfo {author} {\bibfnamefont {S.~X.}\ \bibnamefont {Leong}}, \bibinfo {author} {\bibfnamefont {C.~E.}\ \bibnamefont {Griesbach}}, \bibinfo {author} {\bibfnamefont {R.}~\bibnamefont {Zhang}}, \bibinfo {author} {\bibfnamefont {K.}~\bibnamefont {Darvish}}, \bibinfo {author} {\bibfnamefont {Y.}~\bibnamefont {Zhao}}, \bibinfo {author} {\bibfnamefont {A.}~\bibnamefont {Mandal}}, \bibinfo {author} {\bibfnamefont {Y.}~\bibnamefont {Zou}}, \bibinfo {author} {\bibfnamefont {H.}~\bibnamefont {Hao}}, \bibinfo {author} {\bibfnamefont {V.}~\bibnamefont {Bernales}}, \ and\ \bibinfo {author} {\bibfnamefont {A.}~\bibnamefont {Aspuru-Guzik}},\ }\href {https://chemrxiv.org/engage/chemrxiv/article-details/6764449b6dde43c908a1d8c6} {\  (\bibinfo {year} {2024})}\BibitemShut {NoStop}%
\bibitem [{\citenamefont {Wei}\ \emph {et~al.}(2022{\natexlab{b}})\citenamefont {Wei}, \citenamefont {Wang}, \citenamefont {Schuurmans}, \citenamefont {Bosma}, \citenamefont {Xia}, \citenamefont {Chi}, \citenamefont {Le}, \citenamefont {Zhou} \emph {et~al.}}]{wei2022chain}%
  \BibitemOpen
  \bibfield  {author} {\bibinfo {author} {\bibfnamefont {J.}~\bibnamefont {Wei}}, \bibinfo {author} {\bibfnamefont {X.}~\bibnamefont {Wang}}, \bibinfo {author} {\bibfnamefont {D.}~\bibnamefont {Schuurmans}}, \bibinfo {author} {\bibfnamefont {M.}~\bibnamefont {Bosma}}, \bibinfo {author} {\bibfnamefont {F.}~\bibnamefont {Xia}}, \bibinfo {author} {\bibfnamefont {E.}~\bibnamefont {Chi}}, \bibinfo {author} {\bibfnamefont {Q.~V.}\ \bibnamefont {Le}}, \bibinfo {author} {\bibfnamefont {D.}~\bibnamefont {Zhou}},  \emph {et~al.},\ }\href@noop {} {\bibfield  {journal} {\bibinfo  {journal} {Advances in neural information processing systems}\ }\textbf {\bibinfo {volume} {35}},\ \bibinfo {pages} {24824} (\bibinfo {year} {2022}{\natexlab{b}})}\BibitemShut {NoStop}%
\bibitem [{\citenamefont {Cao}\ \emph {et~al.}(2024{\natexlab{b}})\citenamefont {Cao}, \citenamefont {Lall}, \citenamefont {Bakr}, \citenamefont {Campanaro}, \citenamefont {Fasciati}, \citenamefont {Wills}, \citenamefont {Chidambaram}, \citenamefont {Shteynas}, \citenamefont {Rungger},\ and\ \citenamefont {Leek}}]{PhysRevLett.133.120802}%
  \BibitemOpen
  \bibfield  {author} {\bibinfo {author} {\bibfnamefont {S.}~\bibnamefont {Cao}}, \bibinfo {author} {\bibfnamefont {D.}~\bibnamefont {Lall}}, \bibinfo {author} {\bibfnamefont {M.}~\bibnamefont {Bakr}}, \bibinfo {author} {\bibfnamefont {G.}~\bibnamefont {Campanaro}}, \bibinfo {author} {\bibfnamefont {S.~D.}\ \bibnamefont {Fasciati}}, \bibinfo {author} {\bibfnamefont {J.}~\bibnamefont {Wills}}, \bibinfo {author} {\bibfnamefont {V.}~\bibnamefont {Chidambaram}}, \bibinfo {author} {\bibfnamefont {B.}~\bibnamefont {Shteynas}}, \bibinfo {author} {\bibfnamefont {I.}~\bibnamefont {Rungger}}, \ and\ \bibinfo {author} {\bibfnamefont {P.~J.}\ \bibnamefont {Leek}},\ }\href {\doibase 10.1103/PhysRevLett.133.120802} {\bibfield  {journal} {\bibinfo  {journal} {Phys. Rev. Lett.}\ }\textbf {\bibinfo {volume} {133}},\ \bibinfo {pages} {120802} (\bibinfo {year} {2024}{\natexlab{b}})}\BibitemShut {NoStop}%
\bibitem [{\citenamefont {Cao}\ \emph {et~al.}(2024{\natexlab{c}})\citenamefont {Cao}, \citenamefont {Bakr}, \citenamefont {Campanaro}, \citenamefont {Fasciati}, \citenamefont {Wills}, \citenamefont {Lall}, \citenamefont {Shteynas}, \citenamefont {Chidambaram}, \citenamefont {Rungger},\ and\ \citenamefont {Leek}}]{Cao2024}%
  \BibitemOpen
  \bibfield  {author} {\bibinfo {author} {\bibfnamefont {S.}~\bibnamefont {Cao}}, \bibinfo {author} {\bibfnamefont {M.}~\bibnamefont {Bakr}}, \bibinfo {author} {\bibfnamefont {G.}~\bibnamefont {Campanaro}}, \bibinfo {author} {\bibfnamefont {S.~D.}\ \bibnamefont {Fasciati}}, \bibinfo {author} {\bibfnamefont {J.}~\bibnamefont {Wills}}, \bibinfo {author} {\bibfnamefont {D.}~\bibnamefont {Lall}}, \bibinfo {author} {\bibfnamefont {B.}~\bibnamefont {Shteynas}}, \bibinfo {author} {\bibfnamefont {V.}~\bibnamefont {Chidambaram}}, \bibinfo {author} {\bibfnamefont {I.}~\bibnamefont {Rungger}}, \ and\ \bibinfo {author} {\bibfnamefont {P.}~\bibnamefont {Leek}},\ }\href {\doibase 10.1088/2058-9565/ad37d4} {\bibfield  {journal} {\bibinfo  {journal} {Quantum Science and Technology}\ }\textbf {\bibinfo {volume} {9}},\ \bibinfo {pages} {035003} (\bibinfo {year} {2024}{\natexlab{c}})}\BibitemShut {NoStop}%
\bibitem [{\citenamefont {Zhang}\ and\ \citenamefont {Cao}(2024)}]{kagentsRepo}%
  \BibitemOpen
  \bibfield  {author} {\bibinfo {author} {\bibfnamefont {Z.}~\bibnamefont {Zhang}}\ and\ \bibinfo {author} {\bibfnamefont {S.}~\bibnamefont {Cao}},\ }\href@noop {} {\enquote {\bibinfo {title} {k-agents: knowledge-based agents for automation},}\ }\bibinfo {howpublished} {\url{https://github.com/EvoEvolver/k_agents}} (\bibinfo {year} {2024})\BibitemShut {NoStop}%
\bibitem [{\citenamefont {Research}(2024)}]{NousHermes2}%
  \BibitemOpen
  \bibfield  {author} {\bibinfo {author} {\bibfnamefont {N.}~\bibnamefont {Research}},\ }\href@noop {} {\enquote {\bibinfo {title} {Nous hermes 2 - yi-34b},}\ }\bibinfo {howpublished} {\url{https://huggingface.co/NousResearch/Nous-Hermes-2-Yi-34B}} (\bibinfo {year} {2024})\BibitemShut {NoStop}%
\bibitem [{\citenamefont {Emerson}\ \emph {et~al.}(2005)\citenamefont {Emerson}, \citenamefont {Alicki},\ and\ \citenamefont {{\.{Z}}yczkowski}}]{Emerson_2005}%
  \BibitemOpen
  \bibfield  {author} {\bibinfo {author} {\bibfnamefont {J.}~\bibnamefont {Emerson}}, \bibinfo {author} {\bibfnamefont {R.}~\bibnamefont {Alicki}}, \ and\ \bibinfo {author} {\bibfnamefont {K.}~\bibnamefont {{\.{Z}}yczkowski}},\ }\href {\doibase 10.1088/1464-4266/7/10/021} {\bibfield  {journal} {\bibinfo  {journal} {Journal of Optics B: Quantum and Semiclassical Optics}\ }\textbf {\bibinfo {volume} {7}},\ \bibinfo {pages} {S347} (\bibinfo {year} {2005})}\BibitemShut {NoStop}%
\bibitem [{\citenamefont {Dankert}\ \emph {et~al.}(2009)\citenamefont {Dankert}, \citenamefont {Cleve}, \citenamefont {Emerson},\ and\ \citenamefont {Livine}}]{Unitary2designRB}%
  \BibitemOpen
  \bibfield  {author} {\bibinfo {author} {\bibfnamefont {C.}~\bibnamefont {Dankert}}, \bibinfo {author} {\bibfnamefont {R.}~\bibnamefont {Cleve}}, \bibinfo {author} {\bibfnamefont {J.}~\bibnamefont {Emerson}}, \ and\ \bibinfo {author} {\bibfnamefont {E.}~\bibnamefont {Livine}},\ }\href {\doibase 10.1103/PhysRevA.80.012304} {\bibfield  {journal} {\bibinfo  {journal} {Phys. Rev. A}\ }\textbf {\bibinfo {volume} {80}},\ \bibinfo {pages} {012304} (\bibinfo {year} {2009})}\BibitemShut {NoStop}%
\bibitem [{\citenamefont {Magesan}\ \emph {et~al.}(2012)\citenamefont {Magesan}, \citenamefont {Gambetta}, \citenamefont {Johnson}, \citenamefont {Ryan}, \citenamefont {Chow}, \citenamefont {Merkel}, \citenamefont {da~Silva}, \citenamefont {Keefe}, \citenamefont {Rothwell}, \citenamefont {Ohki}, \citenamefont {Ketchen},\ and\ \citenamefont {Steffen}}]{InterleavedRB}%
  \BibitemOpen
  \bibfield  {author} {\bibinfo {author} {\bibfnamefont {E.}~\bibnamefont {Magesan}}, \bibinfo {author} {\bibfnamefont {J.~M.}\ \bibnamefont {Gambetta}}, \bibinfo {author} {\bibfnamefont {B.~R.}\ \bibnamefont {Johnson}}, \bibinfo {author} {\bibfnamefont {C.~A.}\ \bibnamefont {Ryan}}, \bibinfo {author} {\bibfnamefont {J.~M.}\ \bibnamefont {Chow}}, \bibinfo {author} {\bibfnamefont {S.~T.}\ \bibnamefont {Merkel}}, \bibinfo {author} {\bibfnamefont {M.~P.}\ \bibnamefont {da~Silva}}, \bibinfo {author} {\bibfnamefont {G.~A.}\ \bibnamefont {Keefe}}, \bibinfo {author} {\bibfnamefont {M.~B.}\ \bibnamefont {Rothwell}}, \bibinfo {author} {\bibfnamefont {T.~A.}\ \bibnamefont {Ohki}}, \bibinfo {author} {\bibfnamefont {M.~B.}\ \bibnamefont {Ketchen}}, \ and\ \bibinfo {author} {\bibfnamefont {M.}~\bibnamefont {Steffen}},\ }\href {\doibase 10.1103/PhysRevLett.109.080505} {\bibfield  {journal} {\bibinfo  {journal} {Phys. Rev. Lett.}\ }\textbf {\bibinfo {volume} {109}},\ \bibinfo {pages} {080505} (\bibinfo {year}
  {2012})}\BibitemShut {NoStop}%
\end{thebibliography}%
